\documentclass[preprint,12pt,dvipdfmx]{elsarticle}

\usepackage{amssymb}
\usepackage{amsthm}
\usepackage{algorithm}
\usepackage{algorithmic}
\usepackage{colortbl}
\usepackage{url}
\usepackage{doi}
\usepackage{pdfpages}
\usepackage{booktabs}

\usepackage{amsmath} 
\usepackage{rotating}
\usepackage{lscape}
\usepackage{multirow}

\journal{Applied Soft Computing}

\newcommand{\del}[1]{}

\newcommand{\onodel}[1]{}

\newcommand{\comid}[1]{}

\newlength{\minitwocolumn}
\newlength{\minithreecolumn}
\newcommand{\lw}[1]{\smash{\lower1.8ex\hbox{#1}}}

\graphicspath{{./Figs/}}

\def\Vec#1{\mbox{\boldmath $#1$}}

\newcommand{\figd}[1]{}

\makeatletter
\newcommand{\figcaption}[1]{\def\@captype{figure}\caption{#1}}
\newcommand{\tblcaption}[1]{\def\@captype{table}\caption{#1}}
\makeatother

\begin{document}
\begin{frontmatter}

\title{%
Black-box Adversarial Attacks on Monocular Depth Estimation Using
Evolutionary Multi-objective Optimization
}

\author{Renya Daimo$^1$,Satoshi Ono$^1$,Takahiro Suzuki$^1$}

\address{%
$^1$
Department of Information Science and Biomedical Engineering,
Graduate School of Science and Engineering, Kagoshima University\\
1-21-40, Korimoto, Kagoshima, 890-0065 Japan
}

\begin{abstract}
This paper proposes an adversarial attack method to deep neural
networks (DNNs) for monocular depth estimation, i.e., estimating the
depth from a single image.
Single image depth estimation has improved drastically in recent years
due to the development of DNNs.
However, vulnerabilities of DNNs for image classification have been
revealed by adversarial attacks, and DNNs for monocular depth
estimation could contain similar vulnerabilities.
Therefore, research on vulnerabilities of DNNs for monocular depth
estimation has spread rapidly, but many of them assume white-box
conditions where inside information of DNNs is available, or are
transferability-based black-box attacks that require a substitute DNN
model and a training dataset.
Utilizing Evolutionary Multi-objective Optimization, the proposed
method in this paper analyzes DNNs under the black-box condition where
only output depth maps are available.
In addition, the proposed method does not require a substitute DNN
that has a similar architecture to the target DNN nor any knowledge
about training data used to train the target model.
Experimental results showed that the proposed method succeeded in
attacking two DNN-based methods that were trained with indoor and
outdoor scenes respectively. 

\end{abstract}
 
 \end{frontmatter}
 
\section{INTRODUCTION
  }
Monocular depth estimation, i.e. estimating the scene depth from a
single image, has been widely studied because it is one of the fundamental
problems in computer
vision~\cite{laga2019survey,bhoi2019monocular,KOCH2020102877} and it
includes various applications such as 
robotics, image refocusing, scene
understanding, and so on\cite{bhoi2019monocular,Shi2015,VanHecke2017,Mancini2018}.

Recent advances in Deep Neural Networks (DNNs) have dramatically
improved the
performance~\cite{eigen2014depth,laina2016deeper,casser2019depth,hu2019revisiting}
and increased expectations for its applications to real-world problems.

On the other hand, some studies revealed that DNNs have a particular
vulnerability in which an input image carefully perturbed by an
attacker causes misrecognition~\cite{goodfellow2014explaining}.
Similar vulnerability is expected to exist in DNNs for monocular depth
estimation, and recent studies about adversarial attack methods for DNNs for
monocular depth estimation began to take
place~\cite{weko_200840_1,hu2019analysis}.
The above studies assume a white-box condition where internal
information of DNNs can be used to generate adversarial examples, and
are useful when developers of the DNNs verify the vulnerability by
themselves.
However, it is diffuicult for a third party to assess the security of
the DNNs without permission to the internal information of the DNN.
Since 2020, transferability-based black-box attack methods have been
proposed
\cite{zhang2020adversarial,mathew2020monocular,wong2020targeted};
however, the methods require a substitute DNN model with a
similar architecture to a target DNN and a dataset that is also
the same or similar to the one  used to train the target
DNN.

Therefore, this study proposes an adversarial attack method for
verifying the vulnerability of DNNs for monocular depth estimation
under a black-box condition where the inside information of DNNs is
not available.
Unlike object recognition, DNNs for monocular depth estimation produce
a depth map, i.e., estimated pixel-wise depth values, without
confidential scores.
The proposed method can generate adversarial examples only by
referring to the depth maps because
depth estimation is essentially a pixel-wise regression problem, whose
output can be represented as continuous values.

To analyze the DNNs described above, the proposed method employs
black-box optimization-based approach to design adversarial
perturbations on textures of a target object.
In particular, we adopt Evolutionary Multi-objective Optimization
(EMO) that simultaneously minimizes the difference of depth
values between the estimated depth map and a target map in addition to
the perturbation amount.

The contribution of this paper summarized as follows:
\begin{itemize}

\item To the best of our knowledge, this is the first study about
  black-box optimization-based adversarial attacks that is not based
  on the transferability between DNN-based monocular depth estimation
  models.
  The proposed method can be applicable to proprietary services and
  systems.

\item Black-box global optimization-based approach the proposed method
  employs does not require a substitute DNN to approximate the target
  DNN model, nor any information about training datasets the target
  model used.

\item The proposed method adds perturbations only on the target
  object, so it can be developed into a physical attack.
In addition, unlike the universal adversarial patch, the perturbation
amount is moderate and varies depending on the target object, making
it more difficult to defend.

\item Two DNN models trained with indoor and outdoor scenes
  respectively have analyzed by the proposed method.
  By adding perturbations only to the target object texture, without
  the background or other objects, the proposed method produced
  adversarial examples where DNNs falsely predicted depth maps as if
  the target object had disappeared.

\end{itemize}

The rest of this paper organized as follows:
Sec.~\ref{sec:related_work} introduces some recent DNN-based monocular
depth estimation models and representative adversarial attack methods.
Sec.~\ref{sec:proposed_method} describes the key ideas of the proposed
method, the formulation of adversarial example generation as a
multi-objective optimization problem, and the detailed algorithm of
the proposed method.
Sec.~\ref{sec:experiment} demonstrates the effectiveness of the
proposed method by analyzing two DNN-based monocular depth estimation
models.
%

\section{RELATED WORK}
\label{sec:related_work}

  \subsection{Monocular depth estimation using DNNs
  }
Recent remarkable advances in DNNs have led to increasing studies on
monocular depth
estimation~\cite{laina2016deeper,eigen2014depth,godard2019digging,casser2019depth,hu2019revisiting}.
Eigen et al. proposed a method using a DNN consistng of two networks
that perform global coarse-scare and local fine-scale
prediction~\cite{eigen2014depth}.
Laina et al. employed a fully convolutional network that enables higher
resolution output while suppressing the number of parameters to be
trained~\cite{laina2016deeper}.
Godard et al. adopted a self-supervised learning approach that achieved
sufficient accuracy~\cite{godard2019digging}.
Hu et al. proposed a method of monocular depth estimation that
produces a depth map with more accurate object boundaries by focusing
multi-scale features and loss functions based on surface
normals~\cite{hu2019revisiting}.

With the development of monocular depth estimation technology,
research on its analysis is also being conducted.
Dijk et al. showed that a model trained by in-vehicle camera scenes
estimates an object depth by focusing on the vertical position of the
object in an image while ignoring the image region size of the object,
and that the model only partially recognizes changes of the pitch and
roll angles of the camera~\cite{dijk2019neural}.
Hu et al. clarified that DNN-based models select important edges
independently from the edge strength and focuses not only on the boundary
but the inside region of the objects~\cite{hu2019visualization}.
They also showed that the regions around vanishing points are
prioritized in the outdoor scenes.

  \subsection{Adversarial attacks for DNN
  }
DNNs achieved state-of-the-art performance not only in monocular depth
estimation but in various domains of computer vision~\cite{chen2017rethinking,redmon2017yolo9000},
speech recognition~\cite{hannun2014deep,povey2011kaldi}, and so on; 
however, their vulnerability against adversarial perturbations has
been
clarified~\cite{goodfellow2014explaining,papernot2016limitations,suzuki2019adversarial}.
The early study by Goodfellow et al. proposed the Fast Gradient Sign
Method (FGSM) to generate adversarial examples for image
classification~\cite{goodfellow2014explaining}.
Carlini et al. proposed C\&W attack that formulates adversarial
example generation as an optimization
problem~\cite{carlini2017towards}.

Some black-box attack methods have also been studied that does not
require internal information such as gradients of loss functions.
One main approach for black-box attack is transferability-based approach.
Papernot et al. proposed a method to generate AE under BB conditions
by constructing an alternative model of the target NN model
\cite{papernot2016limitations}.
Liu et al. proposed an ensemble-basd method to generate transferable
adversarial examples~~\cite{liu2016delving}.
Recently, boundary attack methods have been widely studied, which
originated in Brendel et al.~\cite{brendel2017decision}, which assumes
hard-label black-box condition where only an assigned label is
available.

The third approach, alongside the transferability-based approach and boundary
attack, is black-box optimization-based approach, which originated in the
study by Su et al.~\cite{Su2017}.
They proposed one pixel attack method using Differential
Evolution~\cite{Storn1997} that modifies very few pixels to create
adversarial examples.
Suzuki et al. proposed a method using EMO that can generate robust
adversarial examples against image conversion under black-box
conditions~\cite{suzuki2019adversarial}.
Kuang et al. proposed a query efficient black-box optimization-based attack
method using Covariance Matrix Adaptation Evolution
Strategy~\cite{Kuang2019}.
The drawback of of this approach is the computational cost, which
limits the scenarios where the attack methods are applicable;
however, there are certain scenarios where this approach is useful.
For instance, with the permission of the system developer, a large
number of queries are allowed for the third parties to externally
inspect the system for vulnerabilities.
In particular, it is expected for the global black-box
optimization-based approach to find unknown vulnerabilities different
from ones found by transfer-based and boundary-based attacks.

Since 2019, research on adversarial attacks for monocular depth
estimation has also begun.
Yamanaka et al. showed that the depth of the target area is
erroneously estimated by overwriting the local area of the input
image~\cite{weko_200840_1}.
Hu et al. attempted to analyze the mechanism of monocular depth
estimation by generating adversarial examples~\cite{hu2019analysis}.
Zhang et al. proposed a multi-task attack method based on
transferability-based approach that can be applicable to black-box
condition~\cite{zhang2020adversarial}.
Mathew et al. also proposed a transferability-based black-box attack method
generating adversarial patch and investigated cross-data
transferability where the same DNN model with different two training
datasets, assumed that one is opened and another is
proprietary~\cite{mathew2020monocular}.
Wong et al. added perturbations on the whole area of the
captured image to achieve target attacks~\cite{wong2020targeted}. 
They also tested a black-box attack using transferability between two
models developed by the same researchers but reported that the
perturbations optimized for one model does not transfer to another.

The above studies assumed white-box approach were difficult to apply
to commercial services and systems, where the DNNs' internal
information cannot be available, or performed transferability-based
black-box attack that requires  a DNN model with an architecture
similar to the target model and a dataset similar to ones used to train
the target model.
In addition, no studies have conducted for techniques for adding
perturbations only on the target object.

\section{The proposed black-box global optimization-based adversarial attack method
}
\label{sec:proposed_method}

  \subsection{Key Ideas
  }
This paper proposes a black-box adversarial attack method against DNNs
for monocular depth estimation.
Followings are the key ideas of the proposed method.

\begin{description}

\item[\parbox{1\textwidth}{%
Idea 1:
Employing Evolutionary Multi-objective Optimization (EMO) algorithm:%
}]~\\
EMO is a kind of black-box global optimization research field in
evolutionary computation.
Because EMO performs black-box global optimization without any prior
to a target problem, unlike transferability-based approach, the
proposed method does not require a substitute DNN that should have
a similar architecture to a target DNN nor the dataset as ones used to
train the target.

The main advantage of EMO is that it can simultaneously optimize more
than one objective function such as performance versus
cost~\cite{coello2006evolutionary}.
Creating adversarial examples naturally involves multiple objective
functions in trade-off relationship, such as accuracy versus
visibility.
A simple way to minimize the weighted linear sum of the functions can
combine them into a single objective function; however, the weight
parameters must be adjusted and the obtained solutions would be biased 
by the predetermined weights.
Therefore, the proposed method formulates the problem as a
multi-objective optimization and solves it using EMO.
Because various objective funcions including non-differentiable,
multimodal, and noisy ones can be employed, the proposed method is
suitable for seeking unknown vulnerabilities in the monocular depth
estimator.

\item[\parbox{1\textwidth}{%
Idea 2:
Dimension reduction technique based on block-wise perturbation patterns
is introduced:}]~\\
Naive formulation of adversarial example generation enlarges the
problem size.
Therefore this paper introduces block-wise perturbation generation, in
which several small perturbation patterns are created while assigning
appropriate perturbation patterns for all image blocks.

\item[\parbox{1\textwidth}{%
Idea 3:
Add perturbations to textures of a target object:}]
Instead of adding perturbations to the entire region of a captured
image, the proposed method perturbs textures of the target object.
This is because, as discussed in the previous
work~\cite{hu2019visualization}, textures of object surfaces play an
important role in monocular depth estimation in addition to its edges.
Furthermore, by adding perturbations only to the object textures, the
proposed method can be applicable to physical adversarial
attacks~\cite{song2018physical,eykholt2018robust} in future.
\end{description}

 \subsection{Formulation
  }
  \label{ssec:formulation}

The proposed method introduces block-wise perturbation patterns, where
assignments to image blocks of target object textures are
$N_{pat} \times N_{pat}$ pixel-block perturbation patterns and their
simultaneously optimized.
The above scheme allows reducing the number of variables even when 
high-resolution textures are available.

\begin{figure}[t]
      \centering
      \includegraphics[width=1\textwidth]{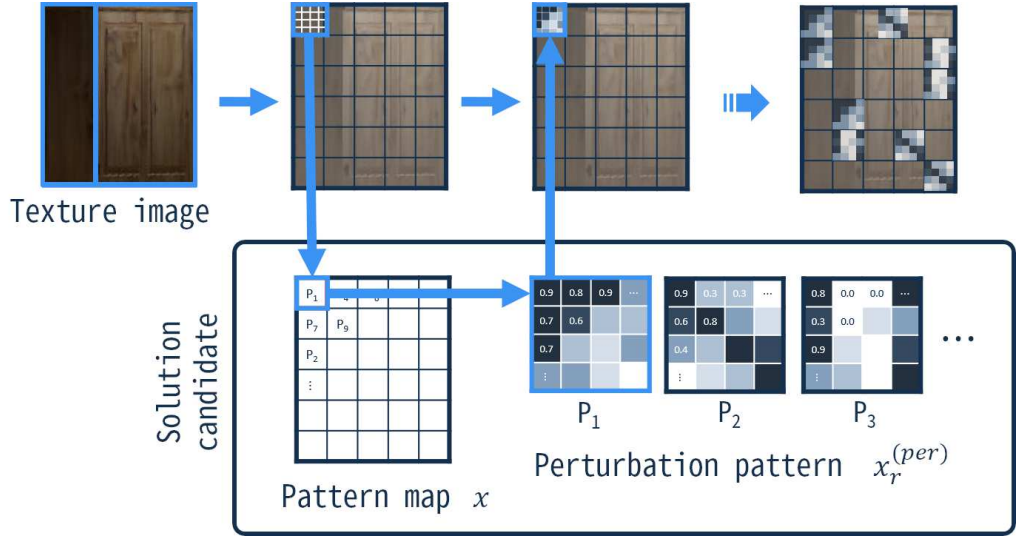}
      \caption{Block-wise perturbation patterns.}
      \label{fig:pattern_map}
    \end{figure}

Fig.\ref{fig:pattern_map} illustrates the block-wise perturbation
patterns.
There are two types of design variables comprising a solution
candidate $\Vec{\chi}$ as follows:
    \begin{eqnarray}
      \Vec{\chi} &=& \Vec{x}^{(map)} \cup \left\{  \Vec{x}^{(pat)}_r   \right\}_{  r \in \{1, \ldots, N_{AP}\}}\\
      \Vec{x}^{(map)} &=& \{x_{u,v}^{(map)}\}_{(u,v) \in \Vec{I}}  \\
      \Vec{x}_r^{(pat)}  &=& \{x_{p,q,r}^{(pat)}\}_{p, q \in \{1,\ldots,N_{pat}\}}
    \end{eqnarray}
Variable $x_{u,v}^{(map)}$ represents a pattern assignment map that
block $(u,v)$ in the texture image $\Vec{I}$, i.e., 
$x_{u,v}^{(map)} \in \{0, 1, 2, \ldots, N_{AP} \}$ .
If $x_{u,v}^{(map)} > 0$ the corresponding block-wise perturbation
pattern is applied to block $(u, v)$, otherwise, no perturbation is
added to the block.
Variable $x_{p,q,r}^{(pat)}$ represents $r$-th block-wise perturbation
pattern, that is, variation amount of intensity of pixel $(u, v)$ in
the local coordinate of pattern $r$ ($r \in \{1,\ldots,N_{AP}\}$).

The proposed method simultaneously minimizes two objective functions: 
the depth estimation error and the perturbation amount.
The proposed method mainly focuses on a target attack, that is, given
a target depth map, the proposed method generated an adversarial
example such that a target DNN estimates a depth map as close as
possible to the given depth map.

The first objective function $f_1$ is the sum of absolute error between the
target and estimated depth values.
\begin{equation}
  \label{fun:objectiveFunction}
        {\rm minimize}~ f_1(\Vec{\chi})
        = \sum_{(w,h) \in \Vec{I}}
        \left| d_{w,h}^{(est)}(\Vec{I} + \Vec{\rho}(\Vec{\chi})) - d_{w,h}^{(target)} \right|
\end{equation}
where $d_{w,h}^{(target)}$ and $d_{w,h}^{(est)}$ denotes the target
and estimated depth values at pixel $(w,h) \in \mathcal{R}$, respectively, and 
$\mathcal{R}$ denotes the image region of the target object.
The second objective function $f_2$ is the L2 norm of the perturbation $\Vec{\rho}(\Vec{\chi})$ 
generated by solution $\Vec{\chi}$
\begin{equation}
  \label{eq:l2}
        {\rm minimize}~ f_2(\Vec{\chi}) = || \Vec{\rho}(\Vec{\chi}) ||_2 
\end{equation}

\subsection{Process flow
  }
The proposed algorithm can adopt any evolutionary multi-objective
optimization algorithms such as NSGA-II\cite{Deb2002} and
MOEA/D\cite{Zhang2007b}.
In this study we employ MOEA/D to simultaneously optimize the two
objective functions described in Sec.~\ref{ssec:formulation}.

MOEA/D converts the approximation problem of the true Pareto Front
into a set of single-objective optimization problems, for instance:
\begin{eqnarray}
 {\rm minimize} && g(\Vec{x} | \Vec{\lambda}^j, z^*) 
                   = \max_{ i \in \{1,\ldots,N_f\} }  
                     \left\{ \lambda_i^j | f_i(\Vec{x}) - z_i^*  \right\}
  \nonumber \\
 {\rm subject~to} && \Vec{x} \in \Vec{\mathcal{F}}
 \end{eqnarray}
where $\Vec{\lambda}^j = (\lambda_1^j, \ldots, \lambda_{N_f}^j )$ are weight
vectors ($\lambda_i^j \geq 0$
) and 
$\sum_{i=1}^{N_f} \lambda_i^j = 1$, and 
$\Vec{z}^*$ is a reference point in the objective function space calculated as follows:
\begin{equation}
z_i^* = \min\{ f_i(\Vec{x}) |  x \in \Vec{\mathcal{F}} \} 
\label{eq:reference_point}
\end{equation}
By preparing $N_D$ weight vectors and optimizing $N_D$ scalar objective
functions, MOEA/D finds various non-dominated solutions at one
optimization.

At the beginning, MOEA/D determines the neighborhood relations for
each weight vector $\Vec{\lambda}_i$ by calculating the Euclidean
distance between them so that $N_n$ neighboring weight vectors
$\{\lambda^k\}$ are selected.
Then MOEA/D generates an initial population of solution candidates
$\Vec{\chi}_1, \ldots \Vec{\chi}_{N_p}$ by sampling at uniform random
from $\Vec{\mathcal{F}}$.
Then, the initial reference point is calculated by
eq.~(\ref{eq:reference_point}).

    \begin{figure}[t]
      \centering
      {\footnotesize
      \begin{tabular}{@{}c@{~}c@{}}
        {\includegraphics[width=0.35\textwidth]{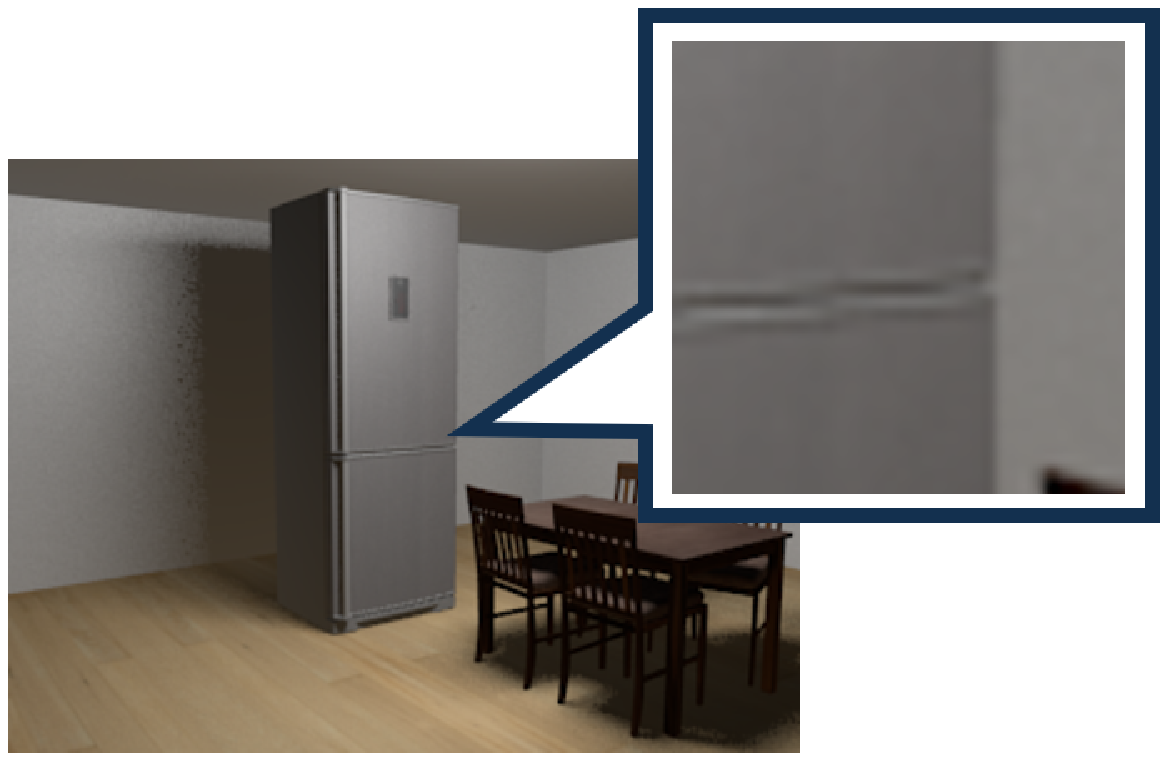}} &
        {\includegraphics[width=0.35\textwidth]{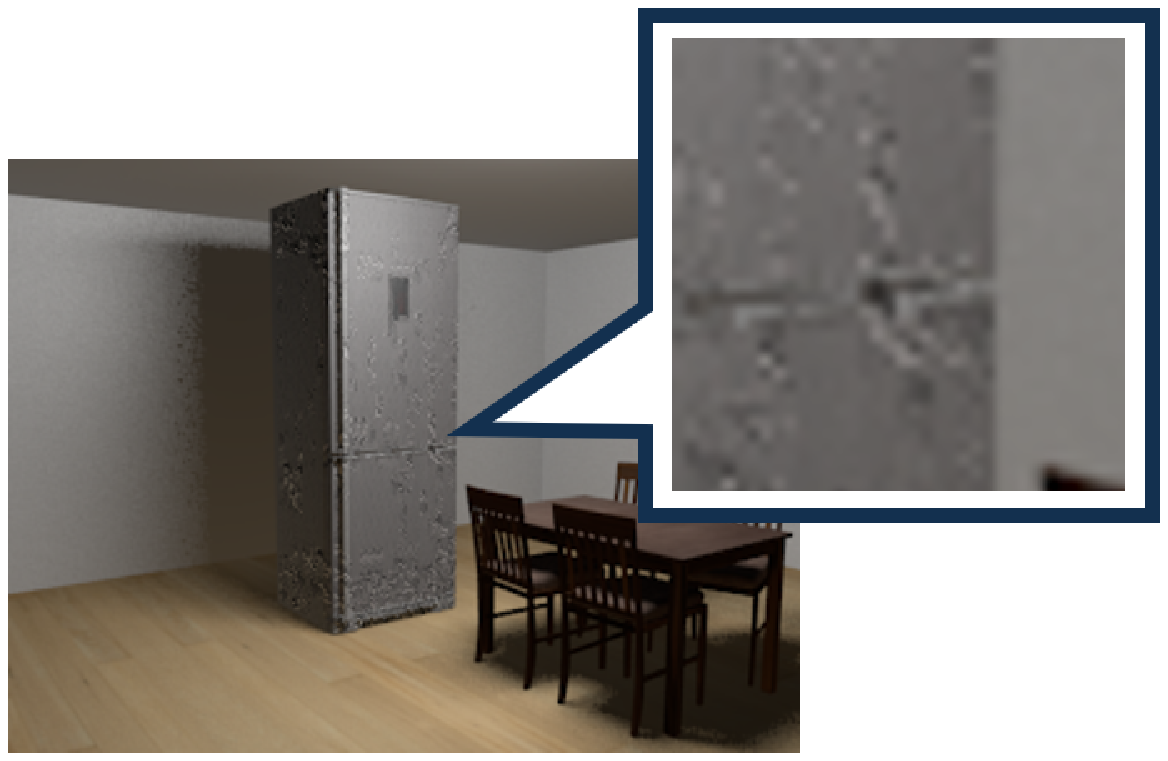}}\\
        ~\\
        \multicolumn{1}{c}{\footnotesize (a) Original image} &
        \multicolumn{1}{c}{\footnotesize (b) Adversarial example} \\
        ~\\
      \end{tabular}}
      \caption{Example of an adversarial perturbation generated by the proposed method.}
      \label{fig:result_generated_AE}
    \end{figure}

After initialization, MOEA/D repeats the processes of selection and
population update including solution generation, evaluation, and solution update.
In the selection process, $N_f$ best individuals are selected for
$N_f$ objective functions respectively, and then, indexes of the
subproblems are selected by tournament selection.
Then the population is updated by applying crossover and mutation
operators for each selected subproblem $i$.
Before applying the operators, mating and update range is selected.
That is, with the probability $\delta$, the update range $\mathcal{P}$
is limited to the neighborhood of $i$, otherwise $\mathcal{P}=
1,\ldots,N_d$.
Then, crossover is performed to randomly selected two indices
 $r_1$ and $r_2$, 
and generates a solution $\Vec{\psi}$ by  
the same way as Differential Evolution (DE)~\cite{Storn1997}.
That is, the component $\psi_k$ of $\Vec{\psi}$ is calculated as
follows:
\begin{equation}
\bar{y}_k = \left\{
\begin{array}{ll}
 \chi_k^{i} + F \left( \chi_k^{r_1} - \chi_k^{r_2} \right) & {\rm with~probability~} CR \\
 \chi_k^{i}                                             & {\rm with~probability~} 1 - CR \\
\end{array}
\right.
\end{equation}
where $F$ and $CR$ are control parameters in DE.
Following the crossover, polynomial mutation~\cite{Deb1996} is applied to
$\psi_k$ with the probability $p_m$ as follows:
\begin{eqnarray}
\psi_k &=& \bar{\psi}_k + \bar{\delta} \Delta_{max} \\
\bar{\delta} &=& \left\{
 \begin{array}{ll}
  (2 u)^\frac{1}{n + 1} - 1       & {\rm if}~ u < 0.5 \\
  1 - [2 (1 - u)]^\frac{1}{n + 1}  & {\rm otherwise}
 \end{array}			
\right.
\end{eqnarray}
where $\Delta_{max}$ represents the maximum permissible perturbance in
the parent value $\bar{\psi}_k$, and $u$ is a random number in
$[0,1]$.

Then the generated solution $\Vec{\psi}$ is evaluated.
The perturbation pattern $\Vec{\rho}$ is generated and the target DNN
model is applied to the perturbed image $\Vec{I} + \Vec{\rho}$.
From the obtained depth map, values of $f_1(\Vec{\chi})$ and
$f_2(\Vec{\chi})$ are calculated.
From the evaluation results, the component $z_j$ of the reference
point is replaced with $f_j(\Vec{\psi})$ if $z_j > f_j(\Vec{\chi})$
for each $j \in \{1, \ldots, N_f \}$.
At the end of the main loop, 
the population is updated by the following procedure:
{\renewcommand{\labelenumi}{(\arabic{enumi})}
\begin{enumerate}
 \item Set $c = 0$.
 \item If $c = n_r$ or $\mathcal{P}$ is empty, then go to
	   (4). Otherwise, pick an index $k$ from $\mathcal{P}$ at random.

 \item If any of the following conditions are satisfied, then replace
	   $\Vec{\chi}^k $ with $\Vec{\psi}$ and set $c = c + 1$.
	   \begin{equation}
		\Vec{\psi} \not\in \Vec{\mathcal{F}} \land 
		 \Vec{\chi}^k \not\in \Vec{\mathcal{F}} \land
		 vio(\Vec{\psi}) < vio(\Vec{\chi}^k)		
	   \end{equation}
	   \begin{equation}
		\Vec{\psi} \in \Vec{\mathcal{F}} \land 
		 \Vec{\chi}^k \not\in \Vec{\mathcal{F}}
	   \end{equation}
	   \begin{equation}
		\Vec{\psi} \in \Vec{\mathcal{F}} \land
		 \Vec{\chi}^k \in \Vec{\mathcal{F}} \land
		 g(\Vec{\psi} | \lambda^k, \Vec{z})
		 \leq g(\Vec{\chi}^k | \lambda^k, \Vec{z})
	   \end{equation}
	   where $vio(\cdot)$ denotes the amount of constraint
	   violations.

 \item Remove $k$ from $\mathcal{P}$ and go back to (2).
\end{enumerate}
}

    \begin{table}[t]
      \centering
      \caption{DNN models analyzed in this study.}
      {\footnotesize
      \begin{tabular}{@{}l@{~}|@{~}p{50mm}@{~}|@{~}p{25mm}@{~}|@{~}c@{~}c@{}} \hline
        Experiment & Model & data set & input / output \\ 
            &       &          & image size [px] \\ \hline
        1 & Full convolution architecture~\cite{laina2016deeper} & NYU Depth v2 & $228\times 304$ / \\
          &  &            & $160\times 124$\\
        \hline 
        2 & Self-Supervised Training \cite{godard2019digging} & KITTI & $640\times 192$ / \\
          &  &       & $640\times 192$ \\
        \hline 
      \end{tabular}}
      \label{tab:Target model}
    \end{table}

\section{Evaluation
}
\label{sec:experiment}

In order to verify the effectiveness of the proposed method, we
analyzed vulnerabilities of two DNN models trained with indoor and
outdoor scenes respectively (Table~\ref{tab:Target model}) by
generating adversarial examples.
The target depth maps are designed as if target objects were removed
from the scenes.
First, the DNN model trained with indoor scenes was  analyzed 
while changing 
target object type
 (Experiment 1-a),
and the presence of objects other than the target
 (Experiment 1-b)
 by creating scenes using CG.
Then, the real images of the actual scenes were tested 
(Experiment 1-c).
Finally, another DNN model trained with outdoor scenes was analyzed
(Experiment 2),
and the proposed method was compared with the previous method 
(Experiment 3).

In this study, the proposed method 
was configured as follows:
the Chebyshev method was selected as the scalarization function, the
neighborhood size $N_n$ and the number of objective functions $N_f$
were set to 10 and 2, respectively, and, $\delta = 0.8$, and $n_r= 1$.
The number of solution candidates $N_{p}$ and
the limit of the number of generations were set to 100 and 5,000, respectively.
The number of block-wise perturbation patterns $N_{AP}$ and its size
$N_{pat}$ were set to 10 and $8\times8$, respectively.

  \subsection{Experiment 1: analyzing DNN trained with indoor scenes
  }
  \subsubsection{Experimental setup}~\\

In Experiment 1, we analyzed Laina's
method~\cite{laina2016deeper} trained using the indoor scene dataset
NYU Depth v2.
We employed the trained model~\footnote{%
\url{https://github.com/iro-cp/FCRN-DepthPrediction} 
}, and did not train the model by ourselves.
In Experiments 
1-a and 1-b, 
to control the components comprising indoor
scenes, such as, walls, ceilings and floors of the room, and furniture
items, CG images were rendered to create simple scenes containing
the floor, walls,
ceiling, and furniture items.
Each target object was approximated as a rectangular shape, and
perturbations were added to its textures of the two surfaces (front and
side) obtained by projective transformation.
Fig.\ref{fig:result_generated_AE}(a) shows an example scene consisting
of one target object (a refrigerator) and other objects (a dining
table and chairs), and Fig.~\ref{fig:result_generated_AE}(b) shows
generated adversarial examples.
In order to quantify the effect of generated adversarial examples on
depth estimation errors, we focused on a pseudo volume of a perturbed
object that is defined as the sum of differences between the estimated
depth of the front side of the target object and a given target depth
$d^{(target)}_{(i,j)}$ for pixels in the target object region
$\mathcal{R}$ as follows:
\begin{eqnarray}
  \label{fun:val_fun}
  V = \Sigma_{(i,j) \in \mathcal{R}}|d^{(est)}_{(i,j)}-d^{(target)}_{(i,j)}|,
\end{eqnarray}
which is essentially the same as $f_1$ but the region of interest is different.

\subsubsection{Experiment 1-a: Influence of target object shape 
  }~\\
In order to investigate the influence of the shape of the target
object, we created scenes
that
have different objects such as 
a wardrobe,
a refrigerator,
a sofa, and a dining table set, 
while the viewpoint and the light source were
fixted.
In scene 
$S^{(in)}_{5}$, 
perturbation was added only to the top side of
the dining tables set, whereas perturbation was added to two sides 
($S^{(in)}_1,\ldots,S^{(in)}_{3}$)
or
three sides
($S^{(in)}_4$).

\begin{figure*}[t]
      \centering
      {\footnotesize
      \begin{tabular}{@{}l@{~}|@{~~~~}c@{~}c@{~}|@{~~~}c@{~}c@{~}|p{18mm}@{}} \hline
         Scene & \multicolumn{2}{c@{~}|@{~~~}}{Images} & \multicolumn{2}{c@{~}|}{Depth maps} & Pseudo volume\\
               &Original & Adversarial ex. & Original & Adversarial ex.     & reduction rate\\
        \hline
        
        $S^{(in)}_1$
        & {\includegraphics[width=26mm]{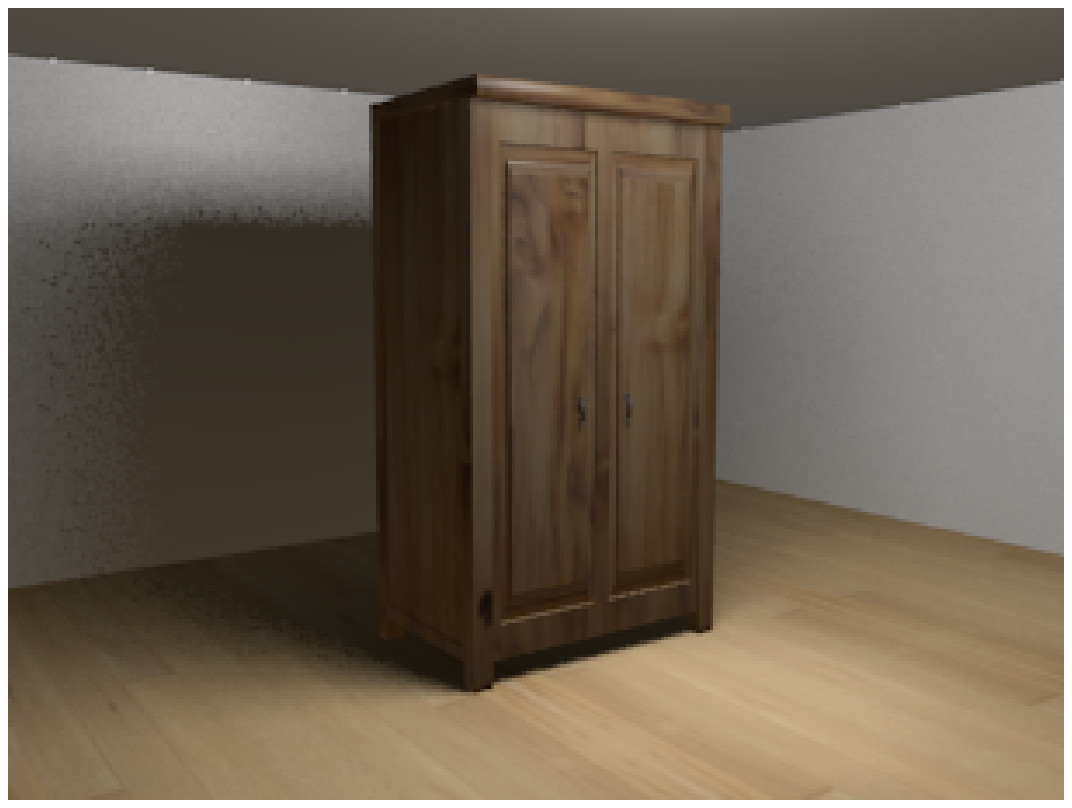}}
        & {\includegraphics[width=26mm]{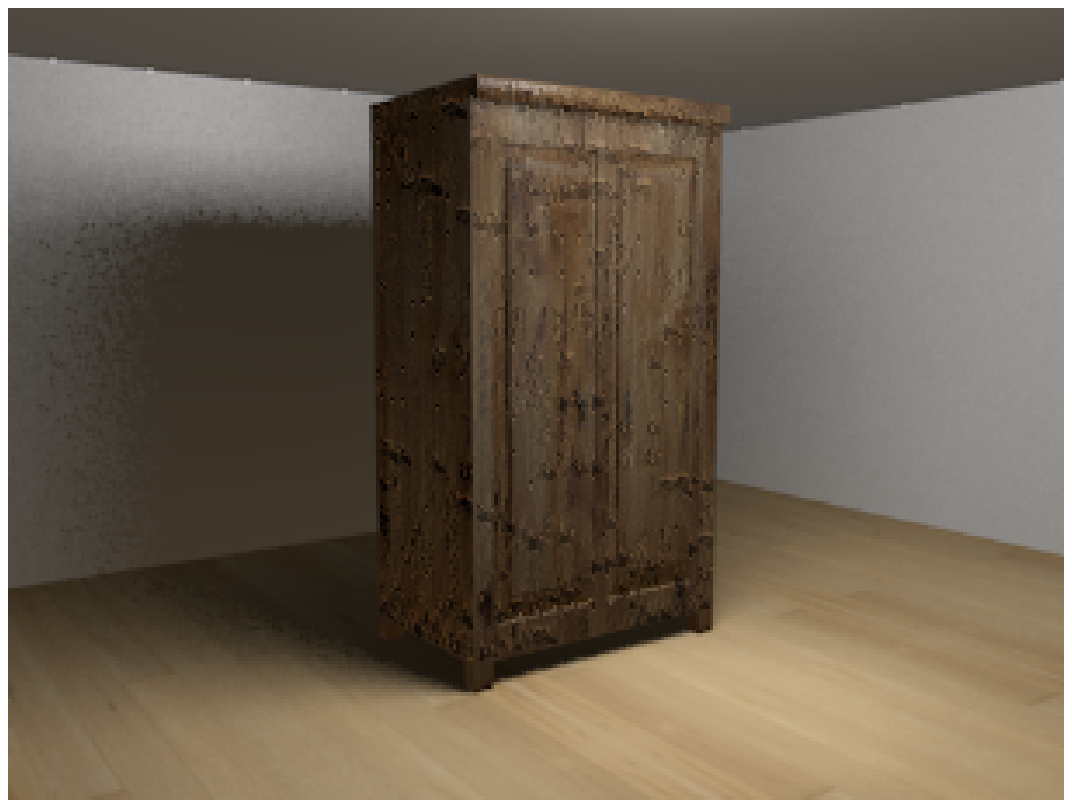}}
        & {\includegraphics[width=26mm]{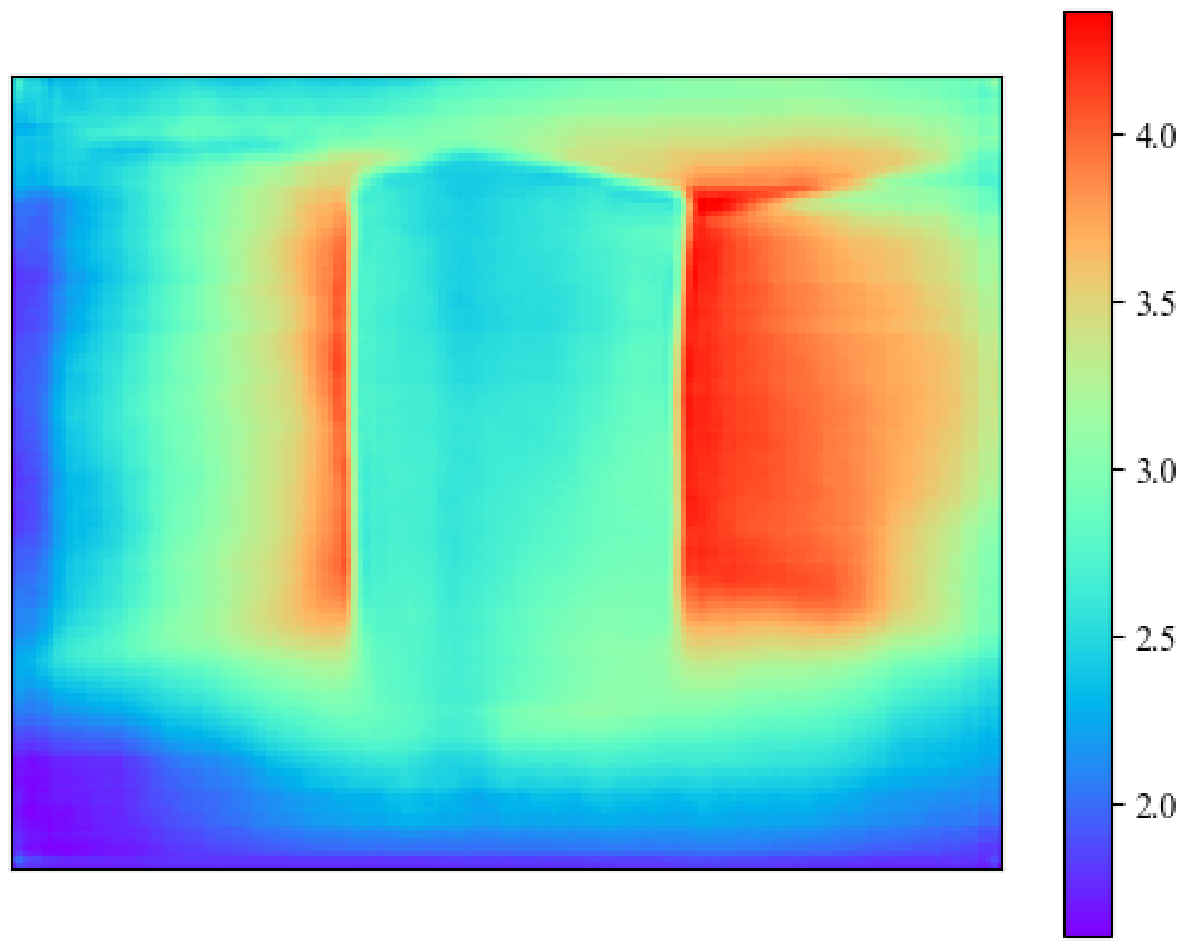}}
        & {\includegraphics[width=26mm]{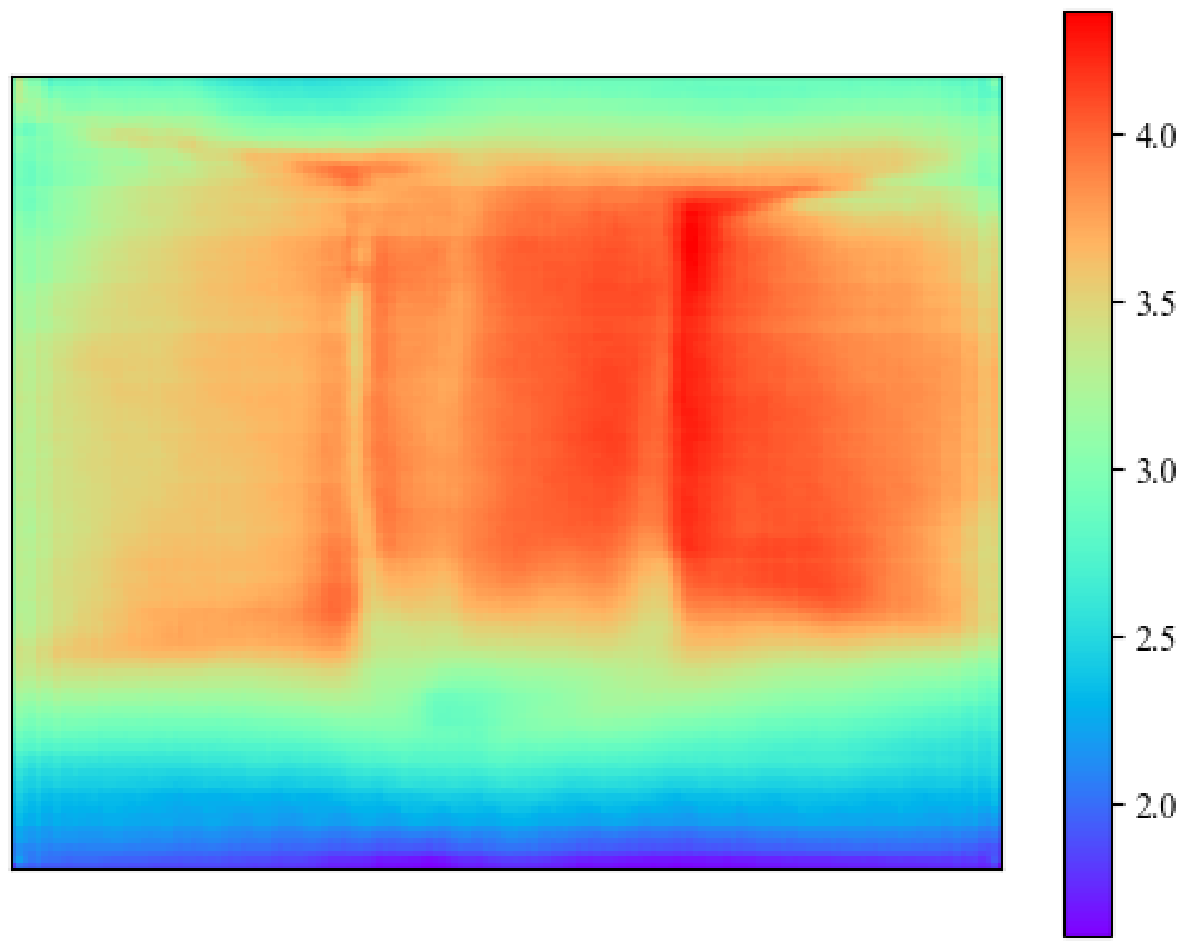}}
        & \vspace*{-18mm}~~~~~~96.2\% 
          \newline \newline $\left(\begin{array}{r}5573.9 \\  \downarrow~~~ \\ 209.5 \end{array}\right)$
        \\[2pt] \hline 

        $S^{(in)}_2$
        & {\includegraphics[width=26mm]{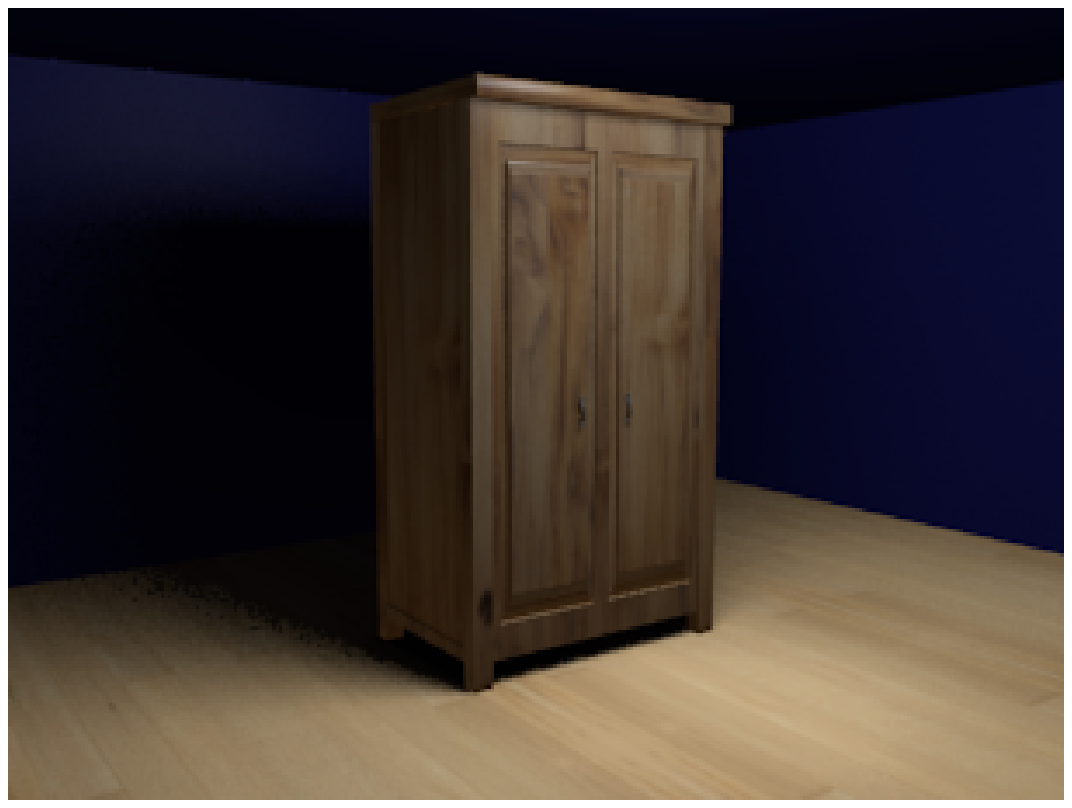}}
        & {\includegraphics[width=26mm]{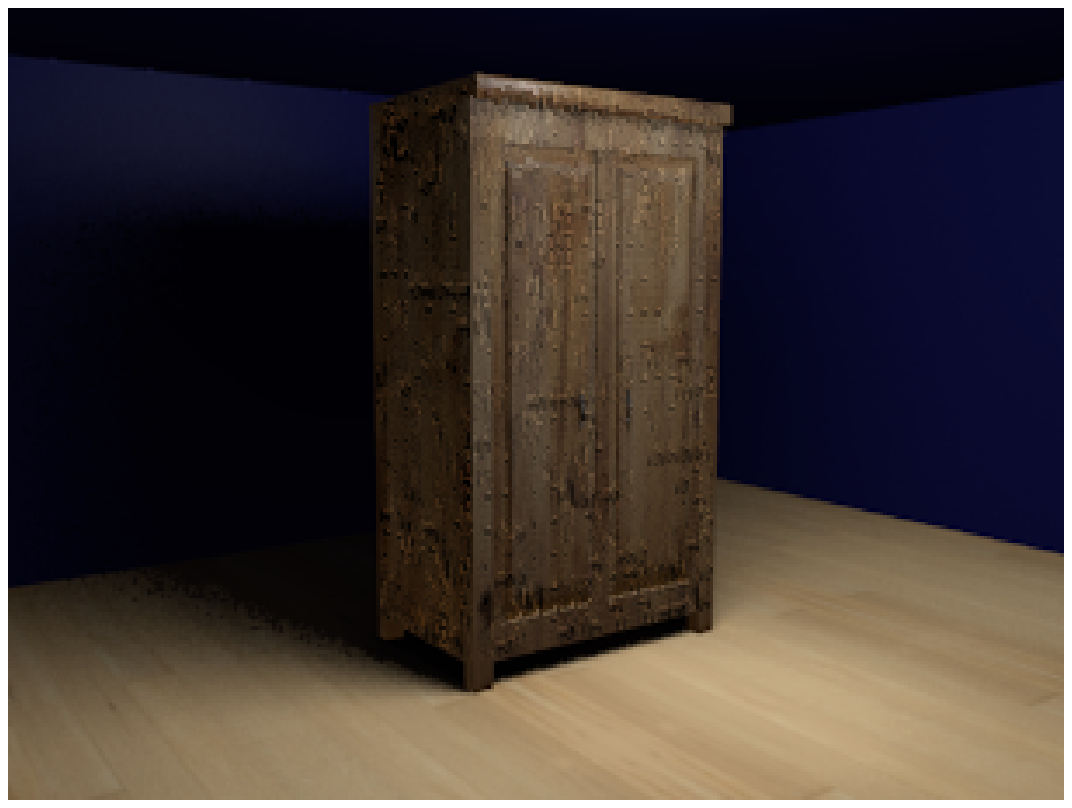}}
        & {\includegraphics[width=26mm]{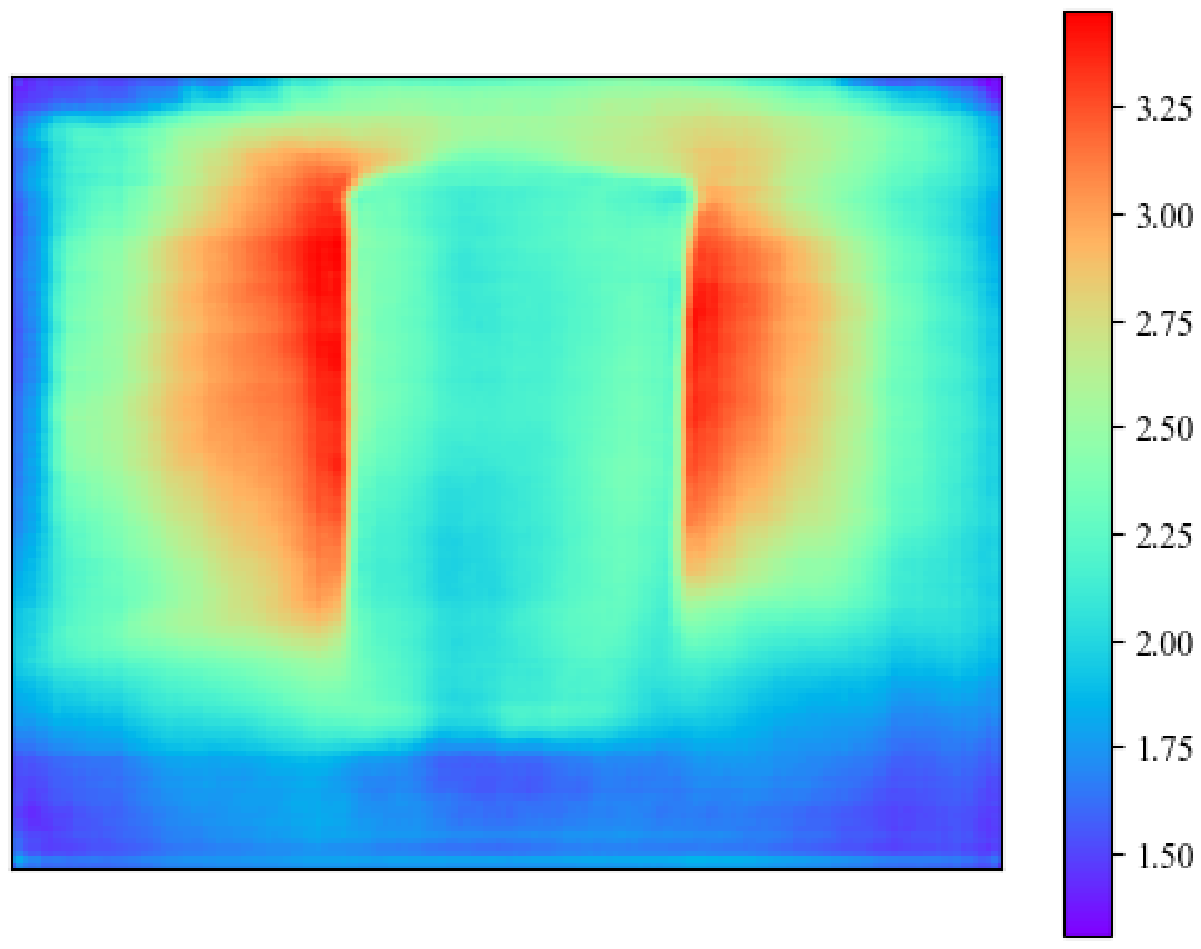}}
        & {\includegraphics[width=26mm]{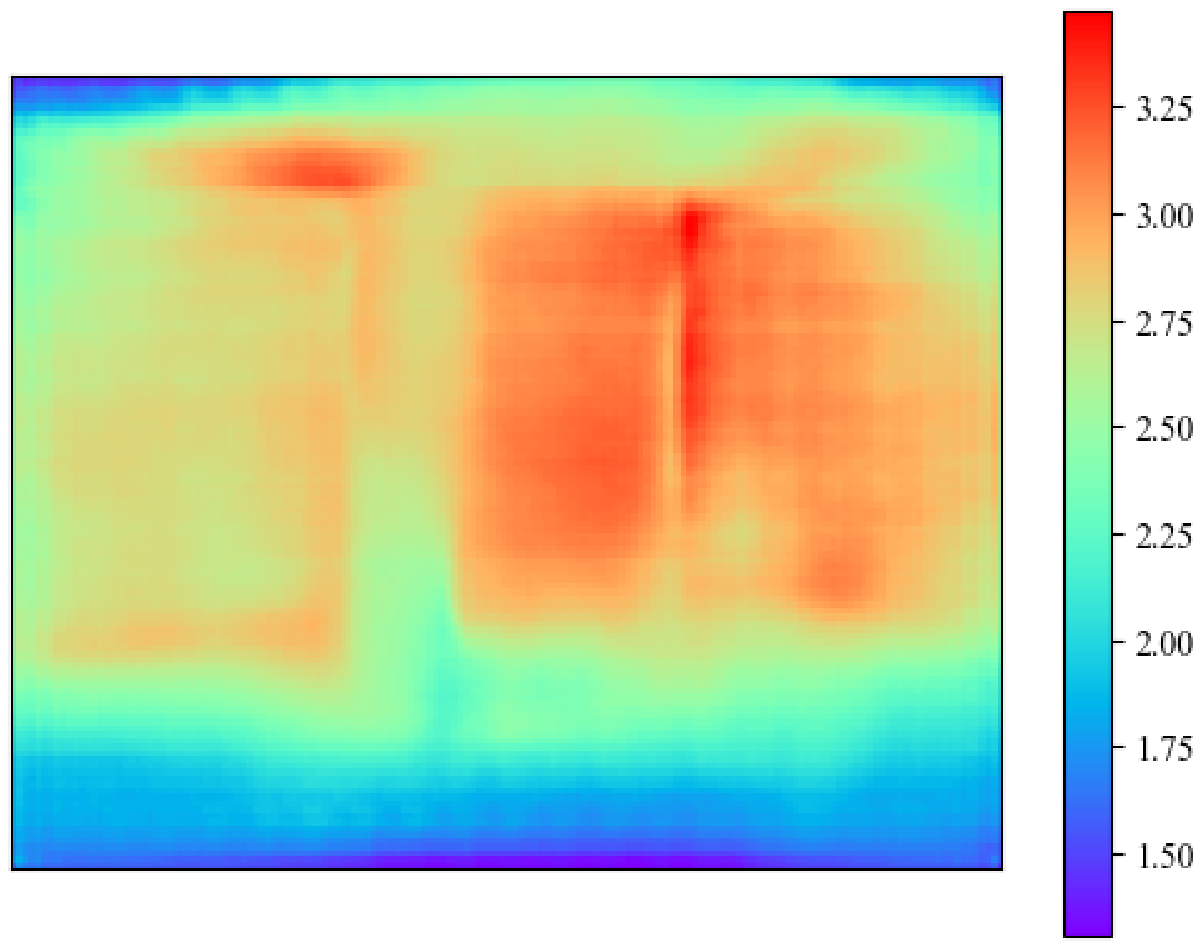}}
        & \vspace*{-18mm}~~~~~~91.9\% 
          \newline \newline $\left(\begin{array}{r}4275.6 \\  \downarrow~~~ \\ 346.3 \end{array}\right)$
        \\[3pt] \hline 

        $S^{(in)}_3$
        & {\includegraphics[width=26mm]{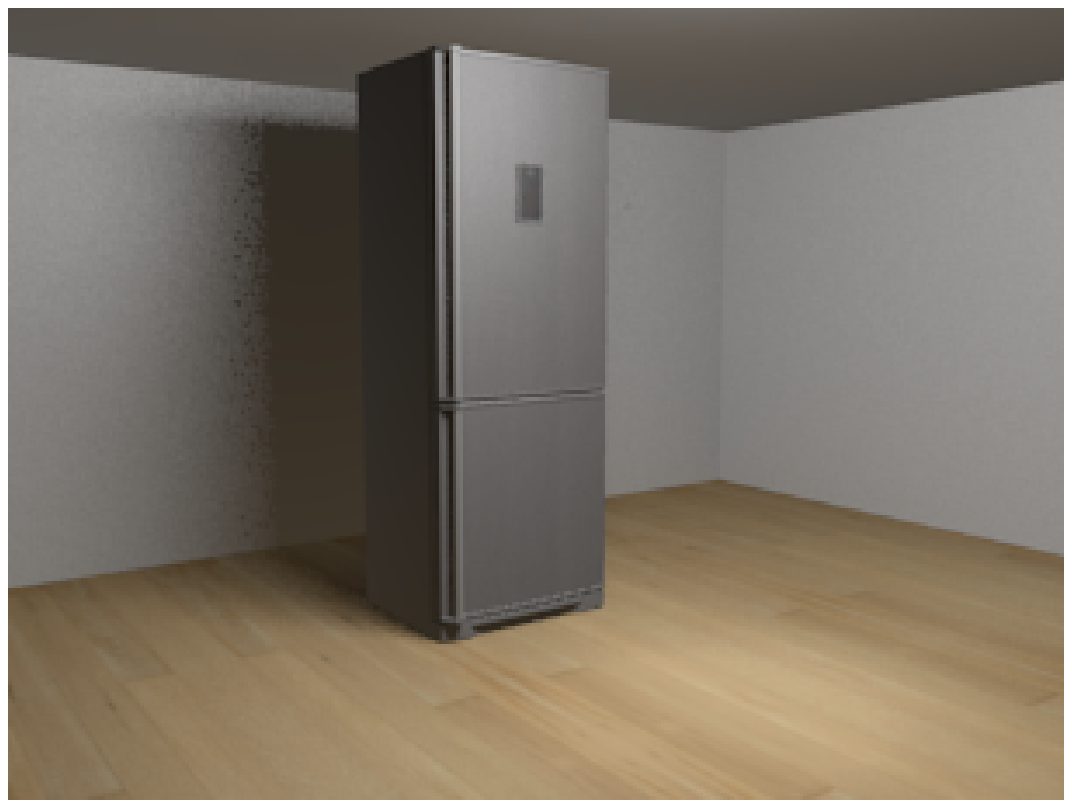}}
        & {\includegraphics[width=26mm]{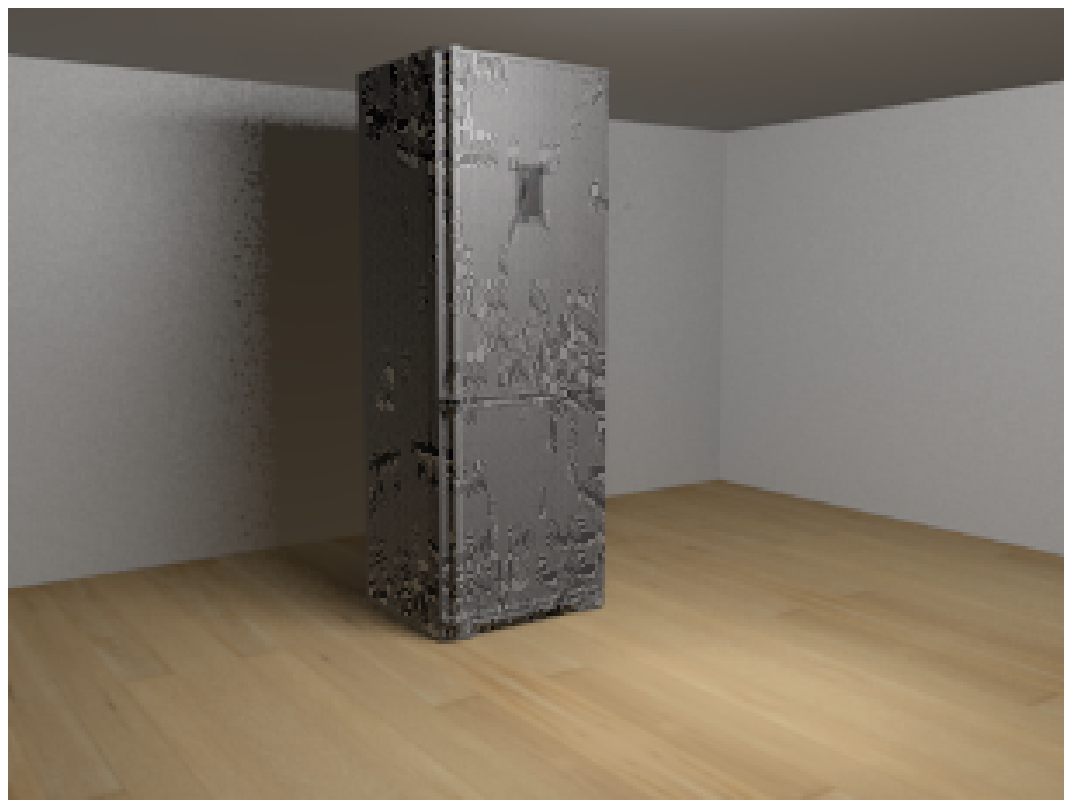}}
        & {\includegraphics[width=26mm]{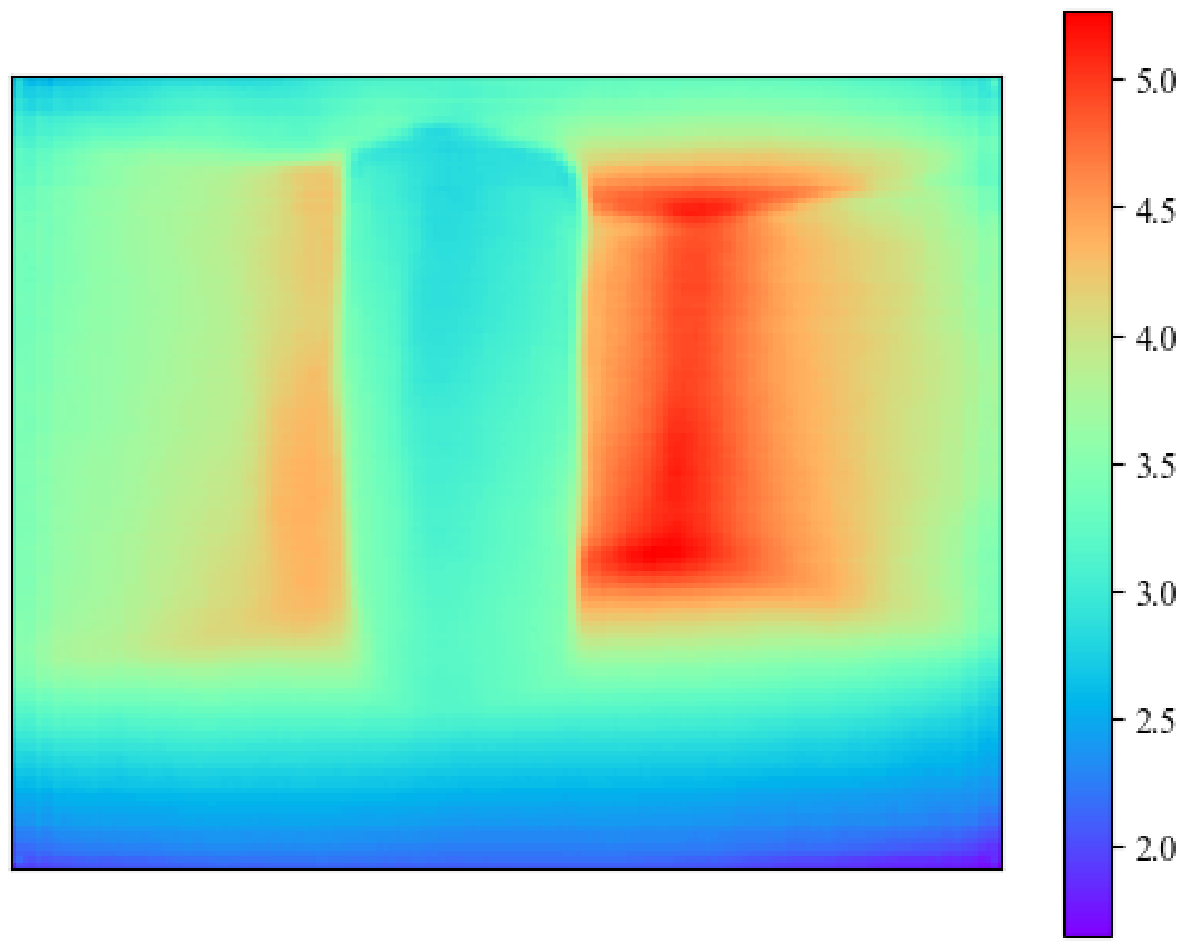}}
        & {\includegraphics[width=26mm]{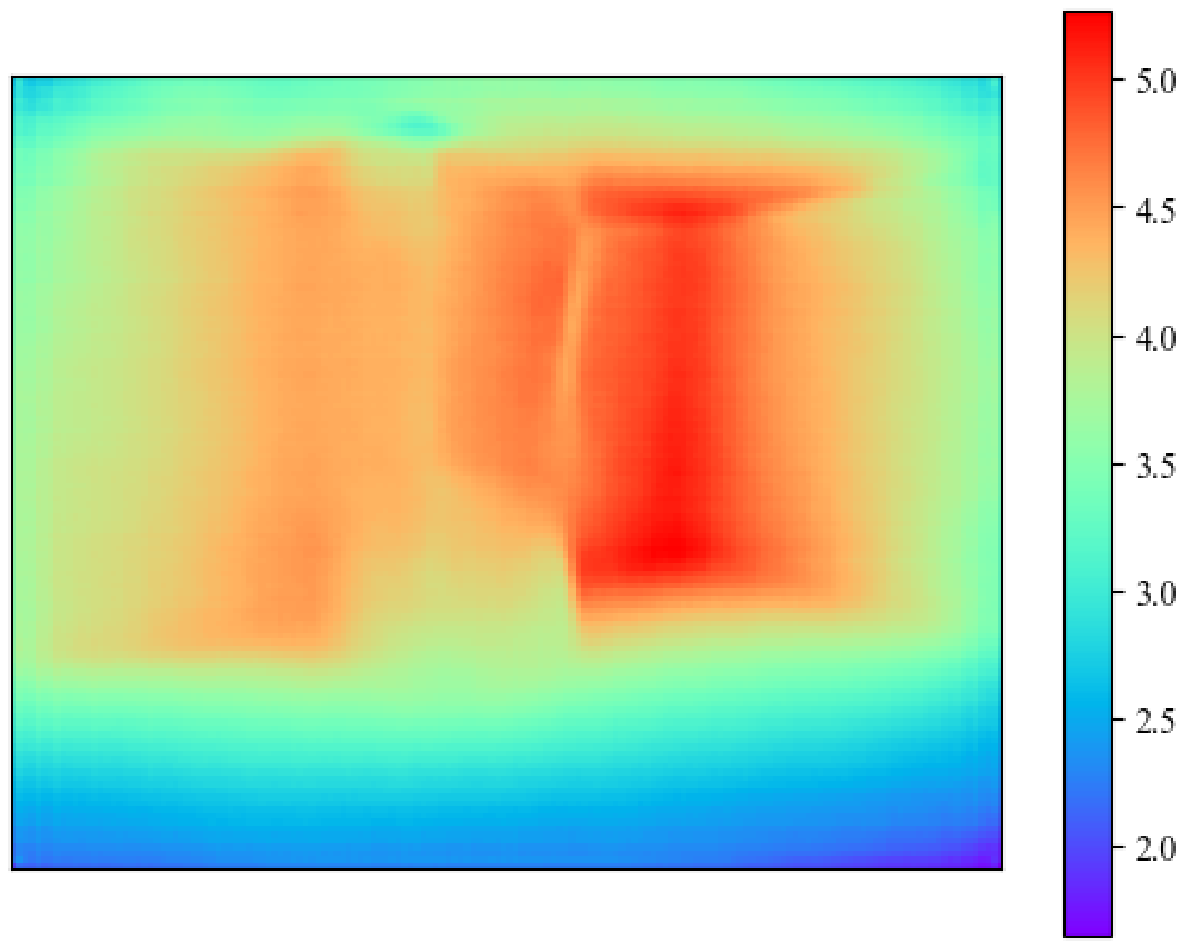}}
        & \vspace*{-18mm}~~~~~~91.8\% 
          \newline \newline $\left(\begin{array}{r}4843.3 \\  \downarrow~~~ \\ 395.6 \end{array}\right)$
        \\ \hline 

        $S^{(in)}_4$
        & {\includegraphics[width=26mm]{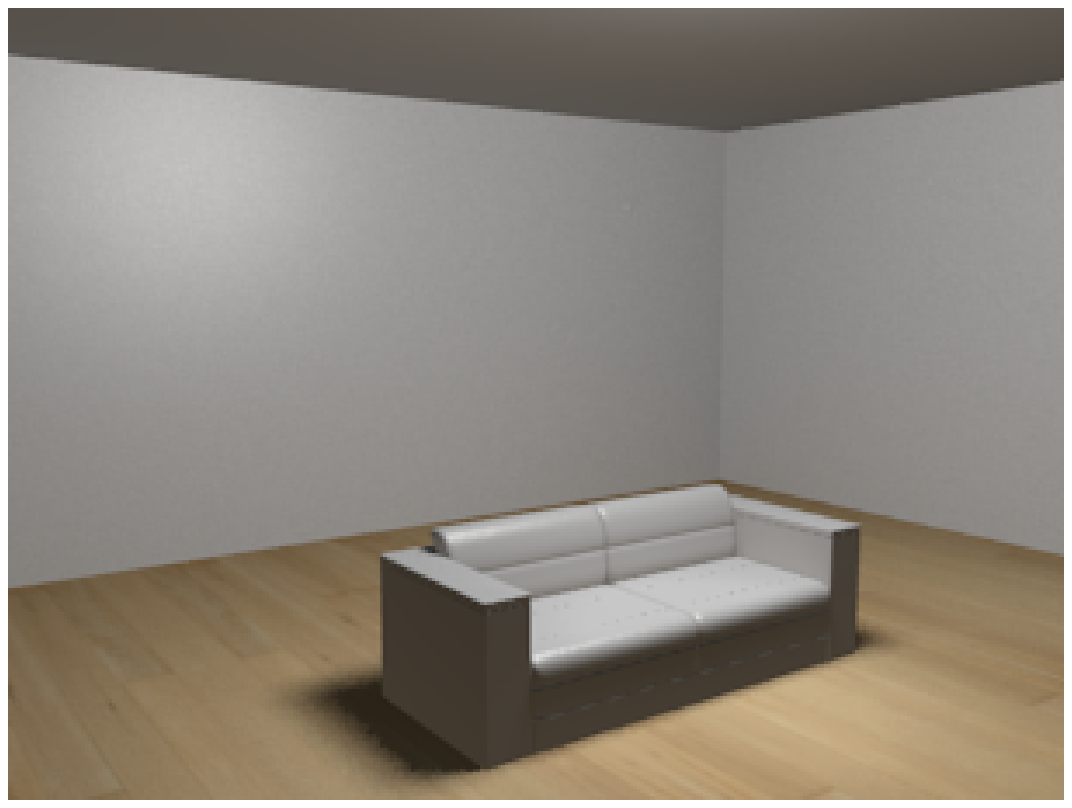}}
        & {\includegraphics[width=26mm]{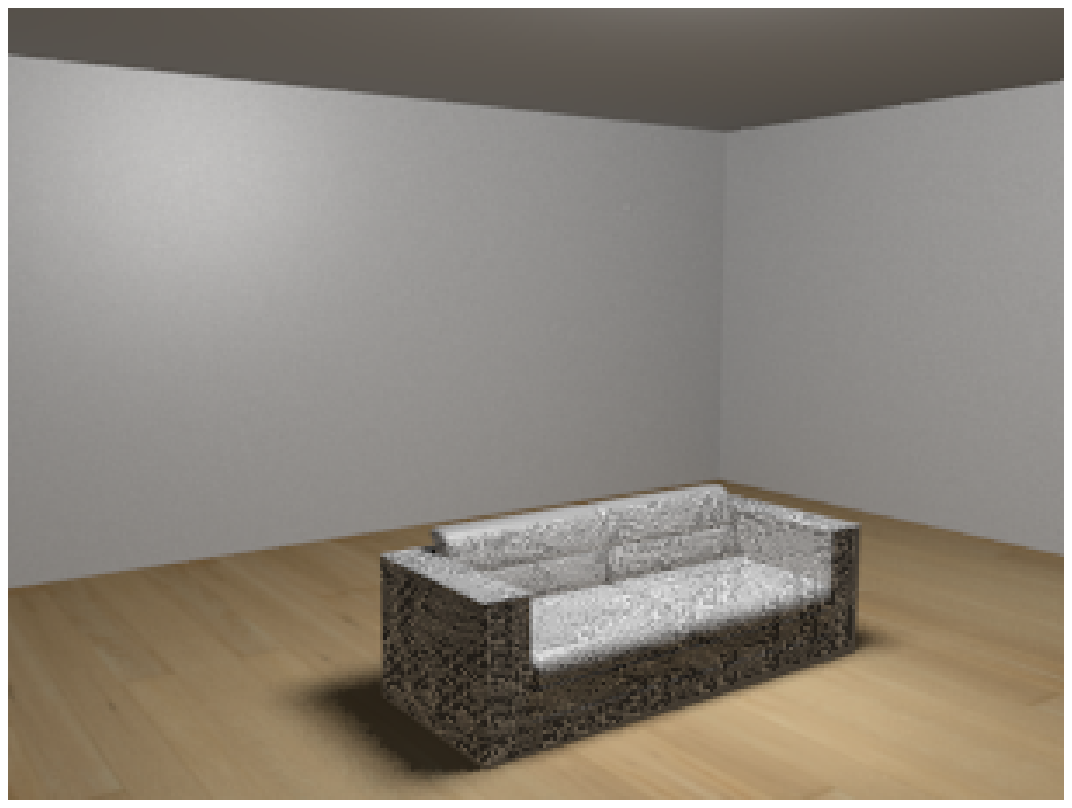}}
        & {\includegraphics[width=26mm]{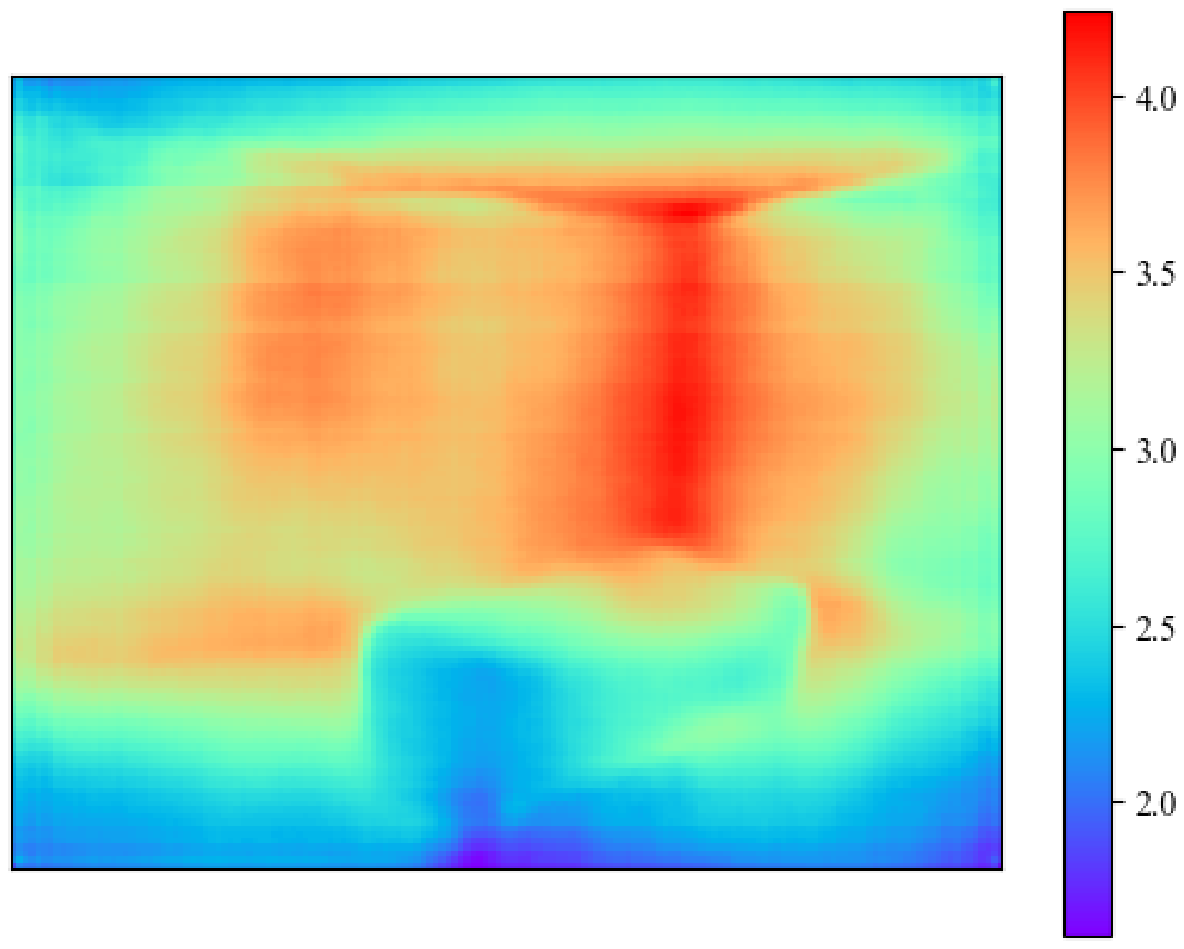}}
        & {\includegraphics[width=26mm]{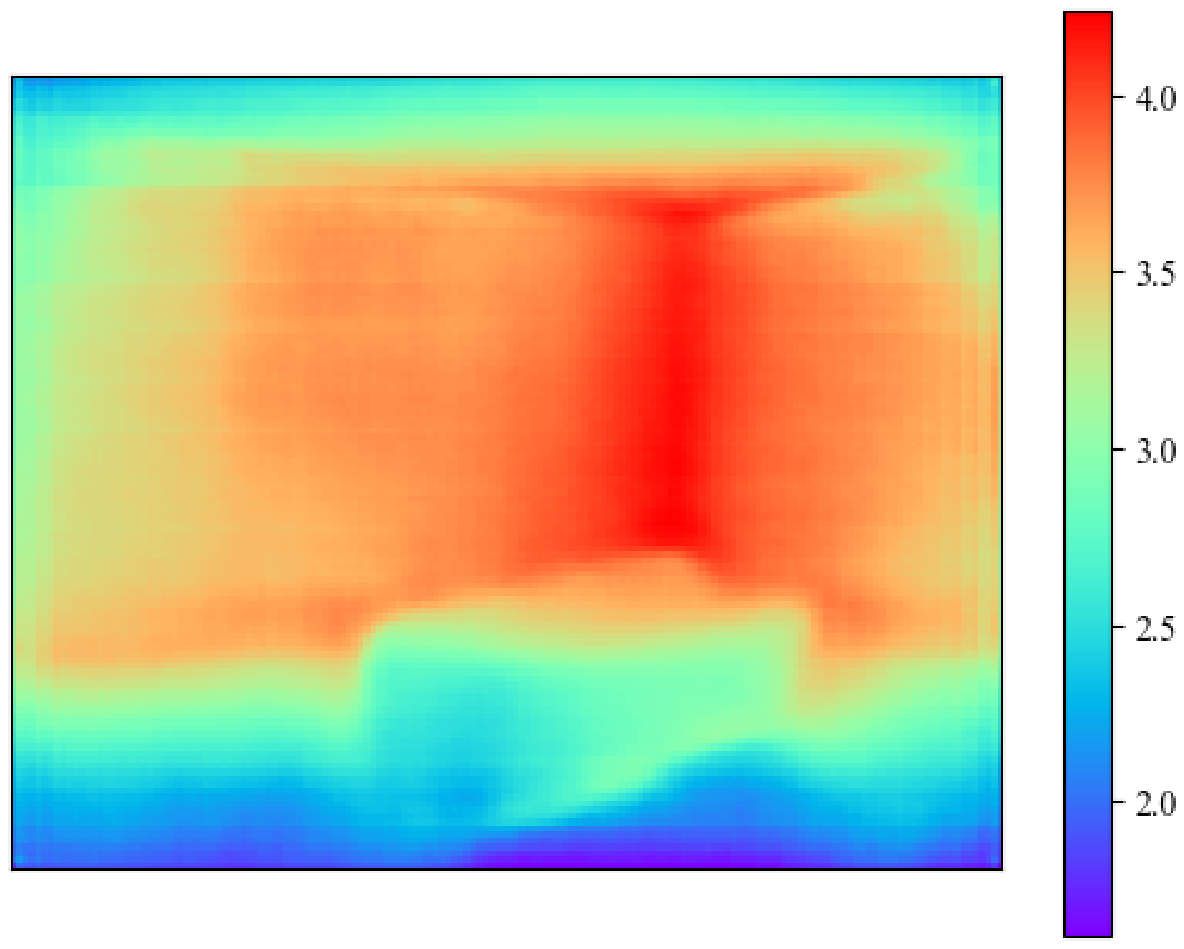}}
        & \vspace*{-18mm}~~~~~~82.2\% 
          \newline \newline $\left(\begin{array}{r}1202.8 \\  \downarrow~~~ \\ 214.0 \end{array}\right)$
        \\ \hline 

        $S^{(in)}_5$
        & {\includegraphics[width=26mm]{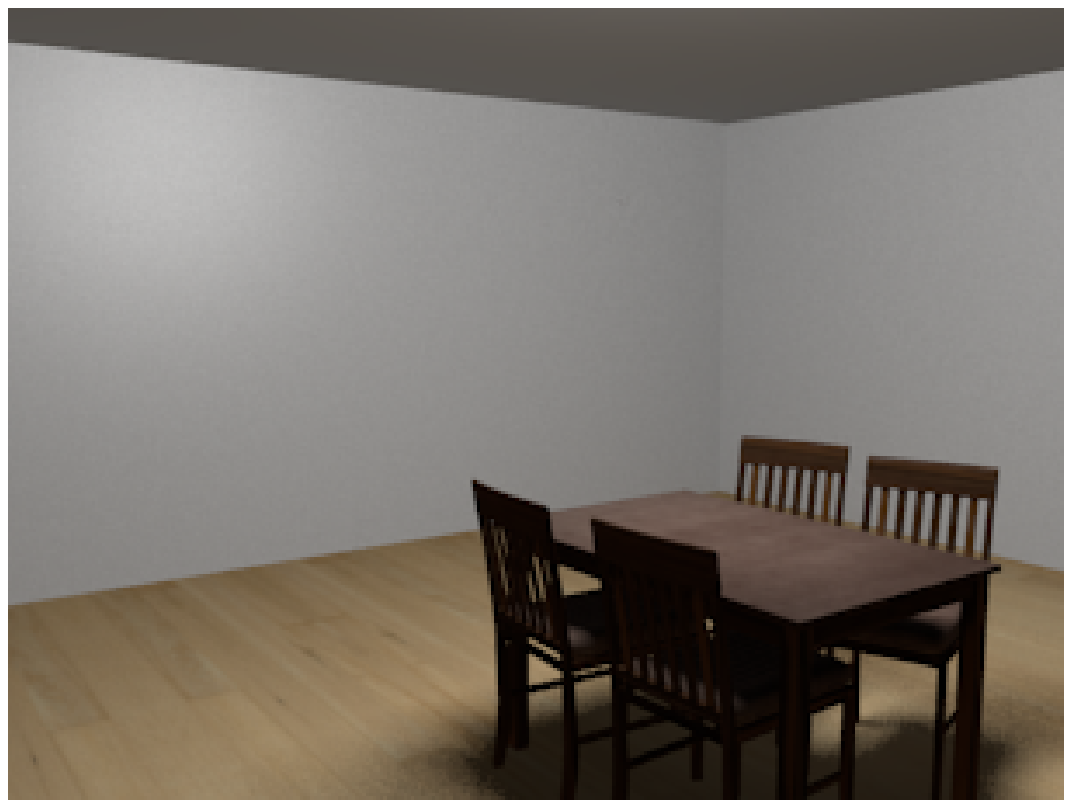}}
        & {\includegraphics[width=26mm]{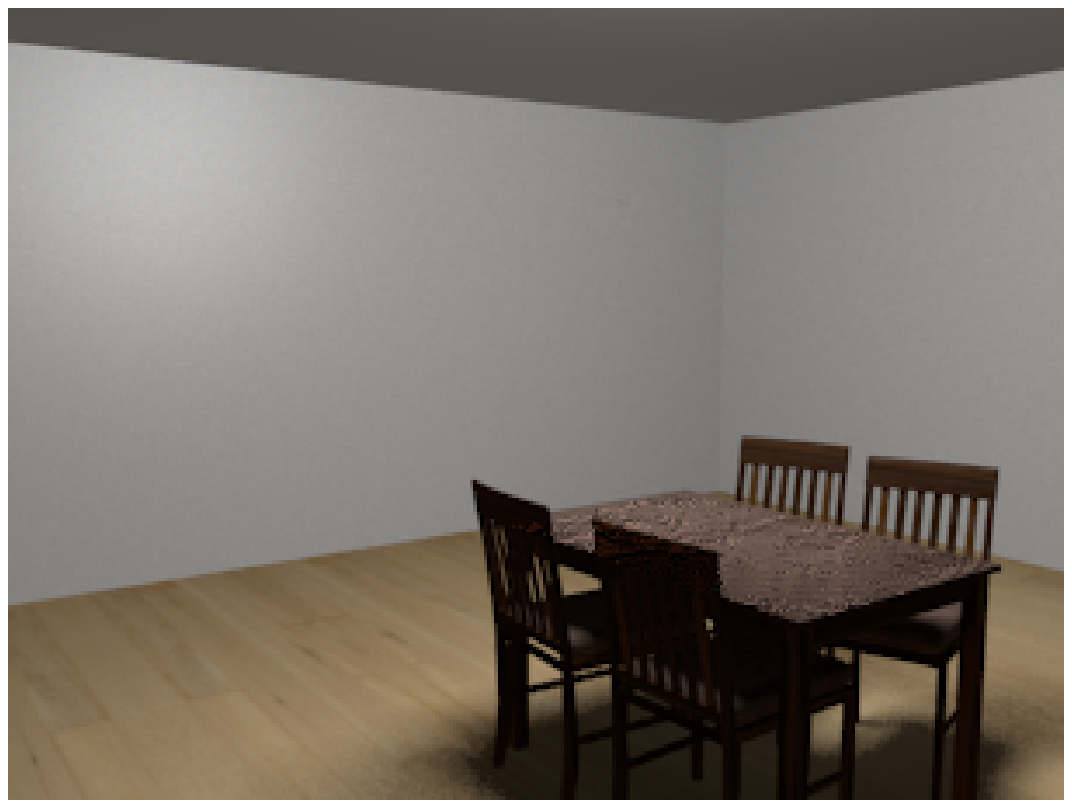}}
        & {\includegraphics[width=26mm]{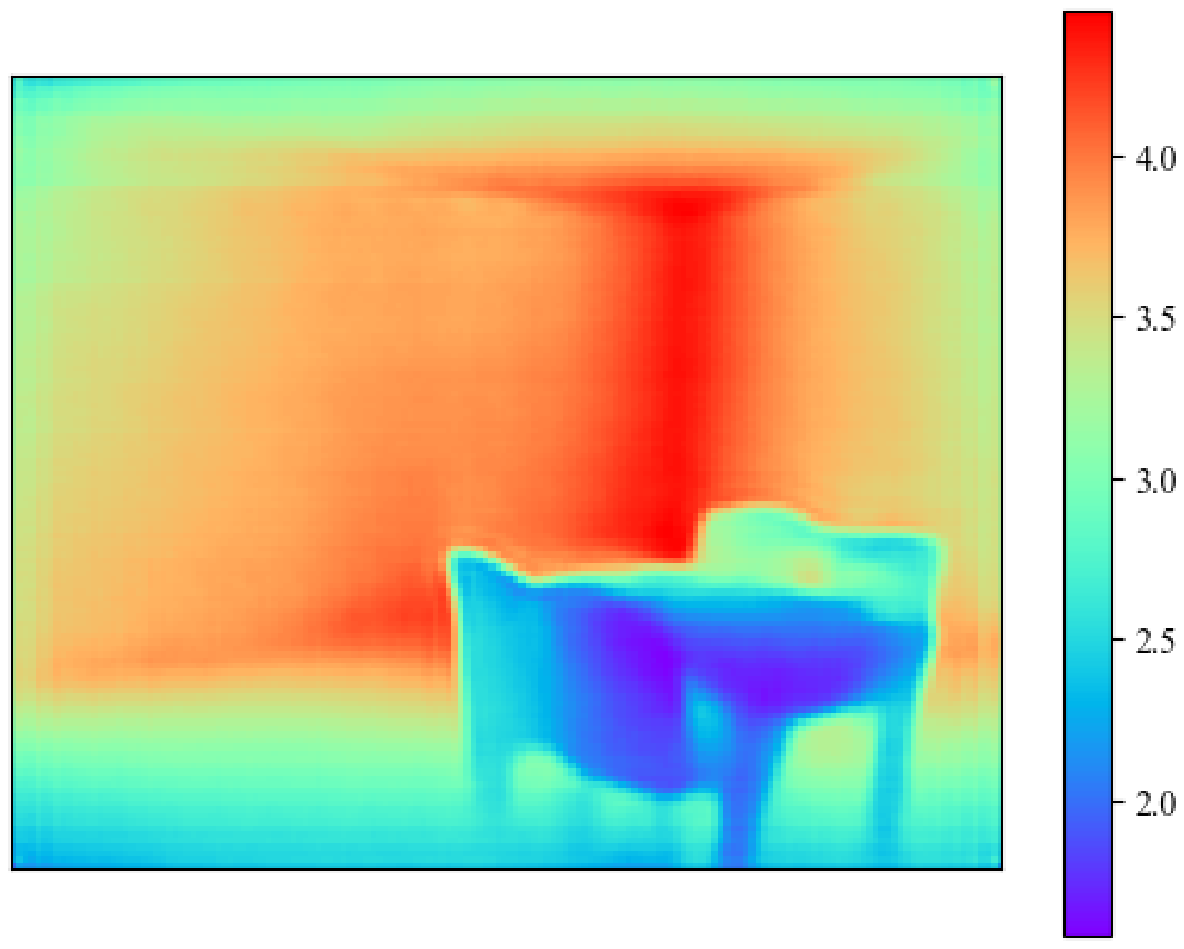}}
        & {\includegraphics[width=26mm]{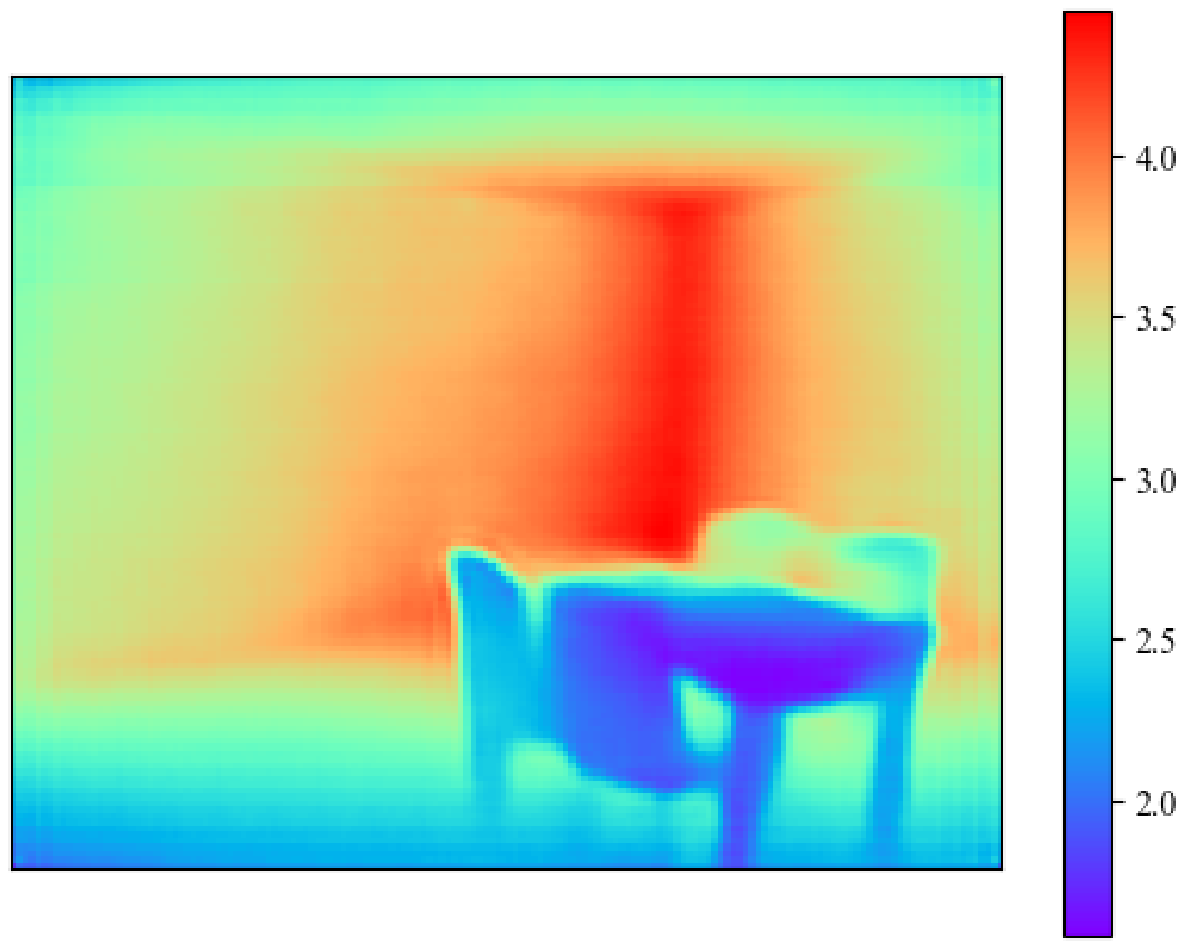}}
        & \vspace*{-18mm}~~~~~~24.7\% 
          \newline \newline $\left(\begin{array}{r}3220.9 \\  \downarrow~~~ \\ 2424.1 \end{array}\right)$
        \\ \hline 
        \multicolumn{6}{c}{~}\\
      \end{tabular}}
      \figcaption{Effect of target object shape (Experiment 1-a)}
      \label{tab:target}
\end{figure*}

From the results shown in Fig.~\ref{tab:target},
in scenes $S^{(in)}_1$ to $S^{(in)}_3$,
the proposed adversarial examples succeeded in having the target DNN
make false depth estimations as if the target object disappeared.
In $S^{(in)}_4$, the sofa seems to have moved backward as well
although it is difficult to see at a first glance.
However, it was difficult to fool the DNN in scene $S^{(in)}_5$ by
adding the perturbation only on the top size of the table set, which
contains many rod-like shapes at the bottom.

  \subsubsection{Experiment 1-b: Influence of the presence other object than the target
  }~\\
Next, we investigated the influence of the presence of objects other
than the target object in the scene.
We created scenes
 $S^{(in)}_6$, $S^{(in)}_7$, $S^{(in)}_8$ by
putting a dining table set on scene 
$S^{(in)}_{1}$, $S^{(in)}_{3}$, $S^{(in)}_{4}$, respectively.

\begin{figure*}[t]
      \centering
      {\footnotesize
      \begin{tabular}{@{}l@{~}|@{~~~~}c@{~}c@{~}|@{~~~}c@{~}c@{~}|p{18mm}@{}} \hline
         Scene & \multicolumn{2}{c@{~}|@{~~~}}{Images} & \multicolumn{2}{c@{~}|}{Depth maps} & Pseudo volume\\
               &Original & Adversarial ex. & Original & Adversarial ex.     & reduction rate\\
        \hline

        $S^{(in)}_{6}$
        & {\includegraphics[width=26mm]{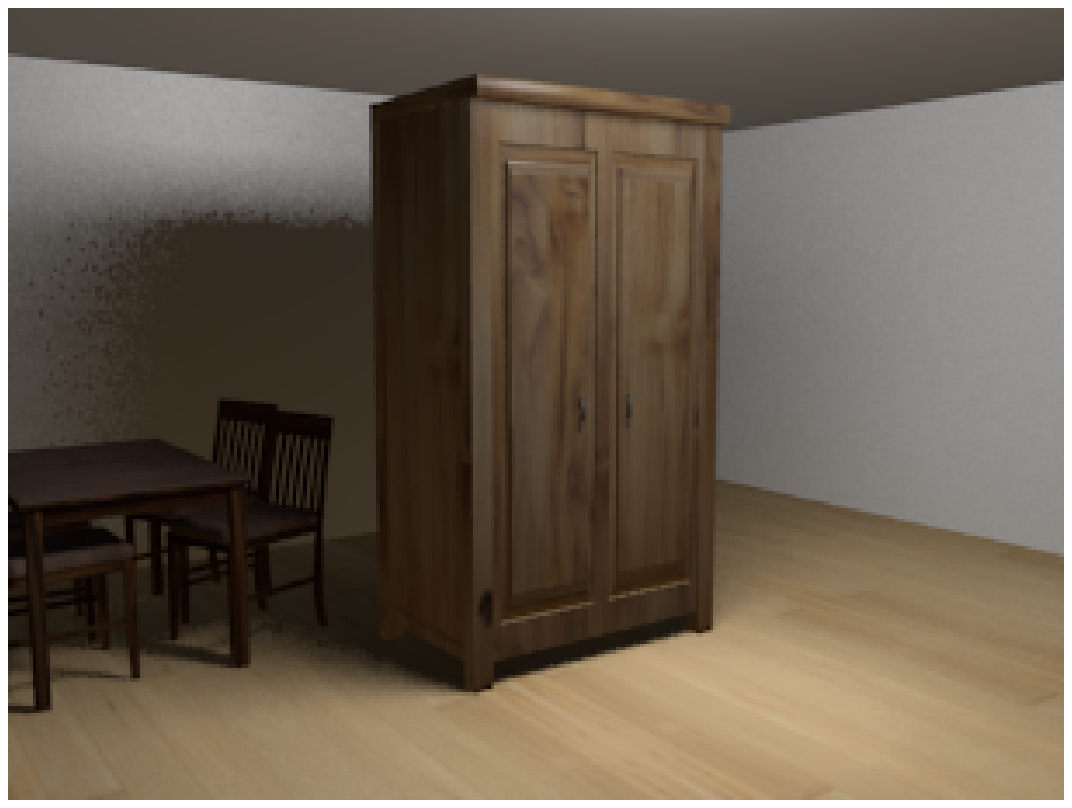}}
        & {\includegraphics[width=26mm]{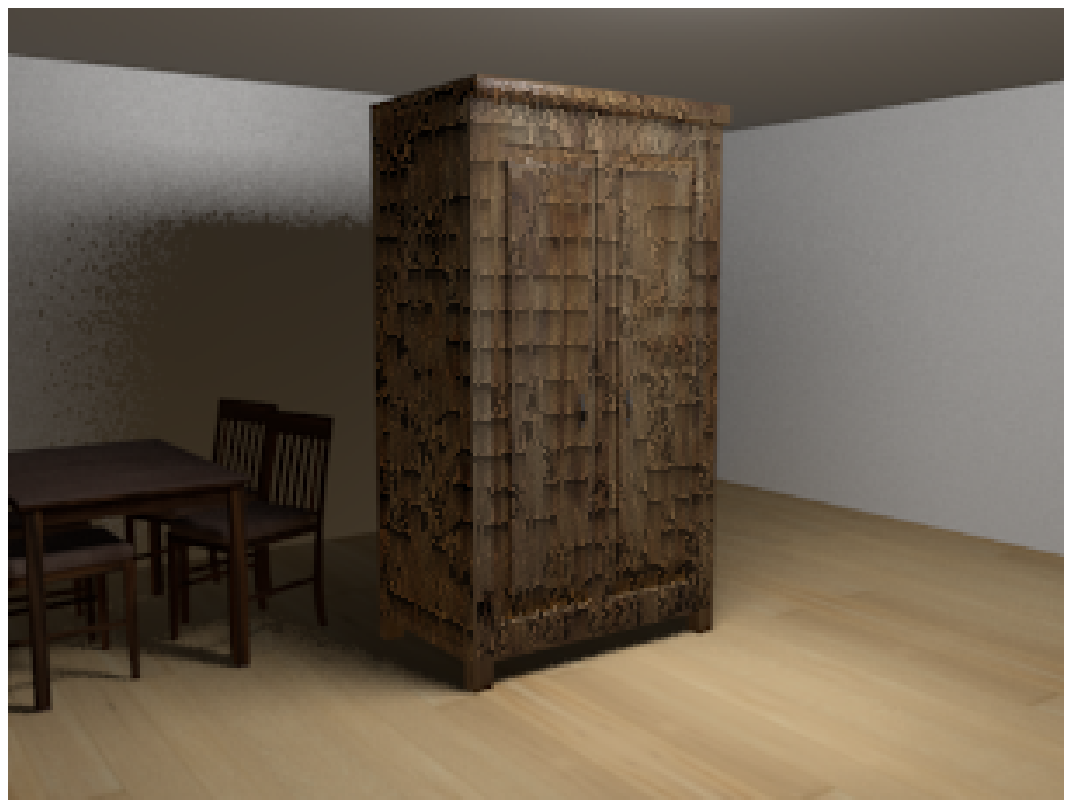}}
        & {\includegraphics[width=26mm]{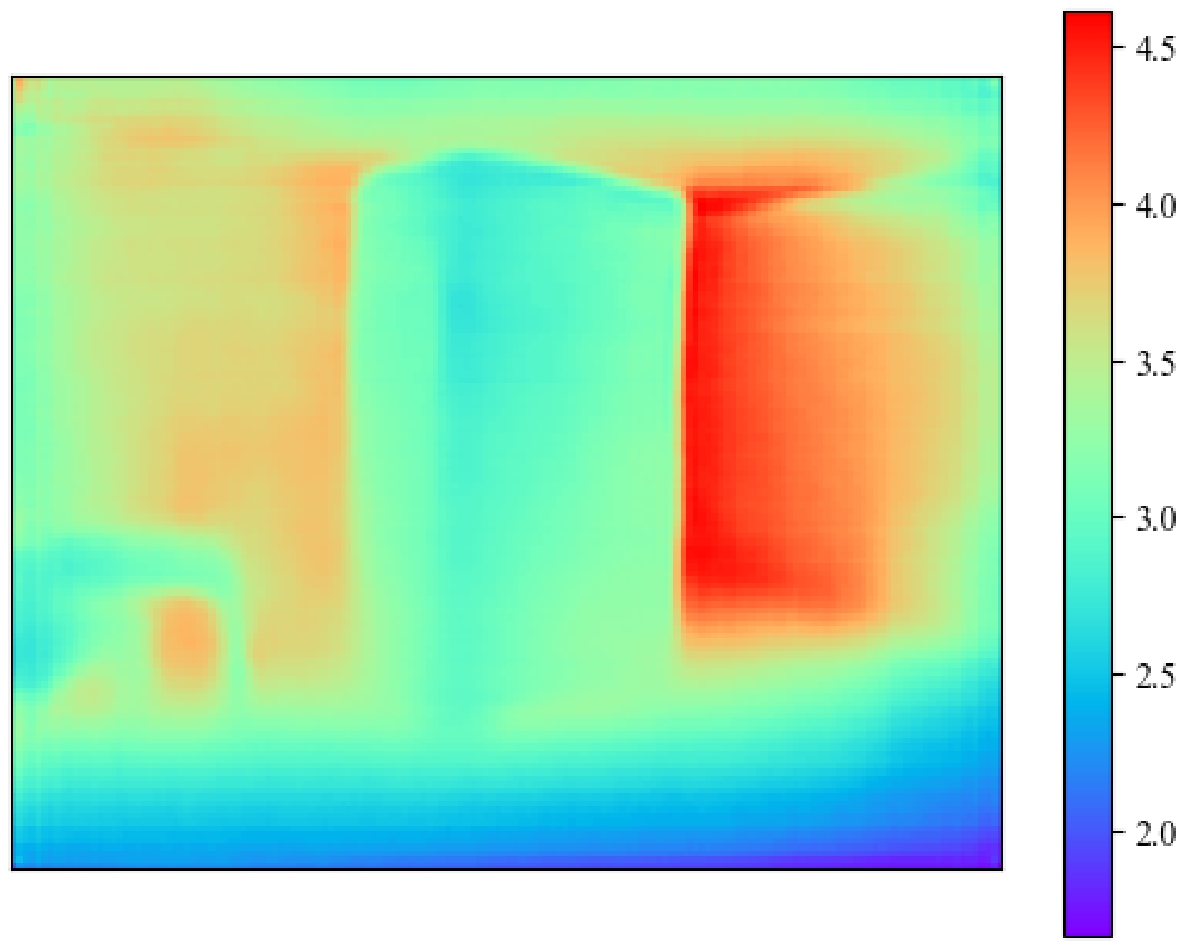}}
        & {\includegraphics[width=26mm]{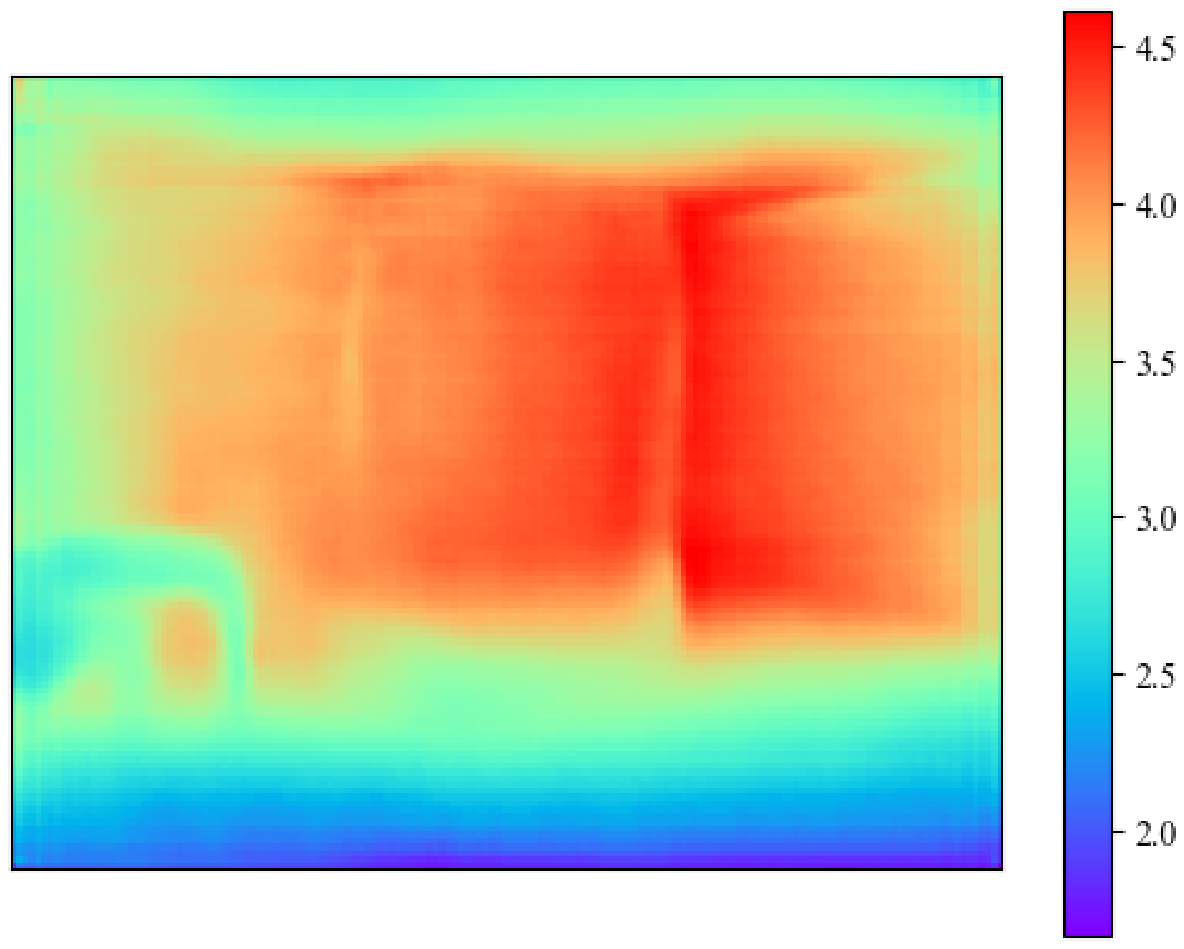}}
        & \vspace*{-18mm}~~~~~~98.2\% 
          \newline \newline $\left(\begin{array}{r}5075.3 \\  \downarrow~~~ \\ 92.3 \end{array}\right)$
        \\ \hline 

        $S^{(in)}_{7}$
        & {\includegraphics[width=26mm]{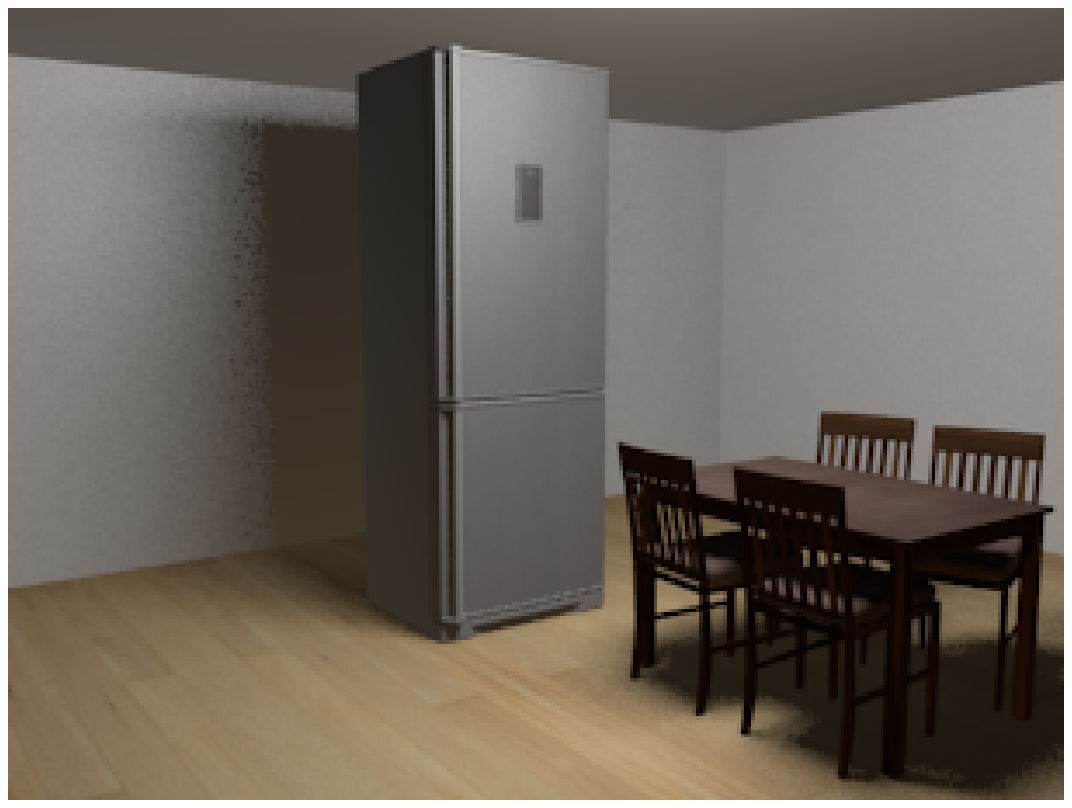}}
        & {\includegraphics[width=26mm]{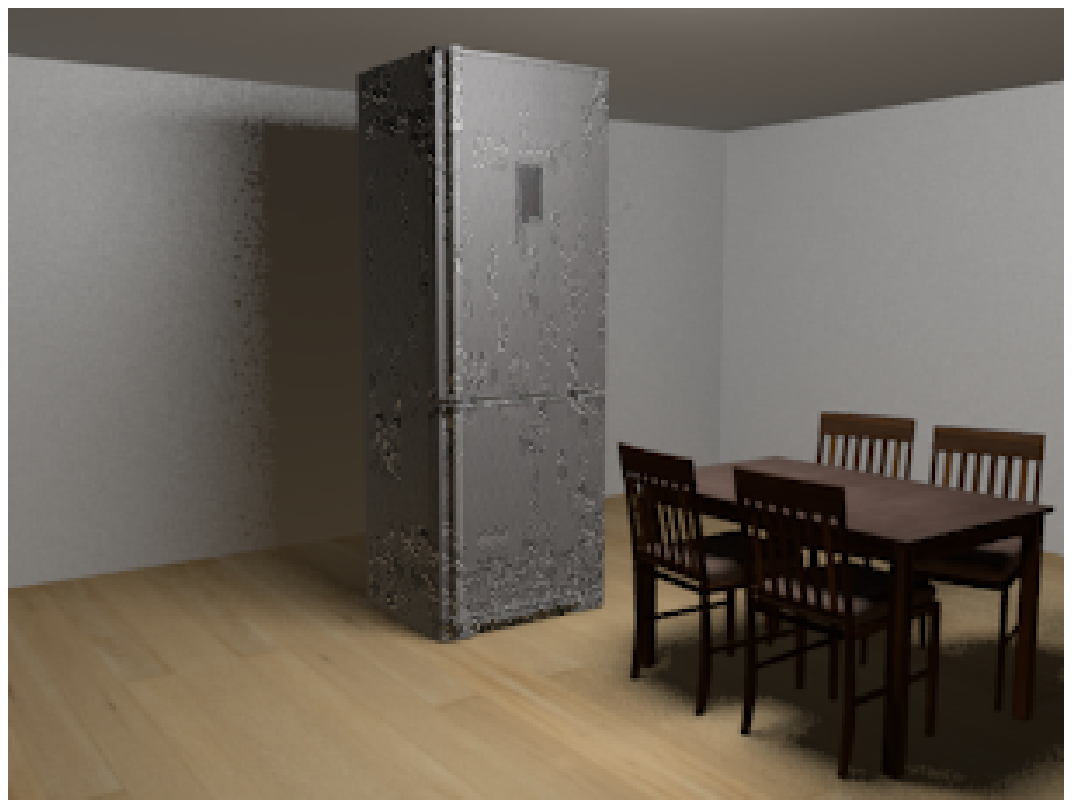}}
        & {\includegraphics[width=26mm]{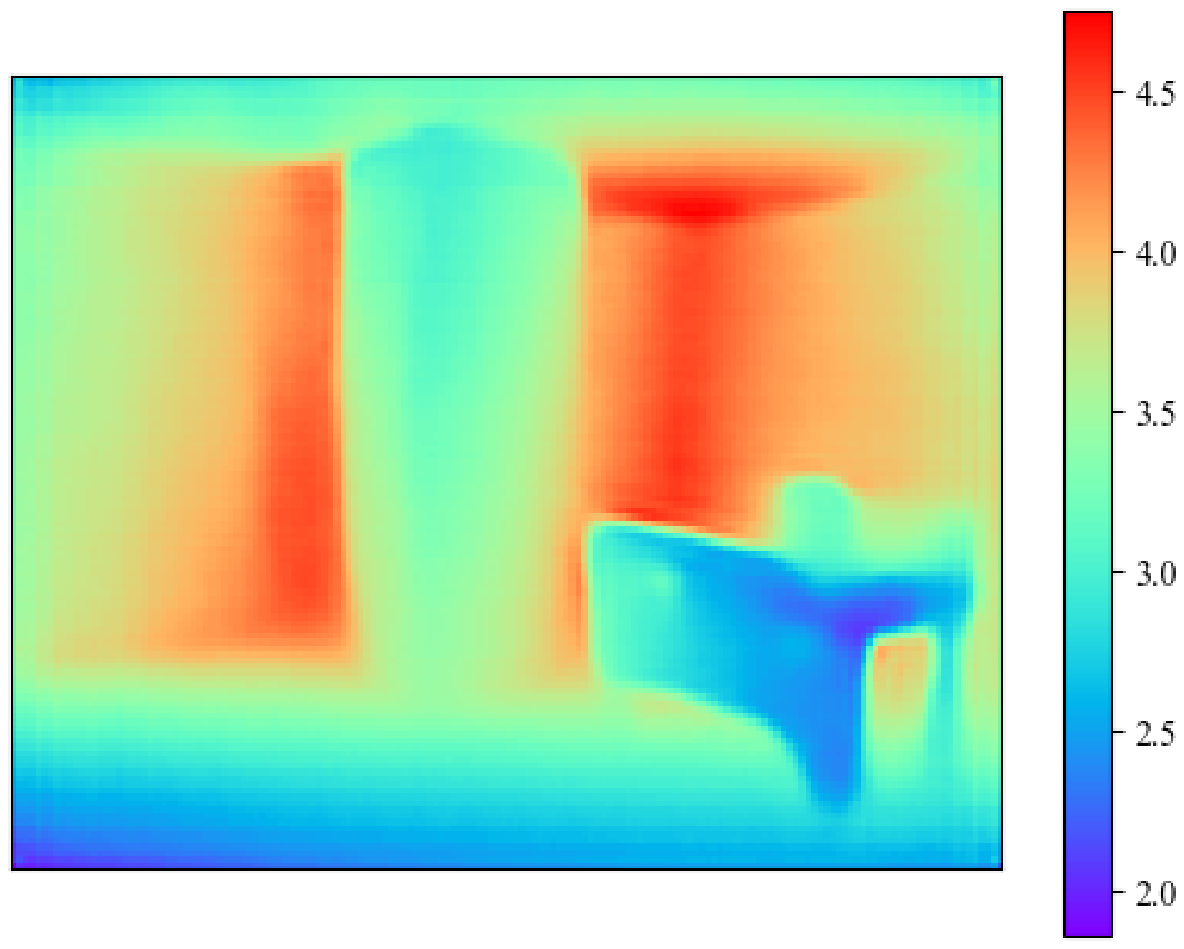}}
        & {\includegraphics[width=26mm]{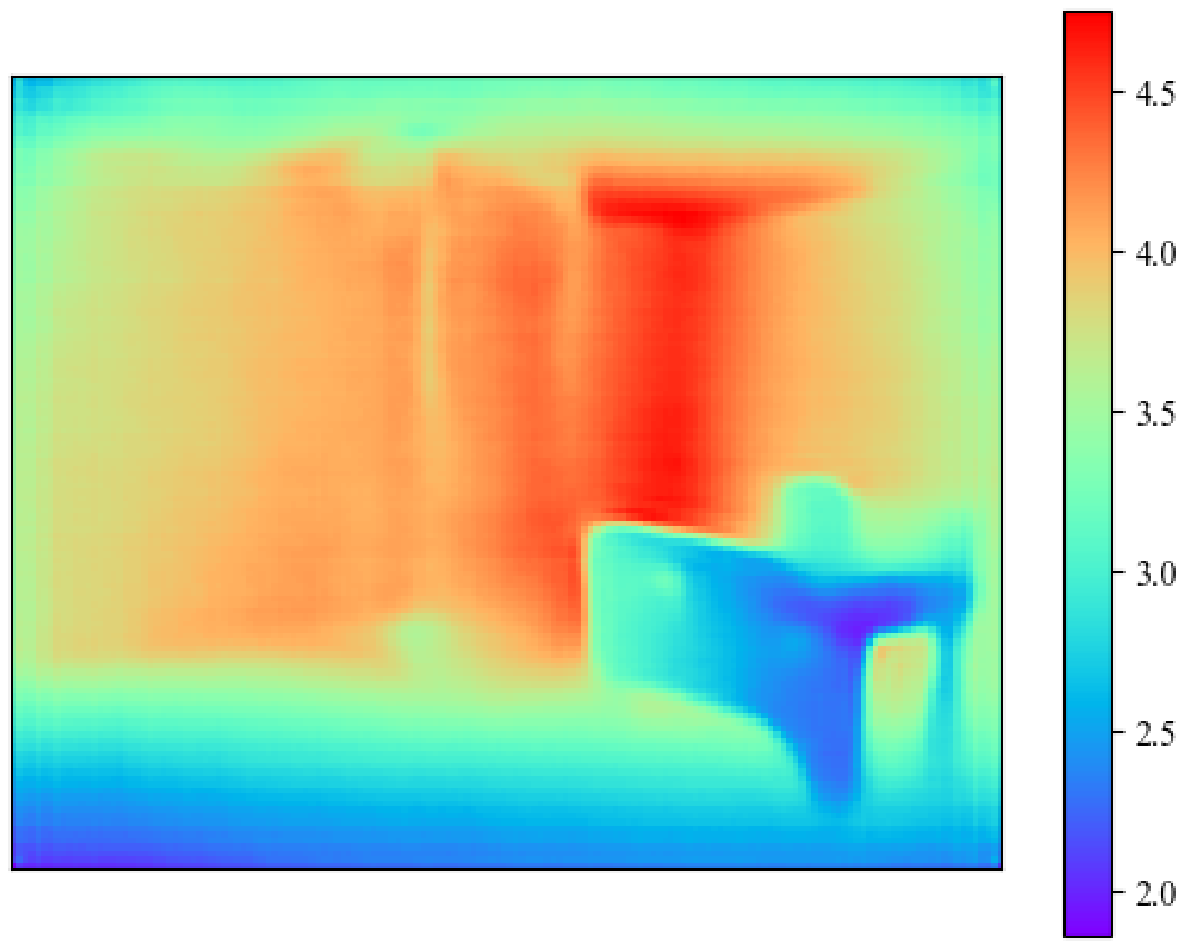}}
        & \vspace*{-18mm}~~~~~~97.8\% 
          \newline \newline $\left(\begin{array}{r}2834.2 \\  \downarrow~~~ \\ 62.8 \end{array}\right)$
        \\ \hline 

        $S^{(in)}_{8}$
        & {\includegraphics[width=26mm]{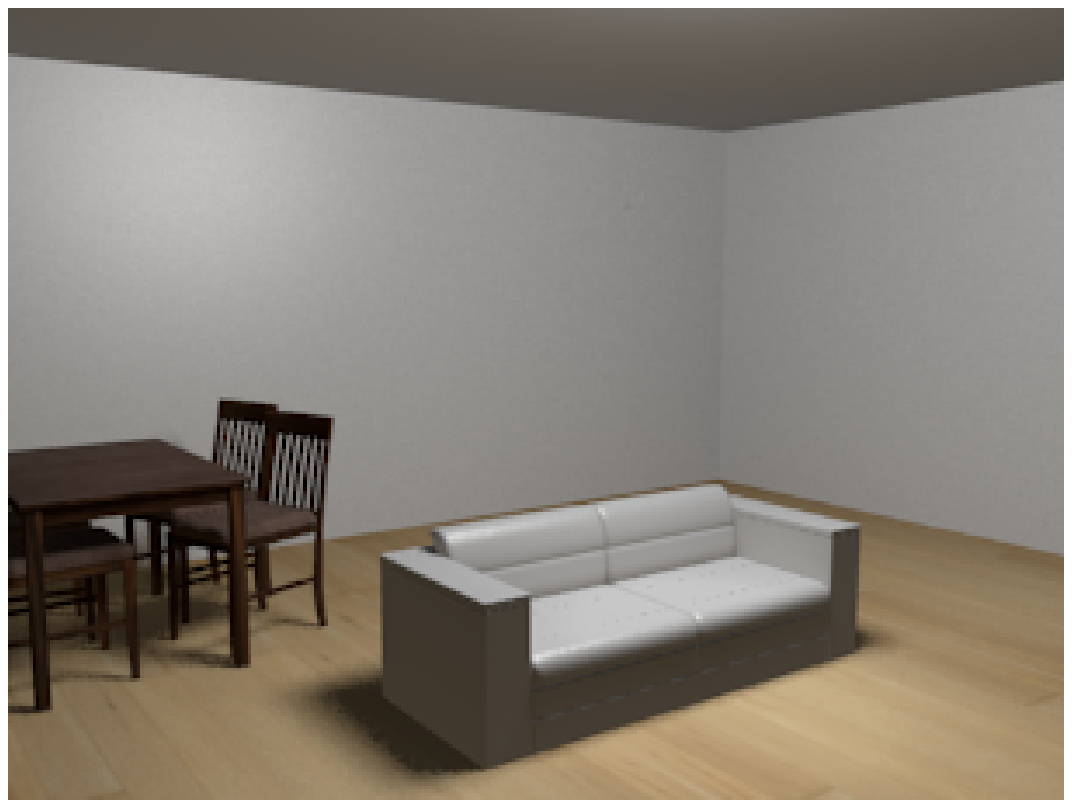}}
        & {\includegraphics[width=26mm]{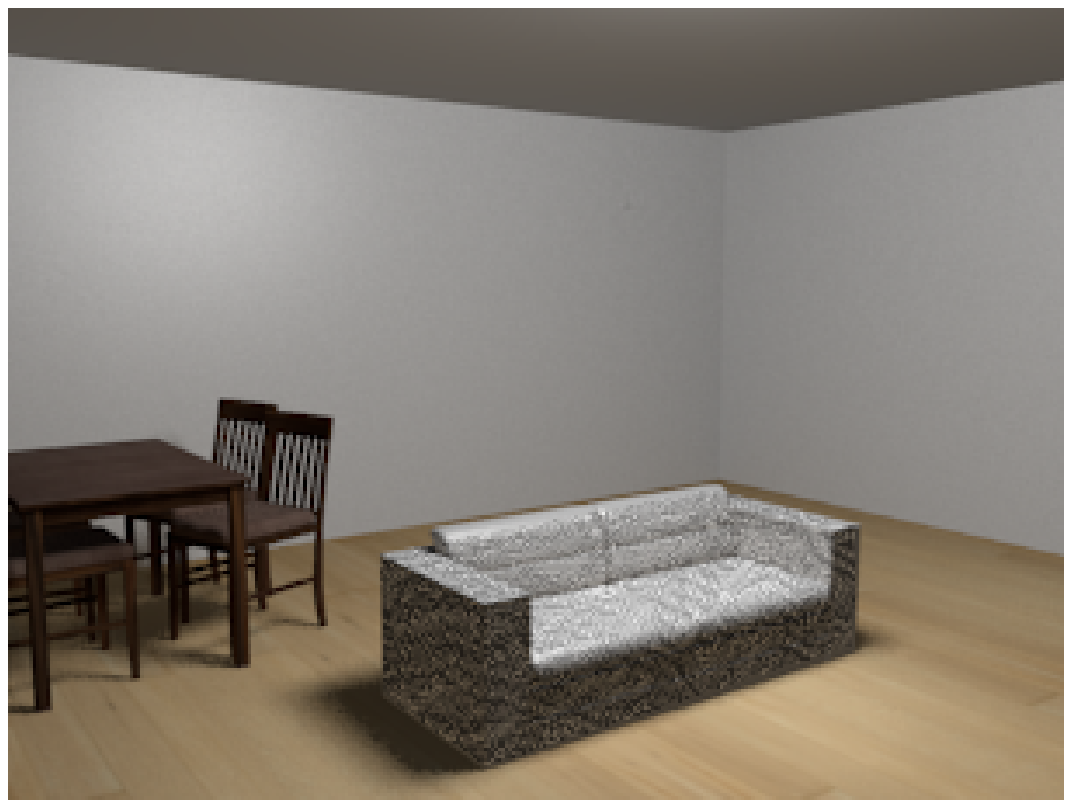}}
        & {\includegraphics[width=26mm]{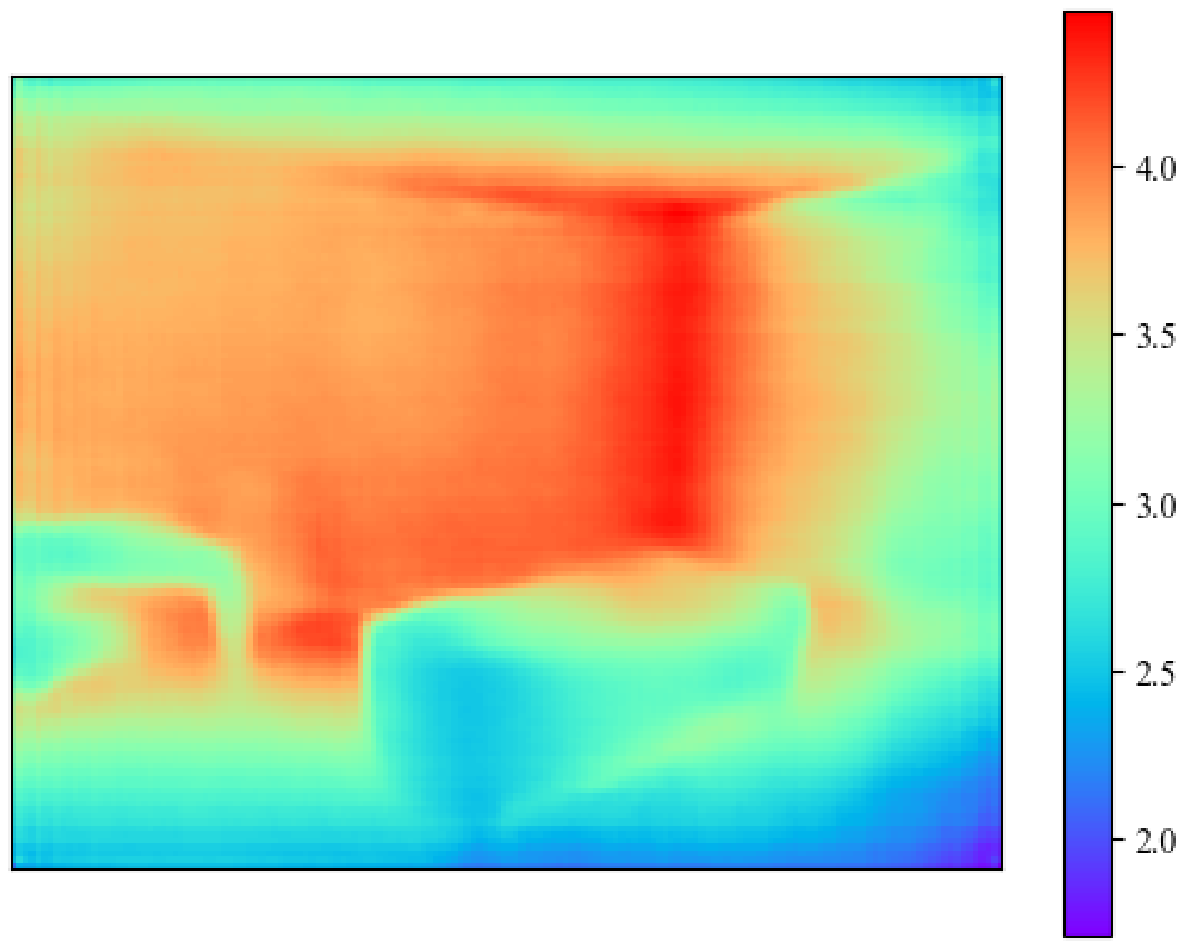}}
        & {\includegraphics[width=26mm]{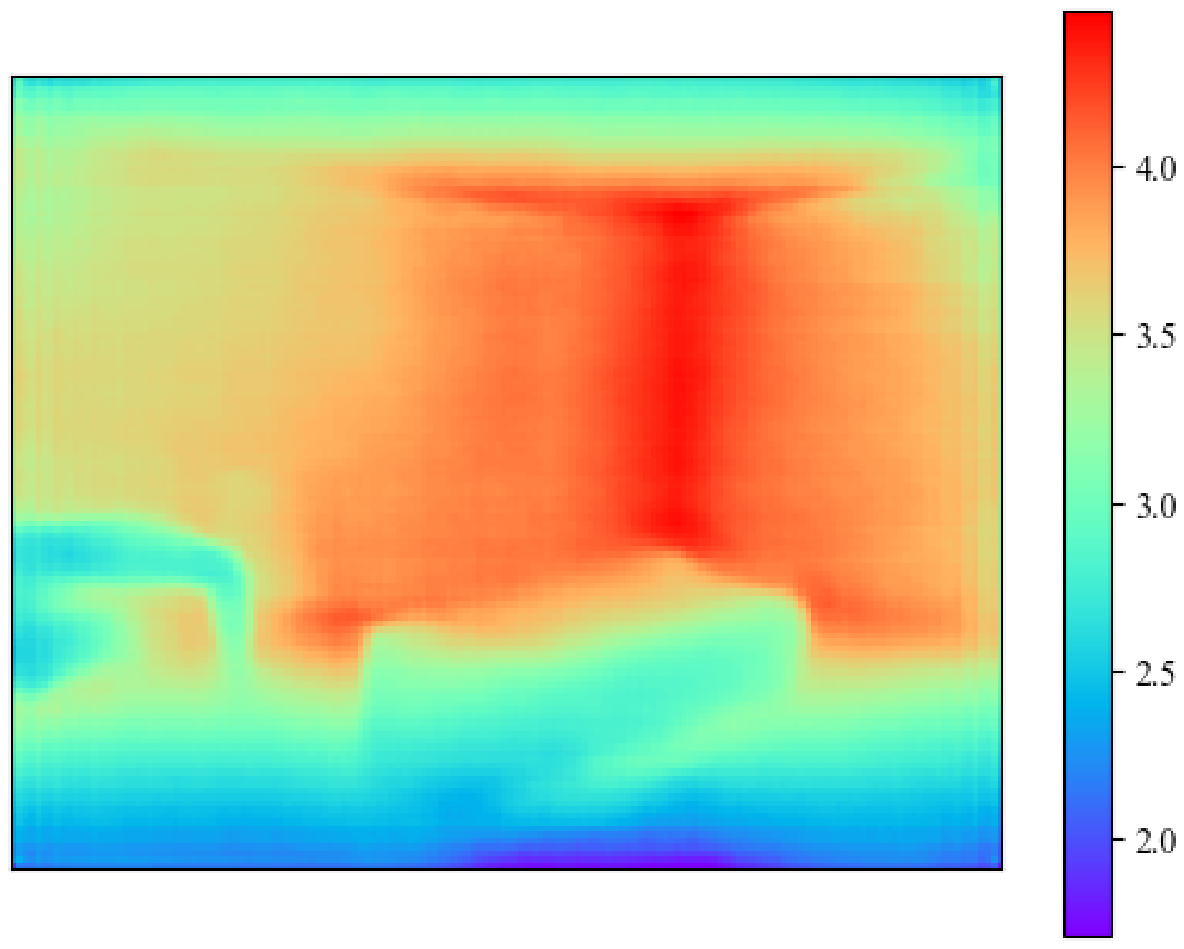}}
        & \vspace*{-18mm}~~~~~~75.4\% 
          \newline \newline $\left(\begin{array}{r}1083.9 \\  \downarrow~~~ \\ 277.4 \end{array}\right)$
        \\ \hline 
        \multicolumn{6}{c}{~}\\
      \end{tabular}}
      \figcaption{Effect of increasing number of objects (Experiment 1-b)}
      \label{tab:table_add}
    \end{figure*}

Fig.~\ref{tab:table_add}
shows
the results.
The depth values of the target objects were changed without being
affected by the table set.

 \subsubsection{Experiment 1-c: Experiments on real images
  }~\\
In addition to scenes rendered using CG in Experiments
1-a and 1-b,
scenes $S^{(in)}_{9}$, $S^{(in)}_{10}$, and $S^{(in)}_{11}$ that
were real images selected from the training dataset were tested in
this experiment.
The target objects were a table shelf, a door, and a sofa on scenes
$S^{(in)}_{9}$, $S^{(in)}_{10}$, and $S^{(in)}_{11}$, respectively.

\begin{figure*}[t]
      \centering
      {\footnotesize
      \begin{tabular}{@{}l@{~}|@{~~~~}c@{~}c@{~}|@{~~~}c@{~}c@{~}|p{18mm}@{}} \hline
         Scene & \multicolumn{2}{c@{~}|@{~~~}}{Images} & \multicolumn{2}{c@{~}|}{Depth maps} & Pseudo volume\\
               &Original & Adversarial ex. & Original & Adversarial ex.     & reduction rate\\
        \hline
        
        $S^{(in)}_{9}$
        & {\includegraphics[width=26mm]{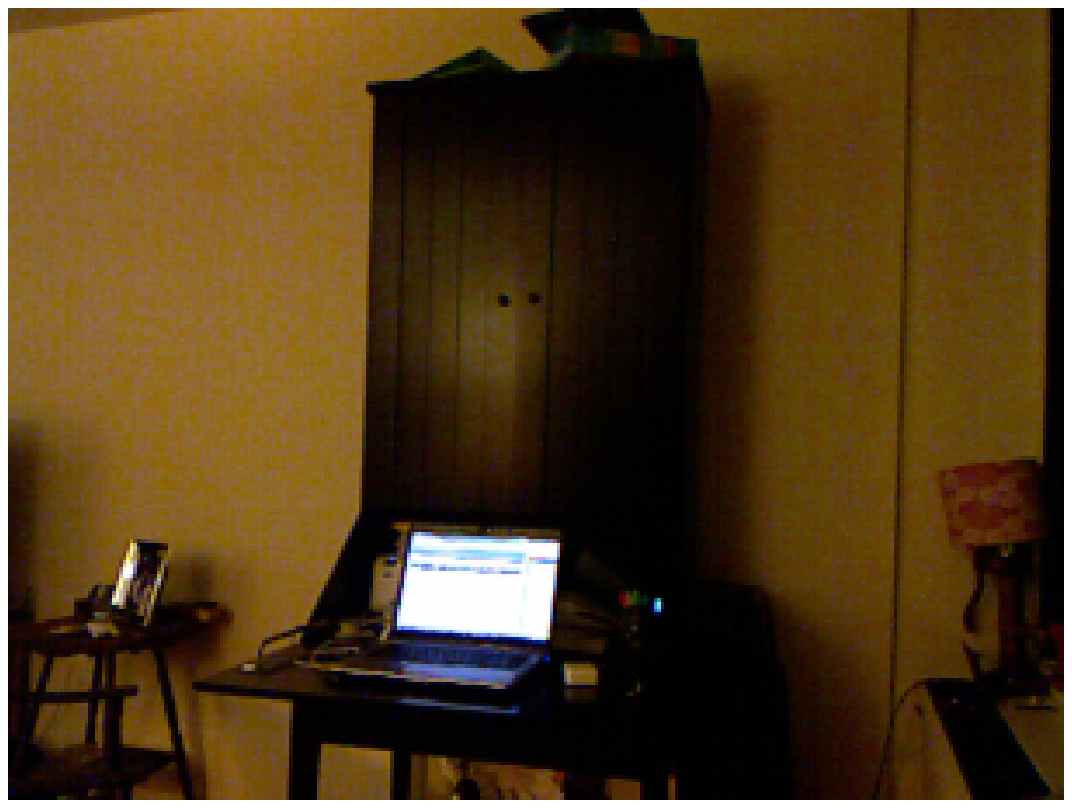}}
        & {\includegraphics[width=26mm]{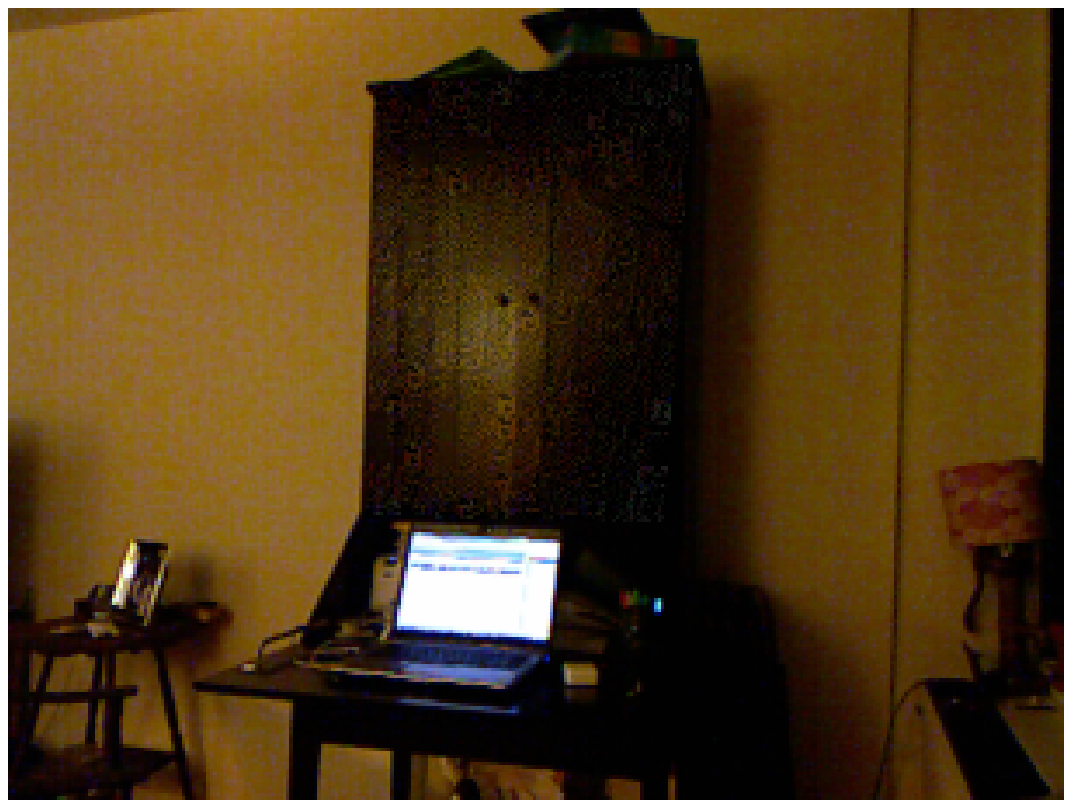}}
        & {\includegraphics[width=26mm]{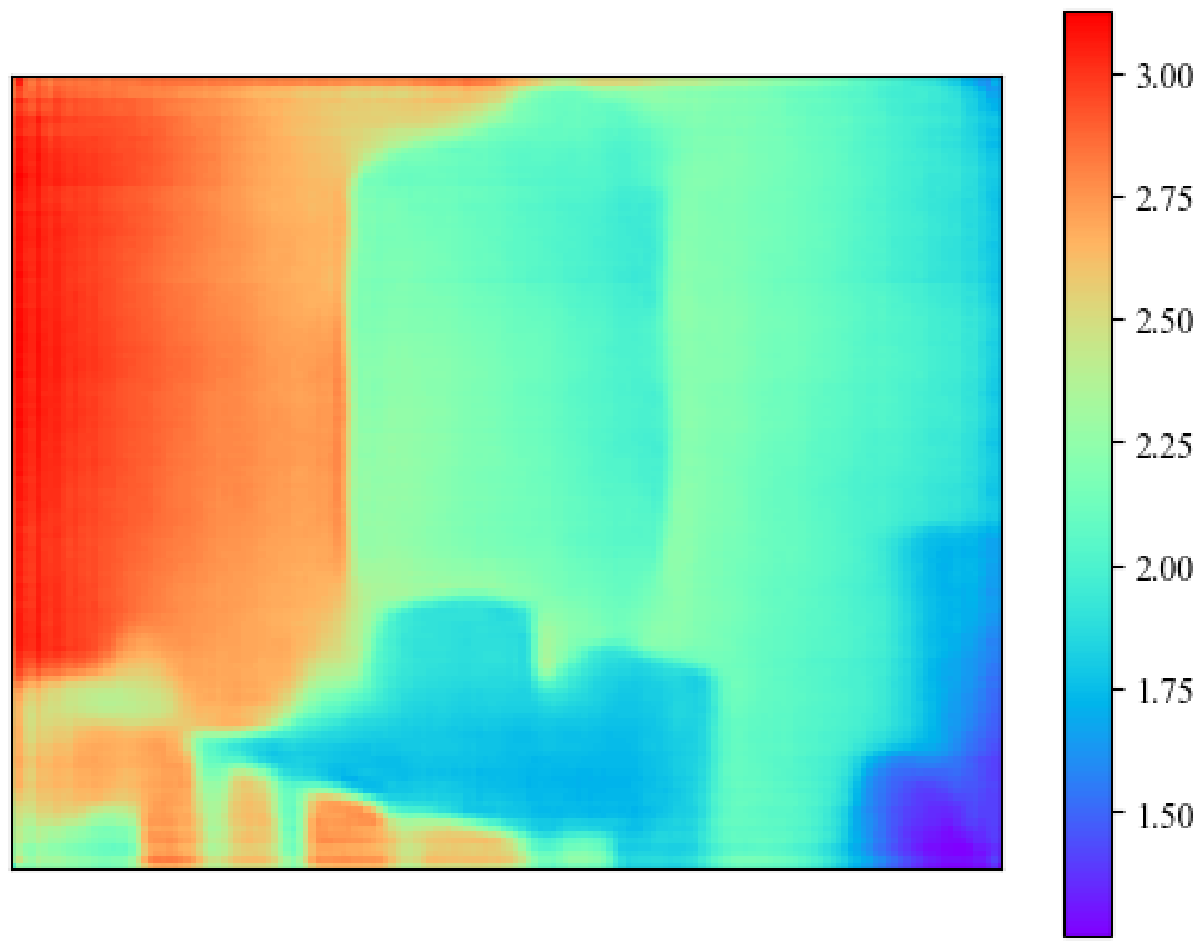}}
        & {\includegraphics[width=26mm]{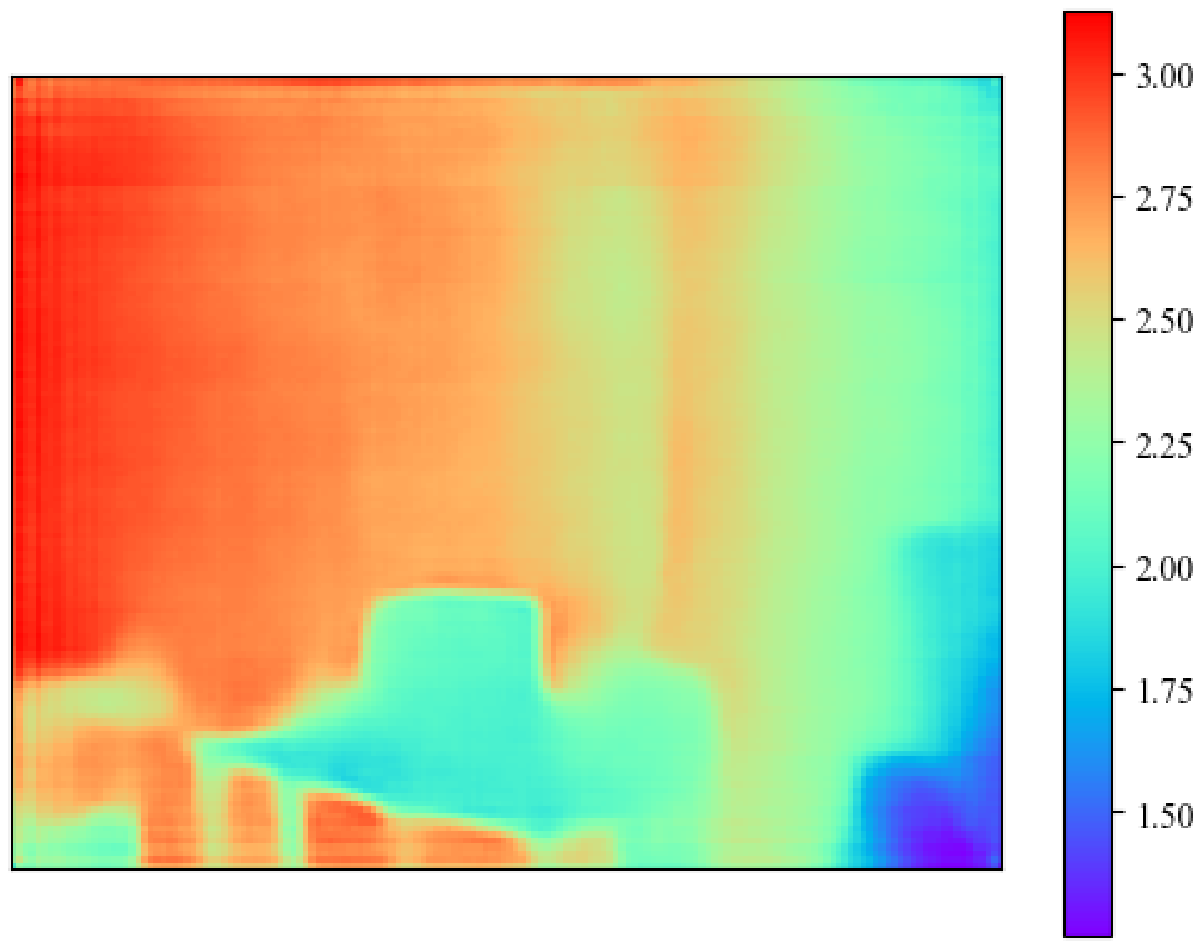}}
        & \vspace*{-18mm}~~~~~~89.9\% 
          \newline \newline $\left(\begin{array}{r}2343.7 \\  \downarrow~~~ \\ 237.1 \end{array}\right)$
        \\ \hline 

        $S^{(in)}_{10}$
        & {\includegraphics[width=26mm]{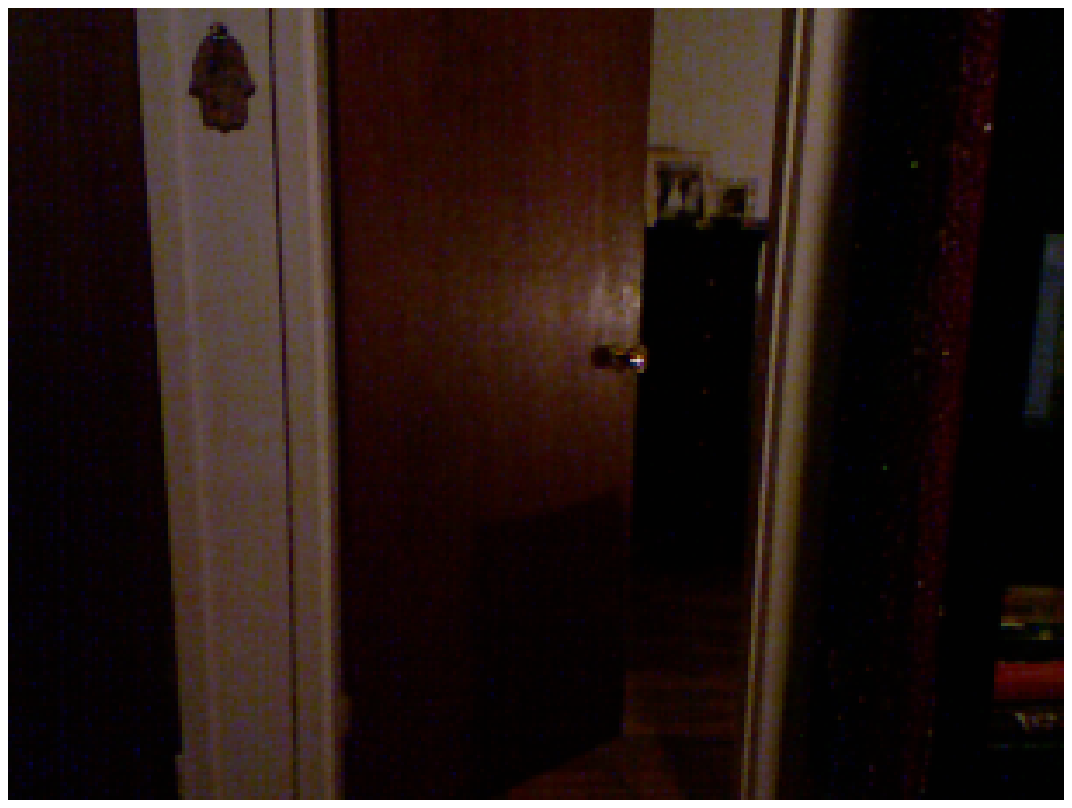}}
        & {\includegraphics[width=26mm]{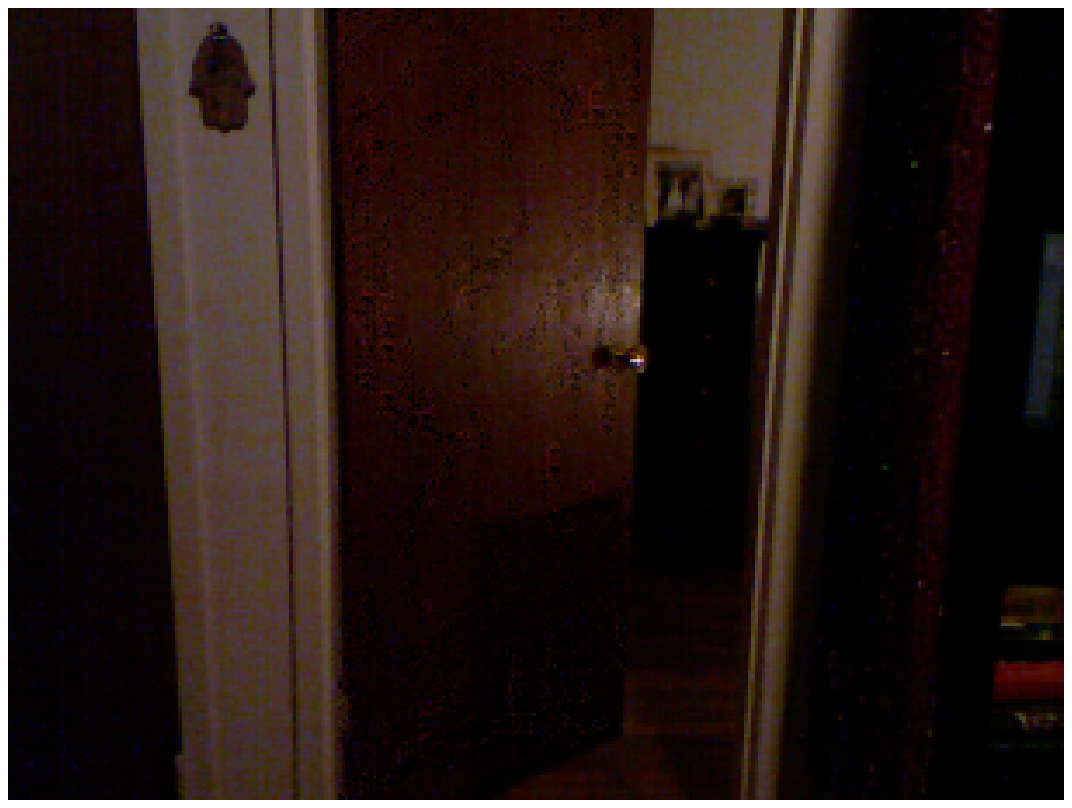}}
        & {\includegraphics[width=26mm]{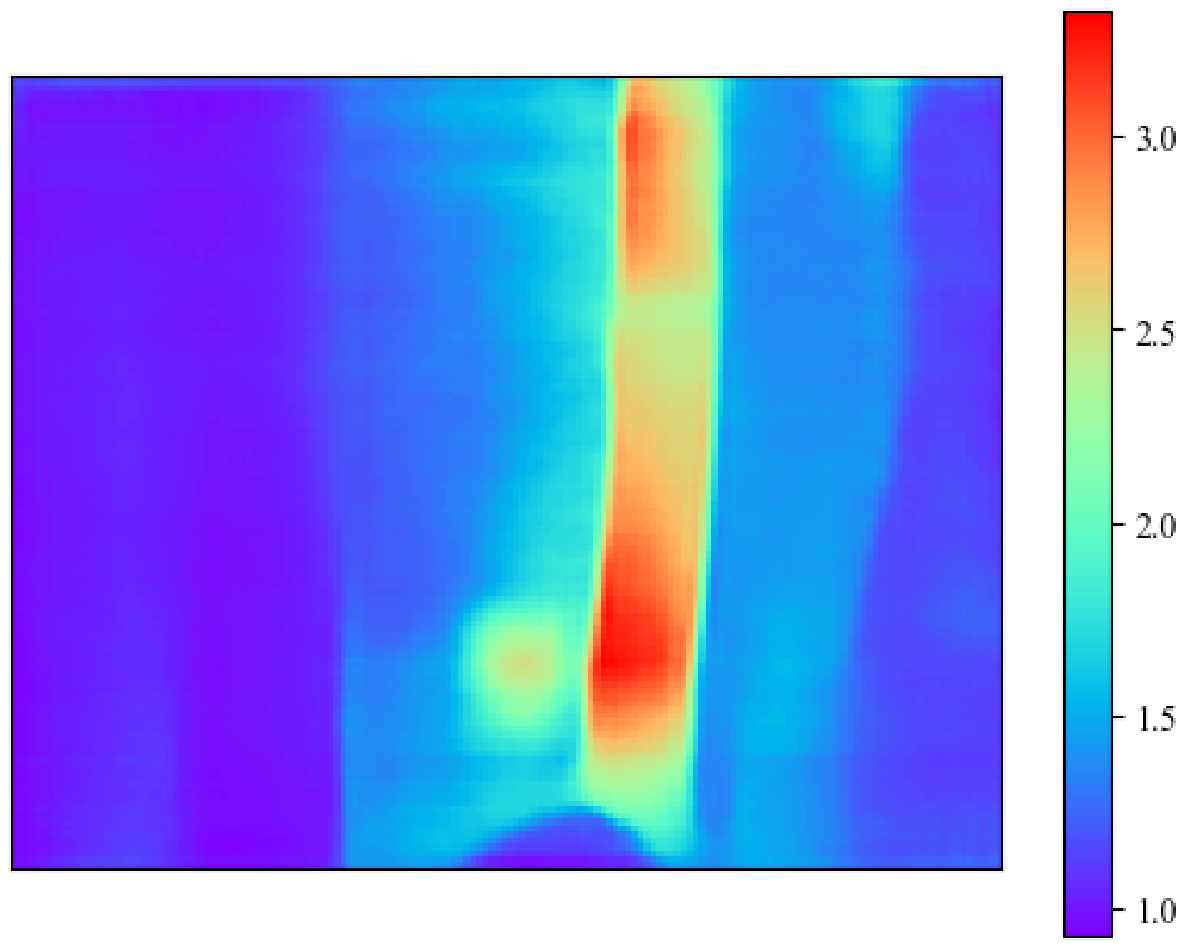}}
        & {\includegraphics[width=26mm]{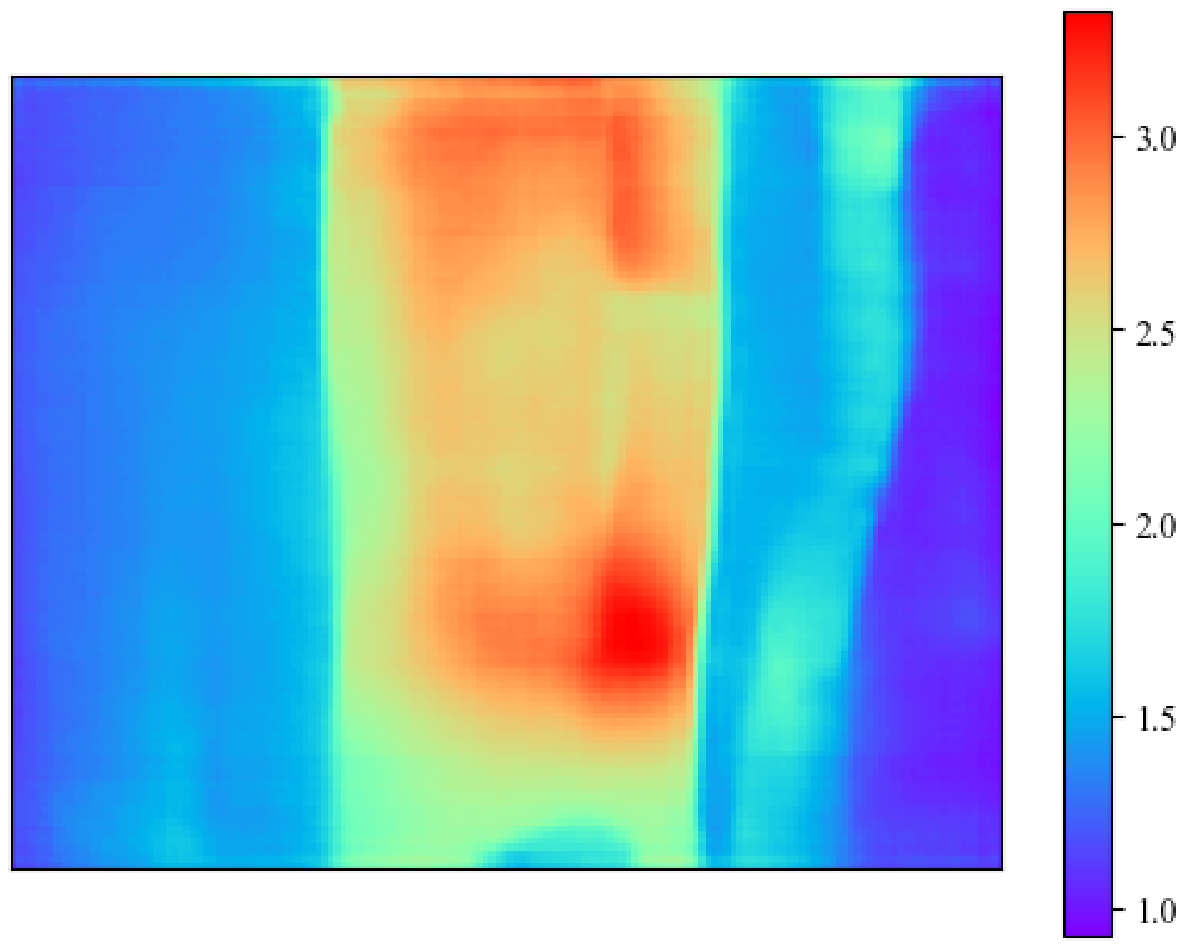}}
        & \vspace*{-18mm}~~~~~~94.8\% 
          \newline \newline $\left(\begin{array}{r}6609.0 \\  \downarrow~~~ \\ 346.2 \end{array}\right)$
        \\ \hline 

        $S^{(in)}_{11}$
        & {\includegraphics[width=26mm]{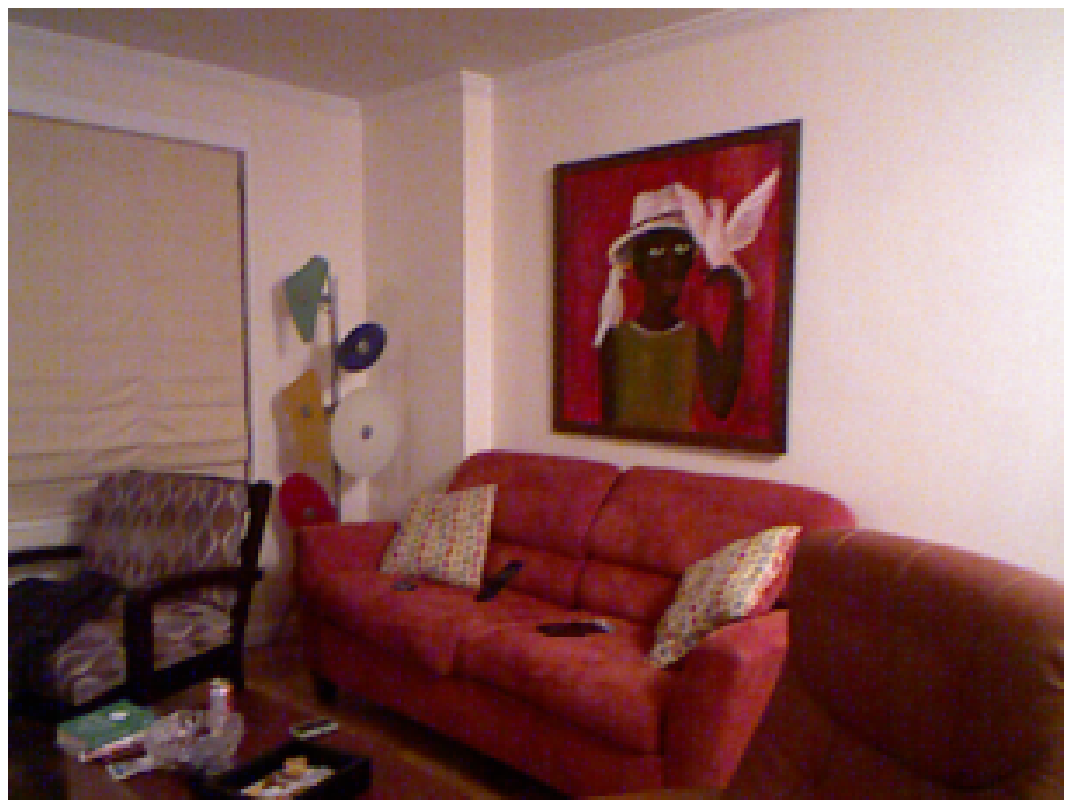}}
        & {\includegraphics[width=26mm]{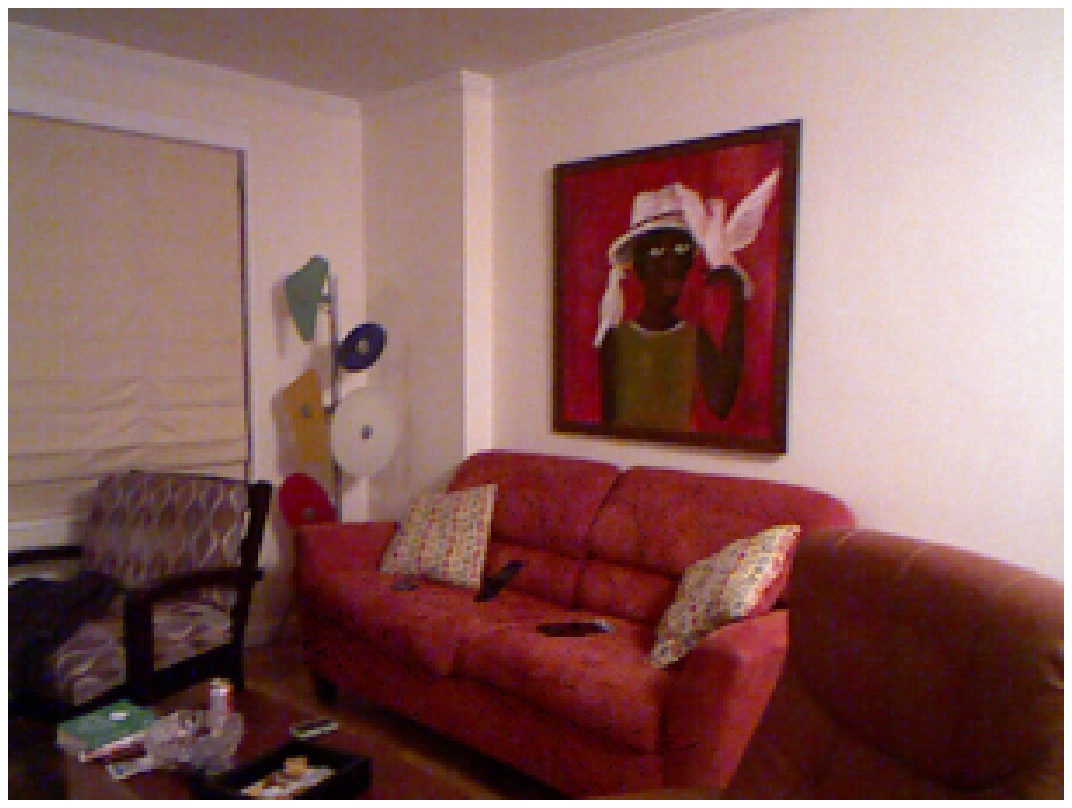}}
        & {\includegraphics[width=26mm]{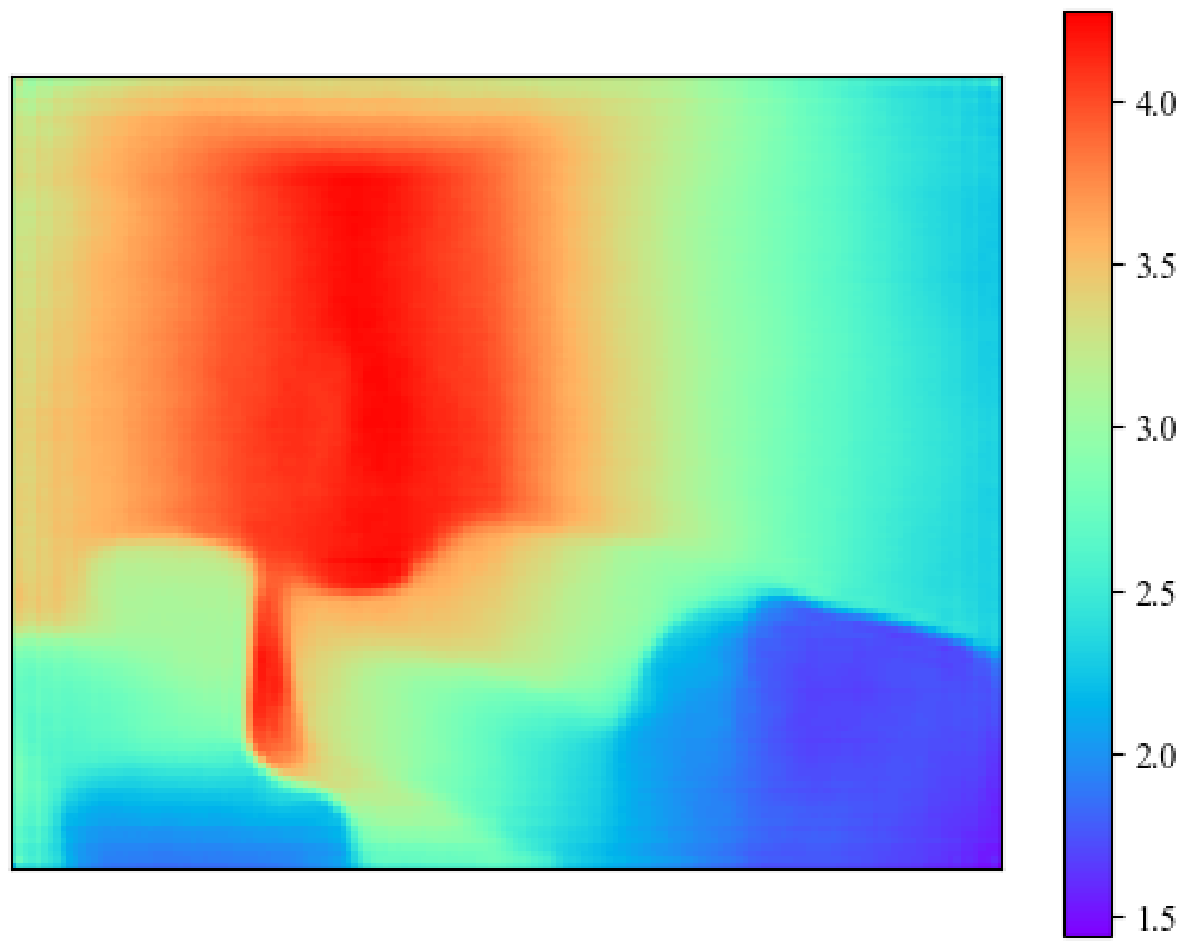}}
        & {\includegraphics[width=26mm]{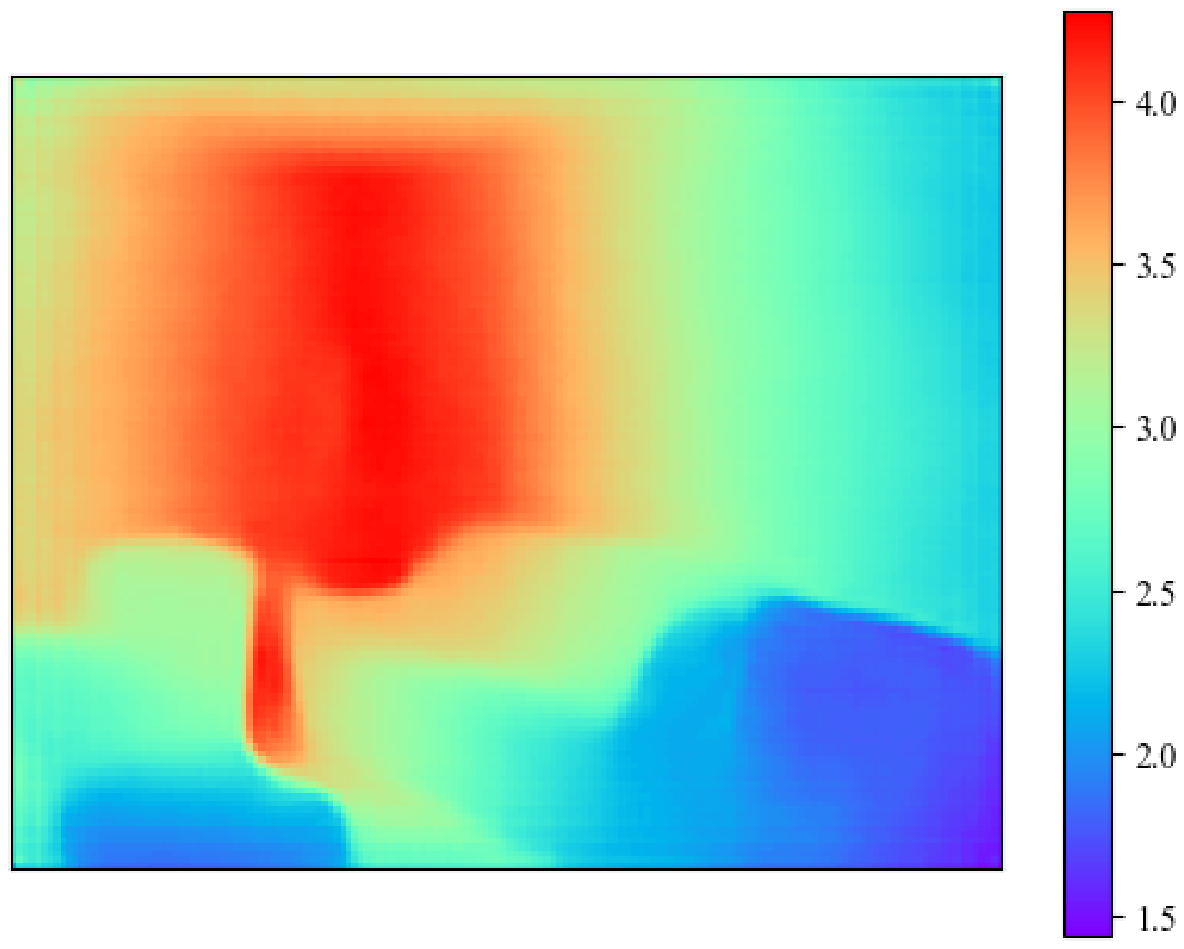}}
        & \vspace*{-18mm}~~~~~~8.2\% 
          \newline \newline $\left(\begin{array}{r}2924.1 \\  \downarrow~~~ \\ 2683.2 \end{array}\right)$
        \\ \hline 
        \multicolumn{6}{c}{~}\\
      \end{tabular}}
      \figcaption{Experiment on real images (Experiment 1-d)}
      \label{tab:table_real}
    \end{figure*}

Fig.~\ref{tab:table_real}
shows
the results.
In scene
 $S^{(in)}_{9}$, 
by adding perturbations, the DNN incorrectly
recognized the scene as if there were no table shelf.
Similarly, in scene 
$S^{(in)}_{10}$, 
the depth estimation result of
the generated adversarial example was as if the door did not exist.
On the contrary, in scene 
$S^{(in)}_{11}$, 
it was difficult to
generate an adversarial example that fools the DNN. 

The results in Experiment 1-a through 1-c suggested that perturbation
on rectangular objects with height is likely to cause erroneous
estimation.

\begin{figure*}[t]
      \centering
      {\footnotesize
      \begin{tabular}{@{}l@{~~}l@{~}|@{~~~}c@{~}|@{~~~}c@{~}|@{~~~}c@{}@{~~~}c@{}} \hline
        \multicolumn{2}{c}{Scene}   & Original & \multicolumn{3}{c@{}}{Adversarial examples}  \\ \cline{4-6}
        \multicolumn{2}{c}{~}       &          & Proposed method & 
        \multicolumn{2}{c@{}}{Previous method~\cite{mathew2020monocular}} \\ 
         & & & & Patch for \cite{godard2019digging} & Patch for \cite{godard2017unsupervised} \\ \hline
        \multirow{2}{*}{\rotatebox{90}{$S^{(out)}_1$}}
        & \rotatebox{90}{Images}
        & {\includegraphics[width=22mm]{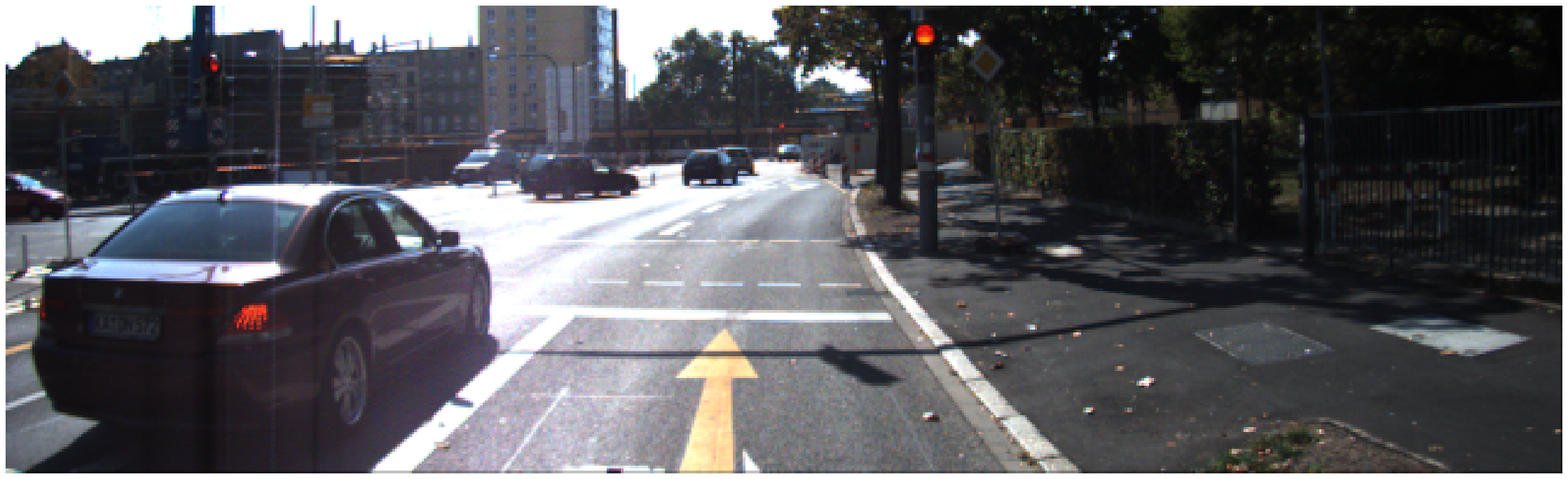}}
        & {\includegraphics[width=22mm]{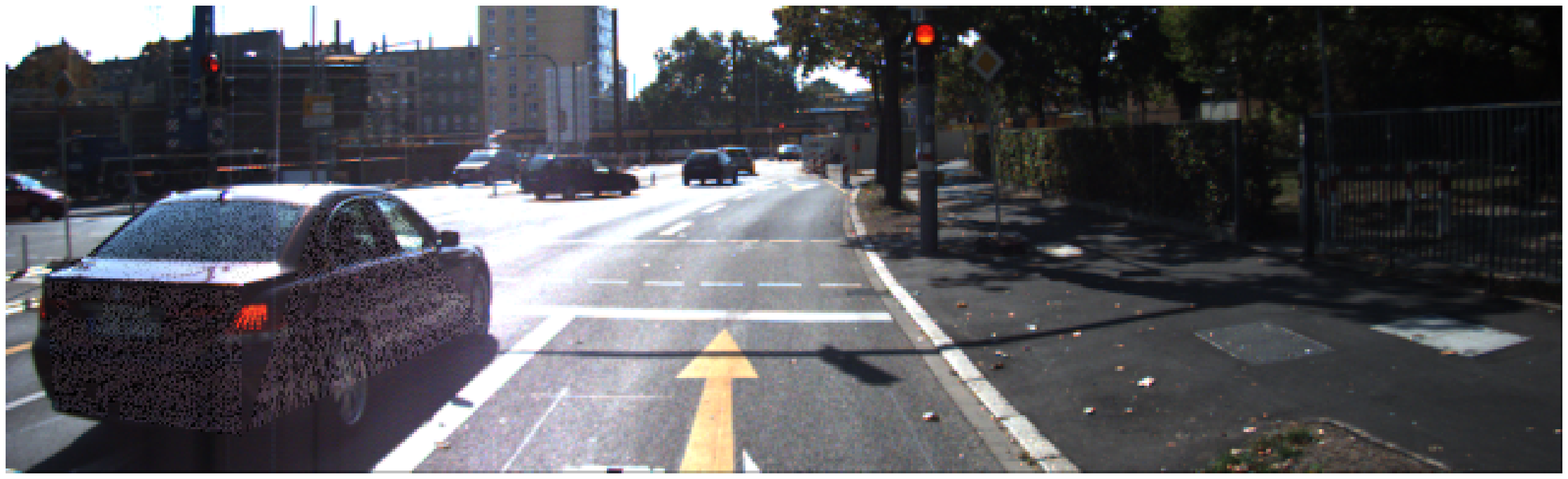}}
        & {\includegraphics[width=22mm]{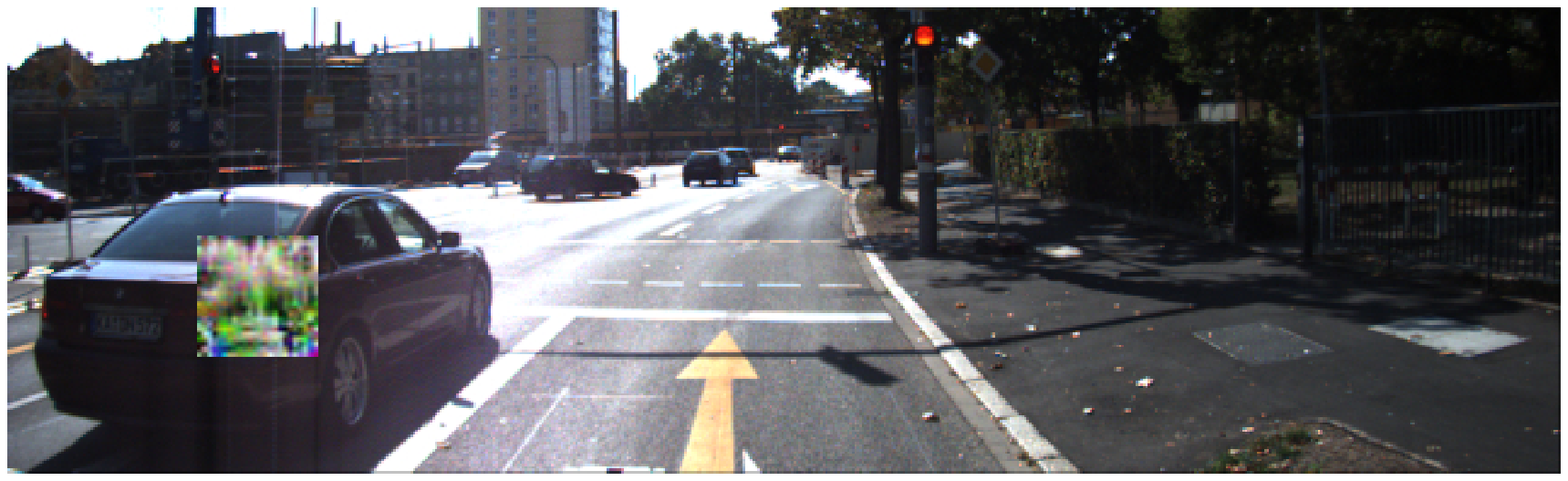}}
        & {\includegraphics[width=22mm]{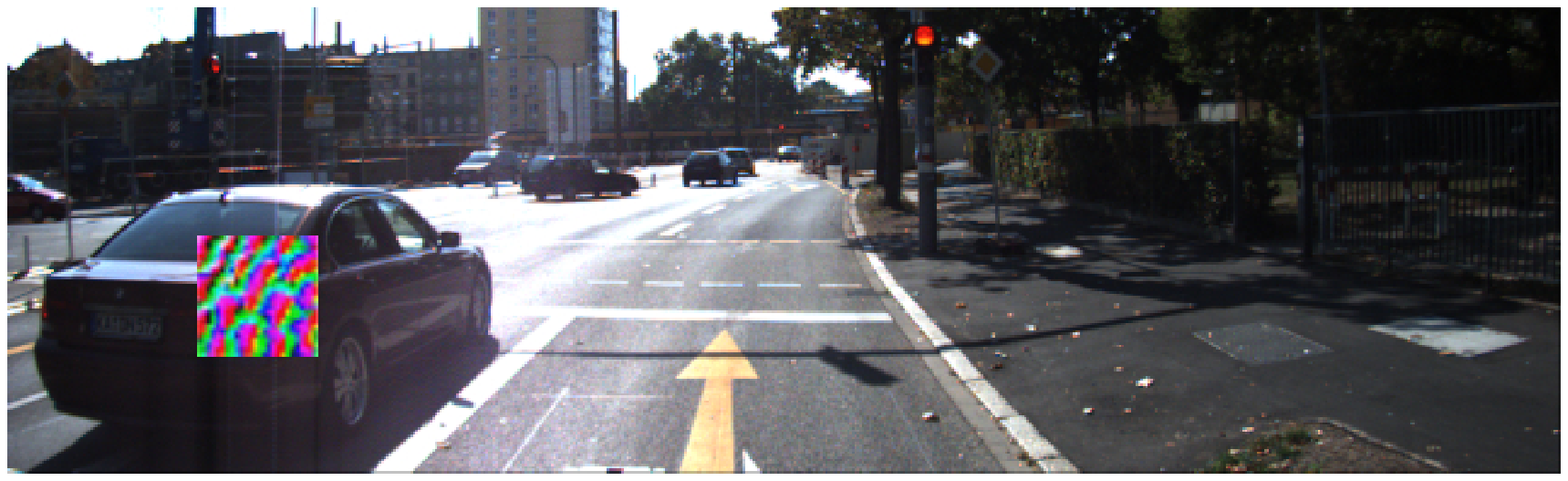}}
        \\
        & \rotatebox{90}{Disparity maps}
        & {\includegraphics[width=28mm]{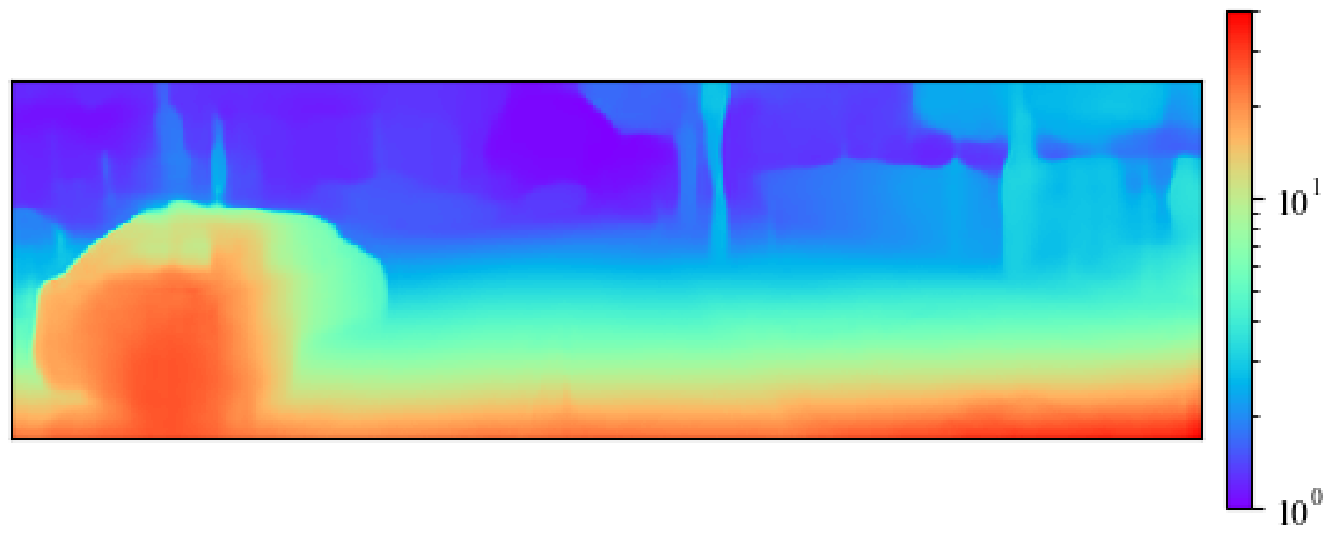}}
        & {\includegraphics[width=28mm]{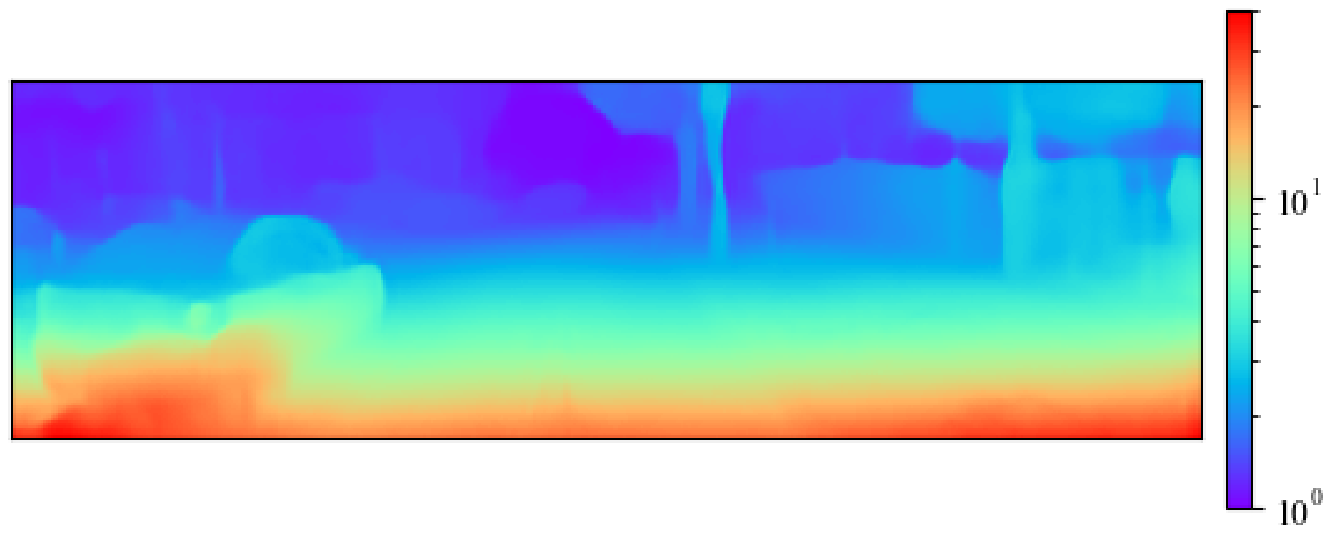}}
        & {\includegraphics[width=26mm]{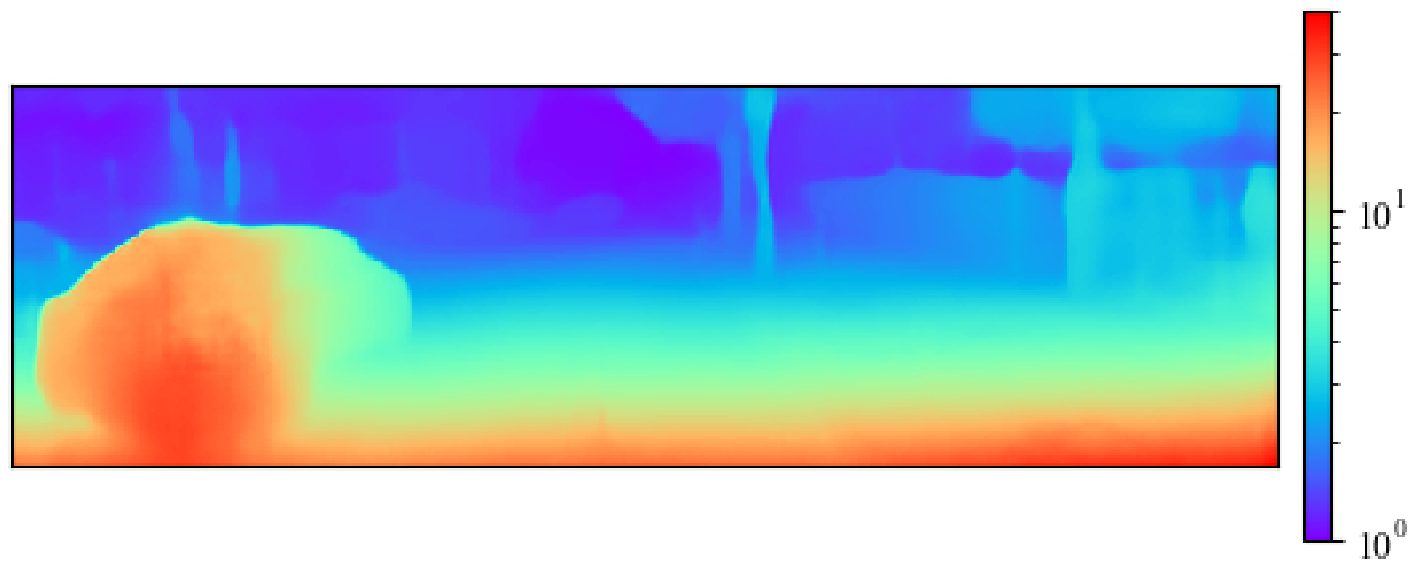}}
        & {\includegraphics[width=28mm]{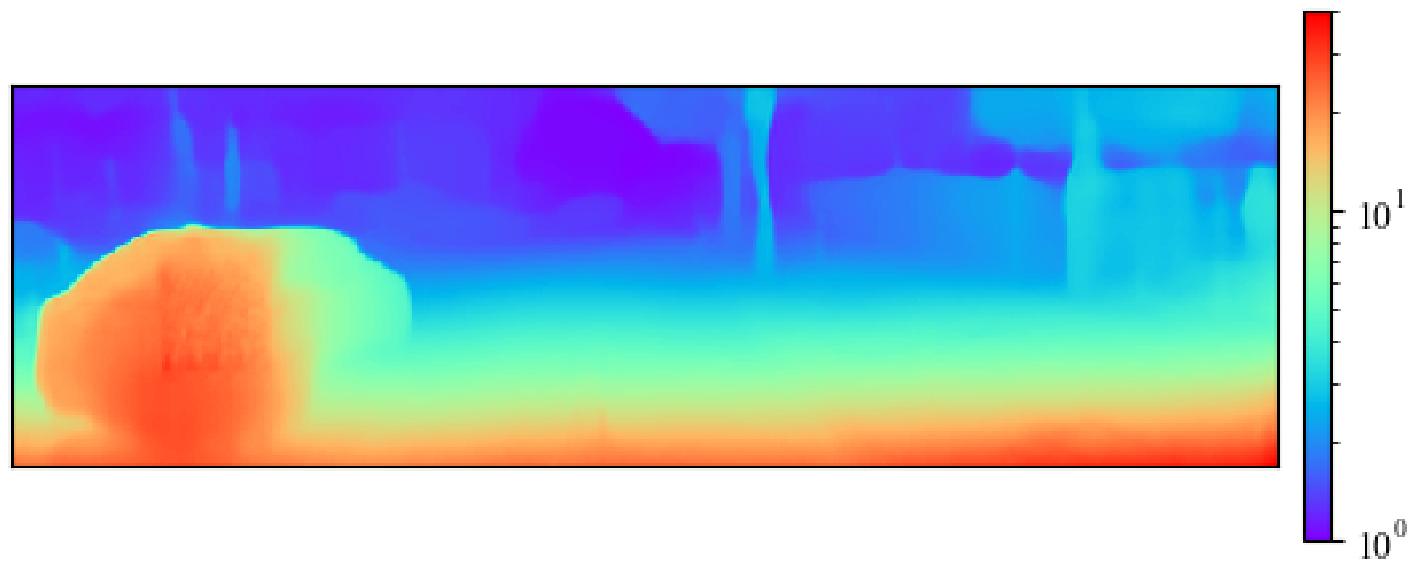}}
        \\
        ~&&&
        \\ \hline 
        \multirow{2}{*}{\rotatebox{90}{$S^{(out)}_2$}}
        & \rotatebox{90}{Images}
        & {\includegraphics[width=22mm]{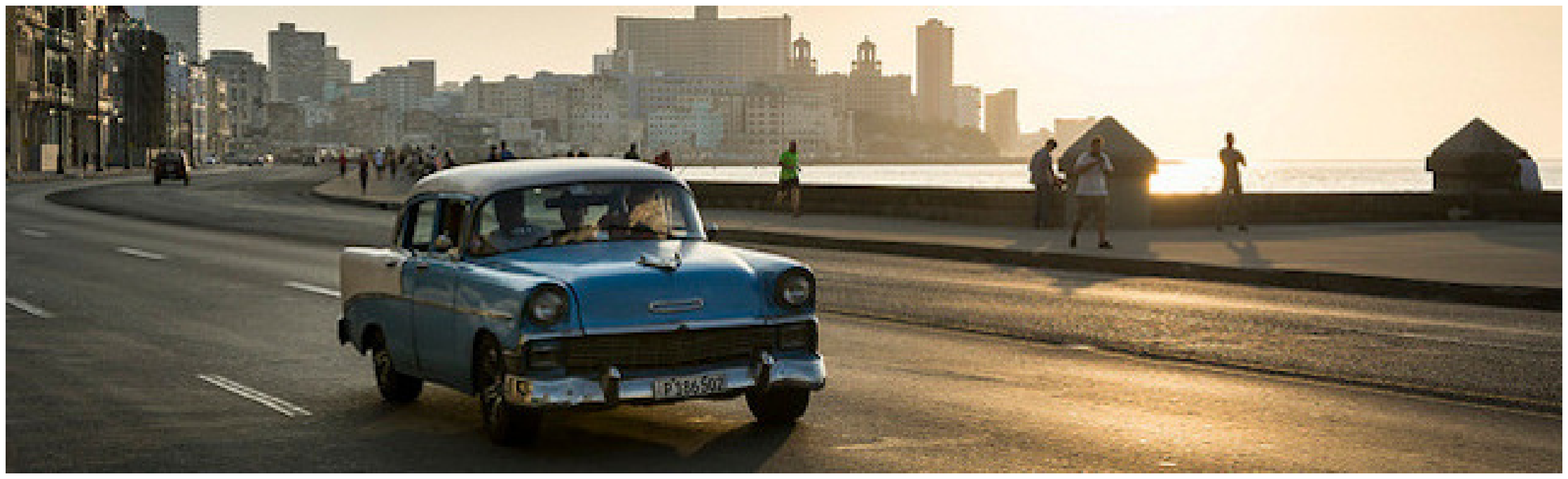}}
        & {\includegraphics[width=22mm]{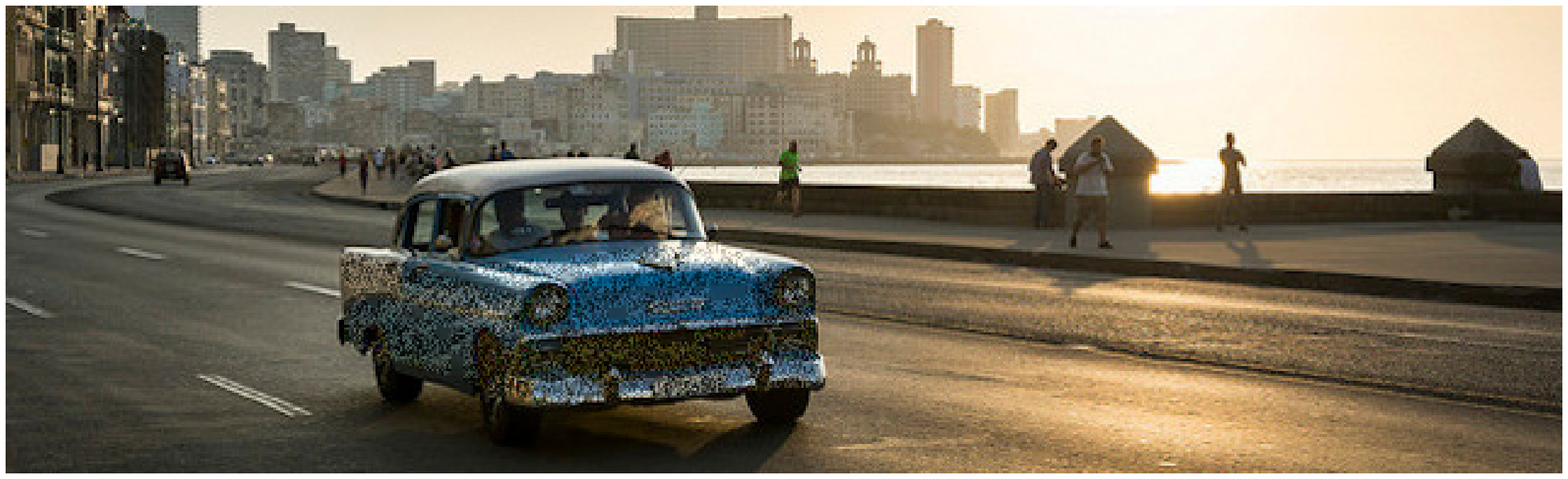}}
        & {\includegraphics[width=22mm]{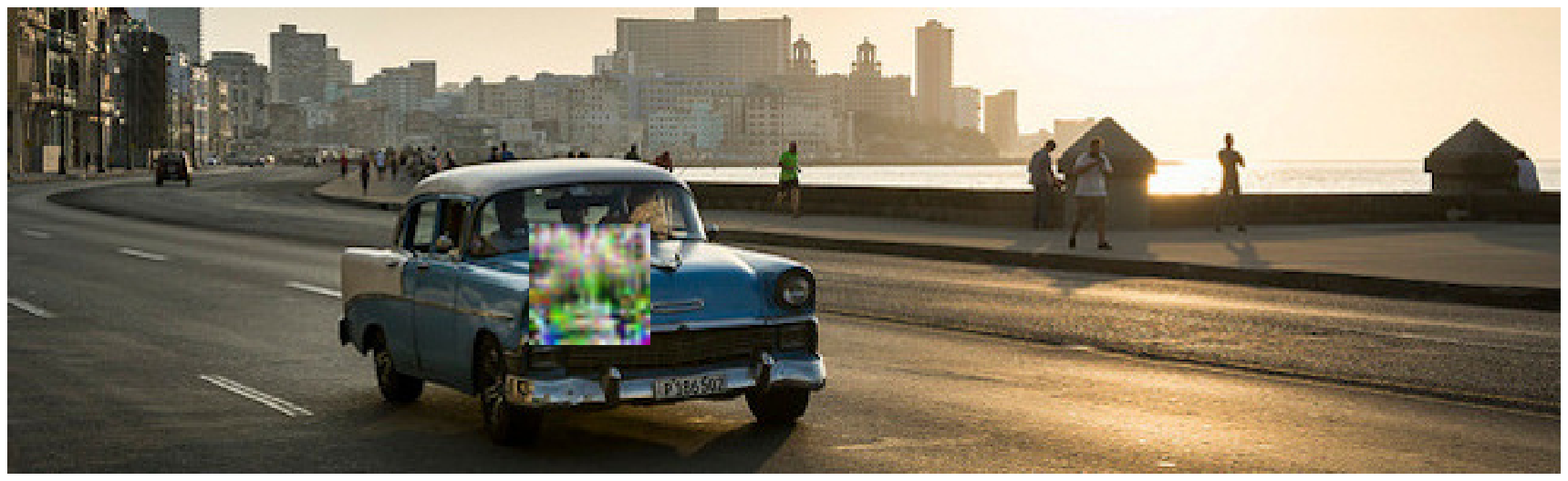}}
        & {\includegraphics[width=22mm]{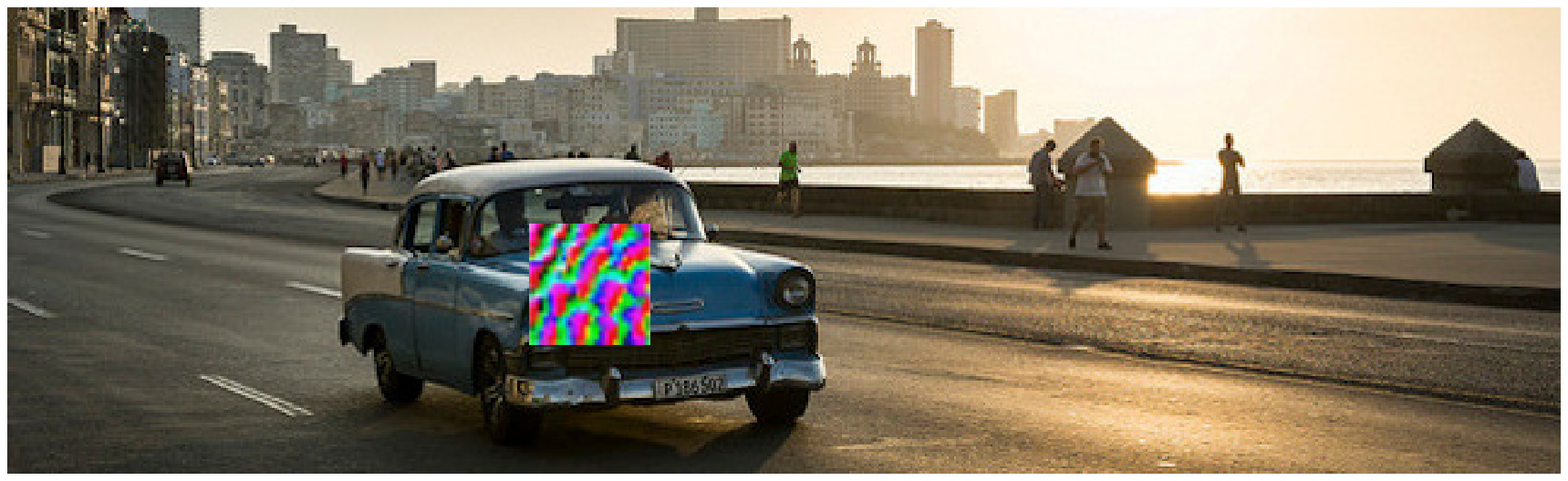}}
        \\
        & \rotatebox{90}{Disparity maps}
        & {\includegraphics[width=28mm]{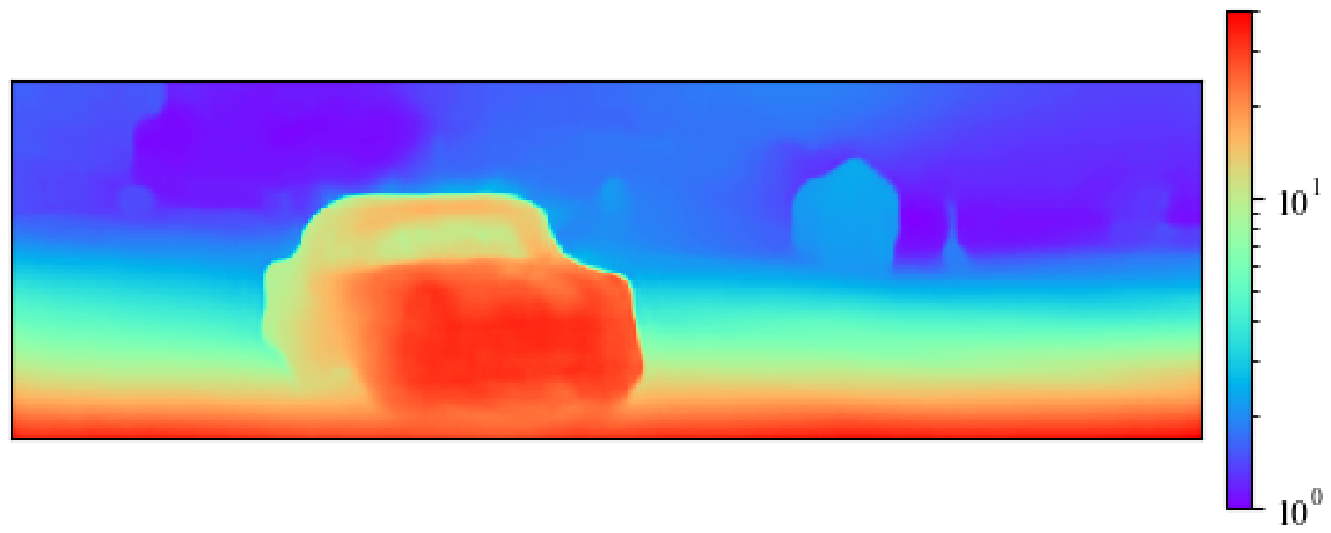}}
        & {\includegraphics[width=28mm]{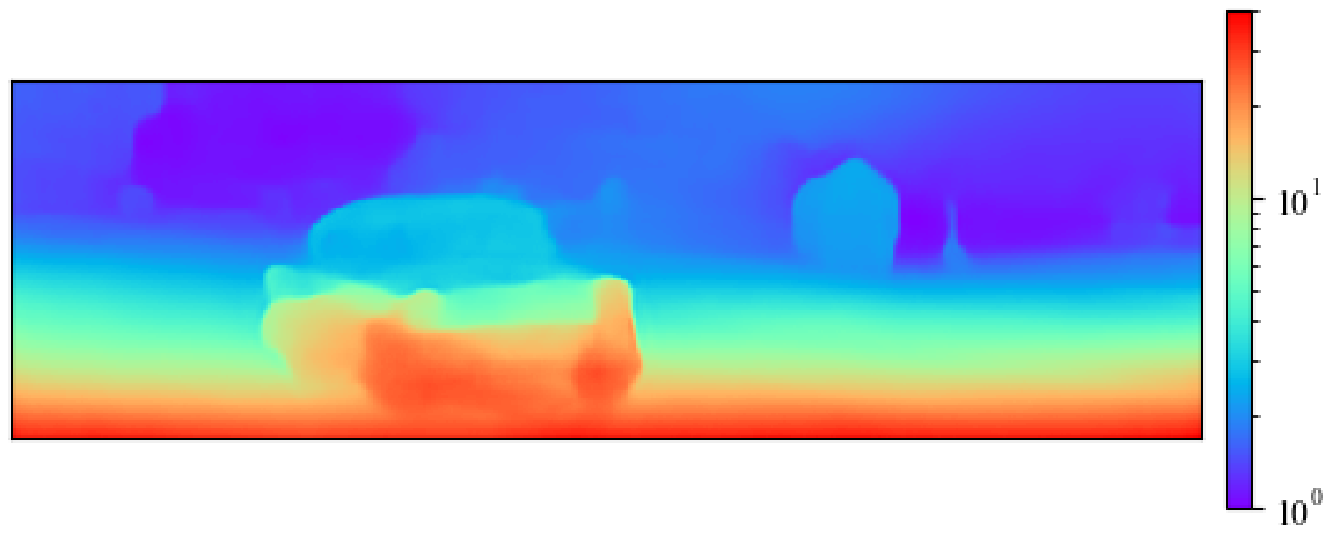}}
        & {\includegraphics[width=28mm]{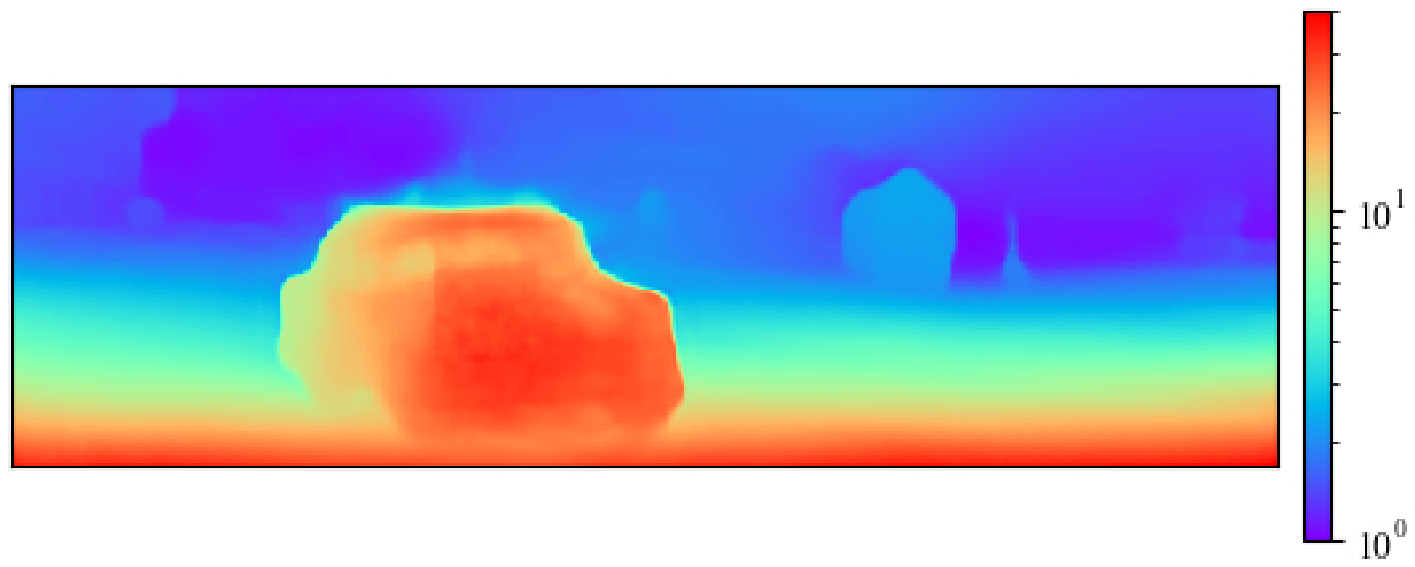}}
        & {\includegraphics[width=28mm]{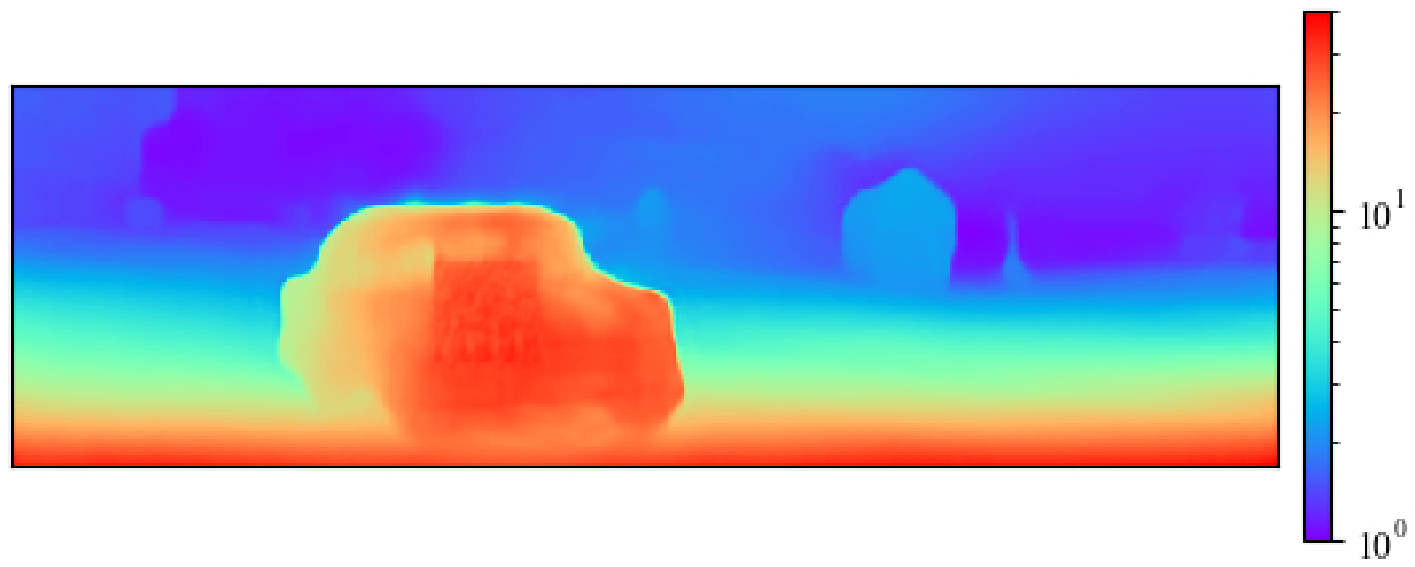}}
        \\
        ~&&&
        \\ \hline

        \multirow{2}{*}{\rotatebox{90}{$S^{(out)}_3$}}
        & \rotatebox{90}{Images}
        & {\includegraphics[width=22mm]{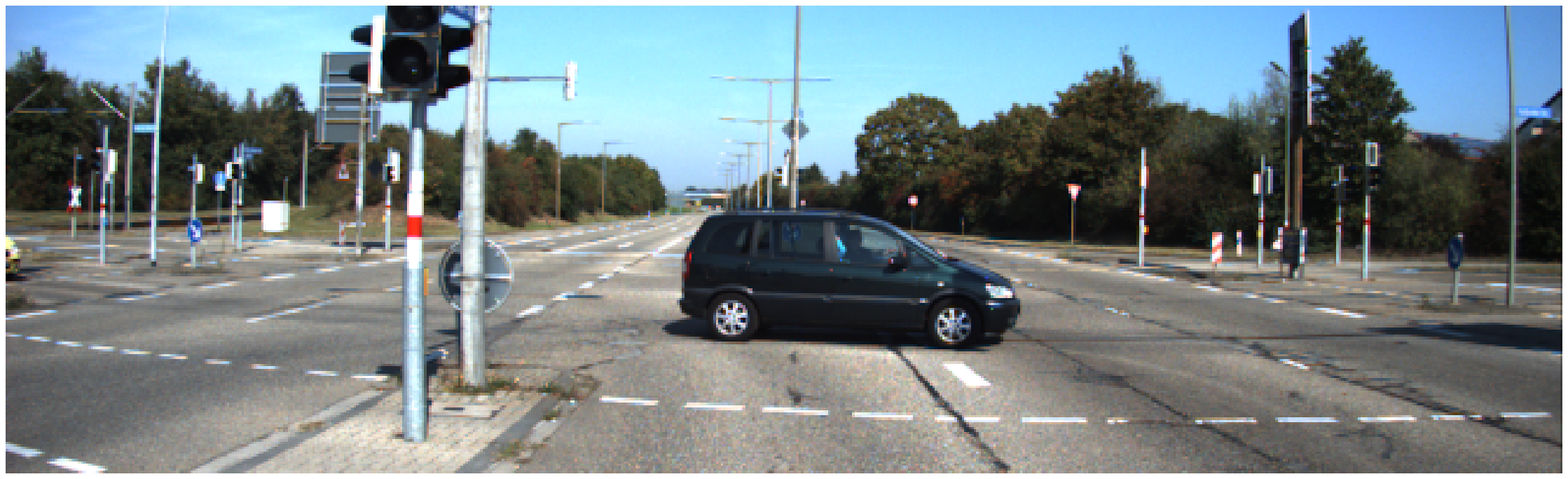}}
        & {\includegraphics[width=22mm]{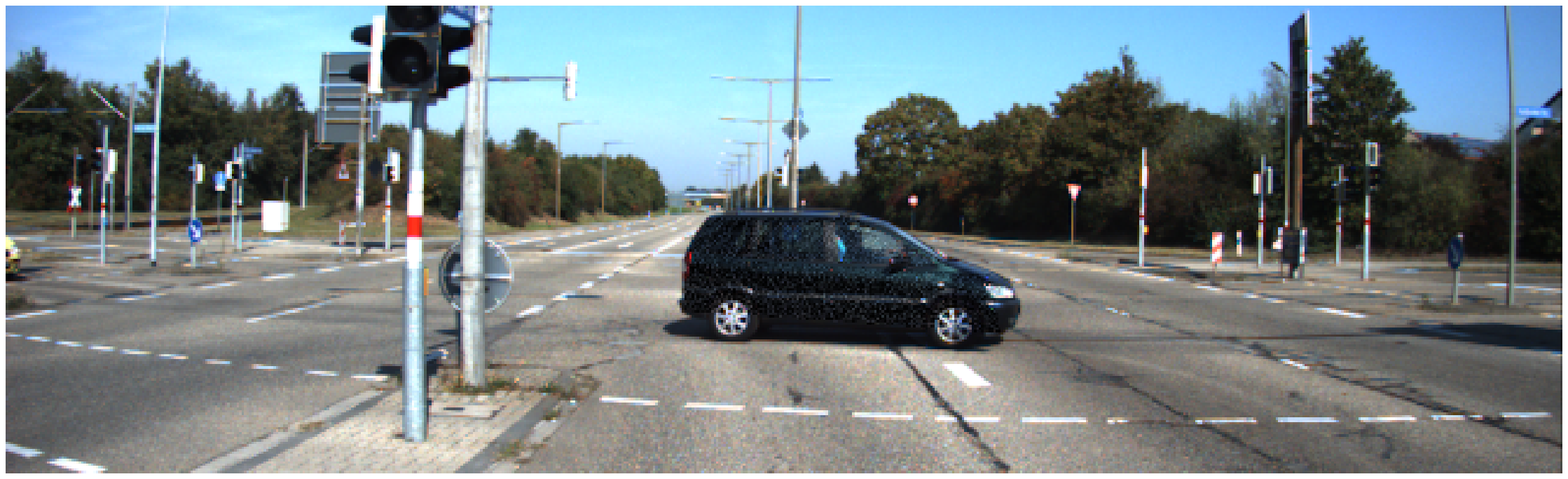}}
        & {\includegraphics[width=22mm]{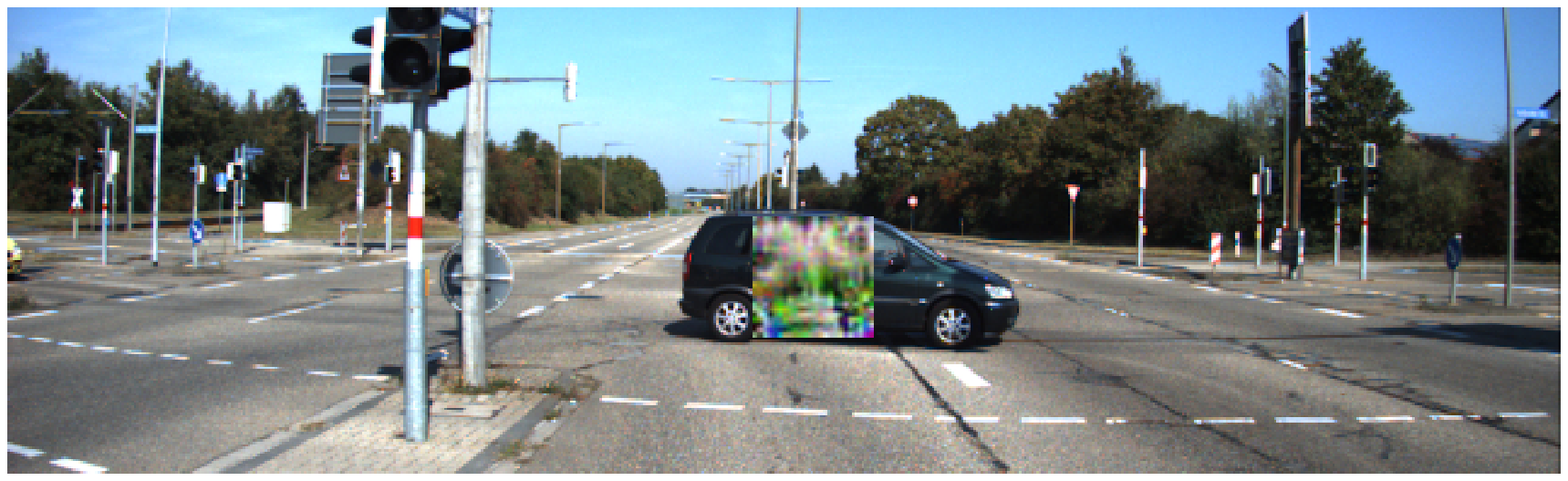}}
        & {\includegraphics[width=22mm]{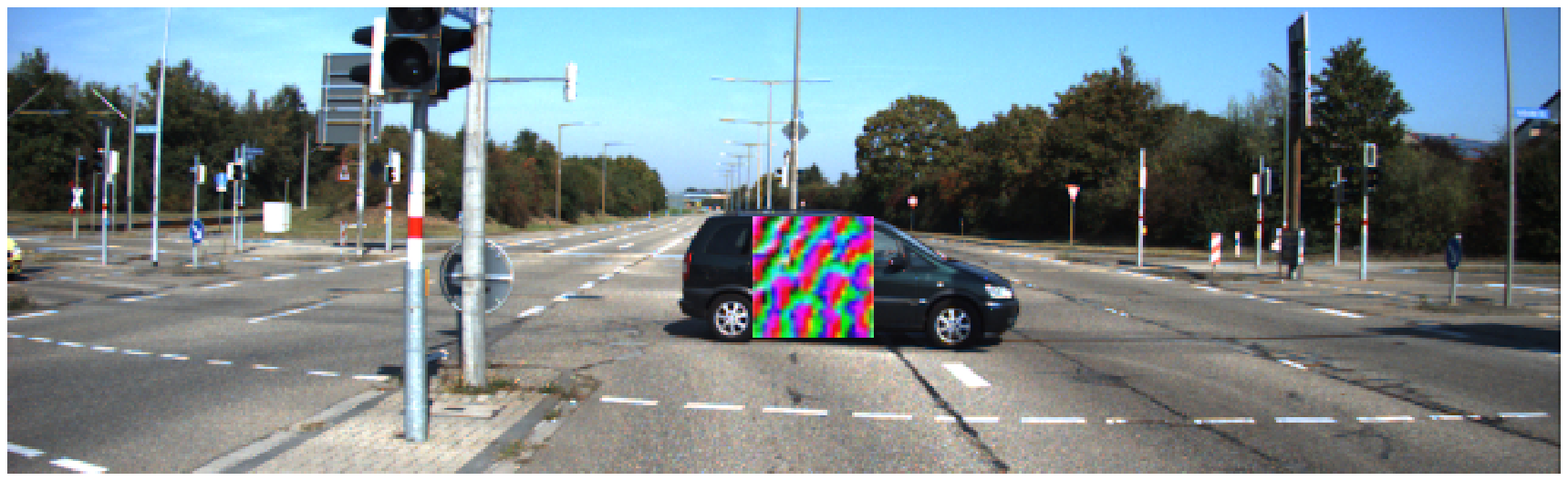}}
        \\
        & \rotatebox{90}{Disparity maps}
        & {\includegraphics[width=28mm]{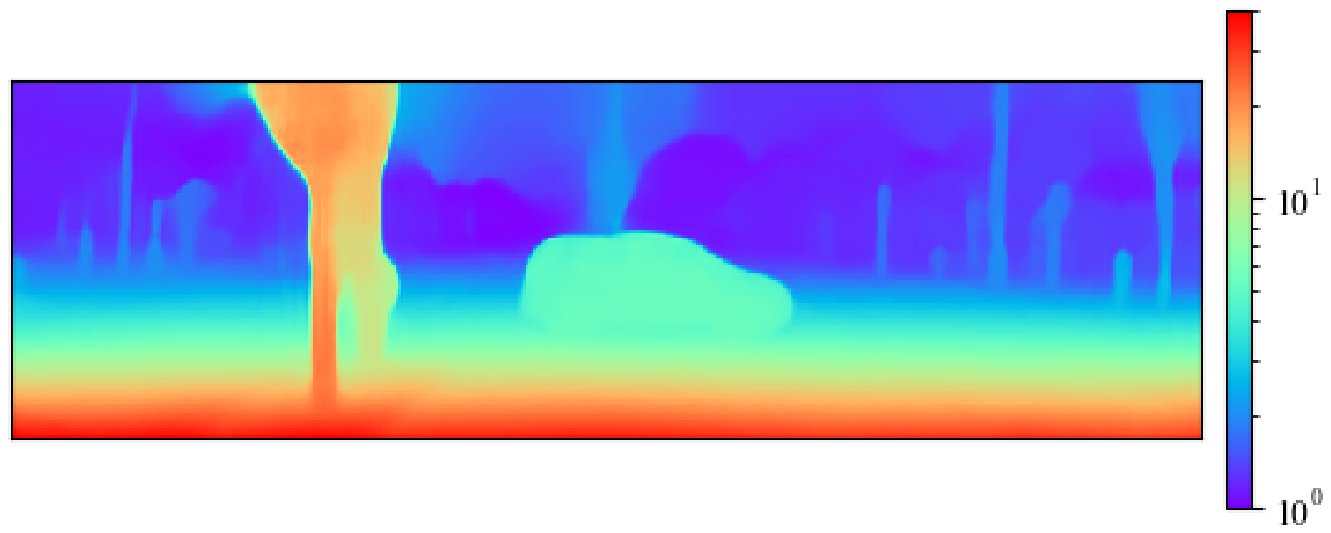}}
        & {\includegraphics[width=28mm]{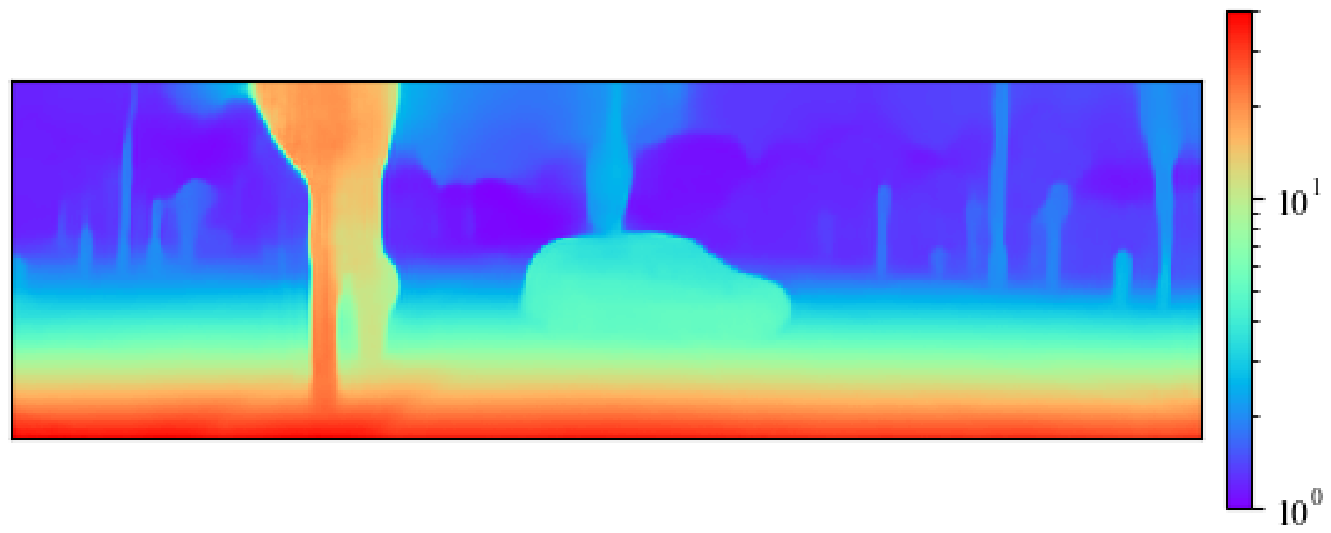}}
        & {\includegraphics[width=28mm]{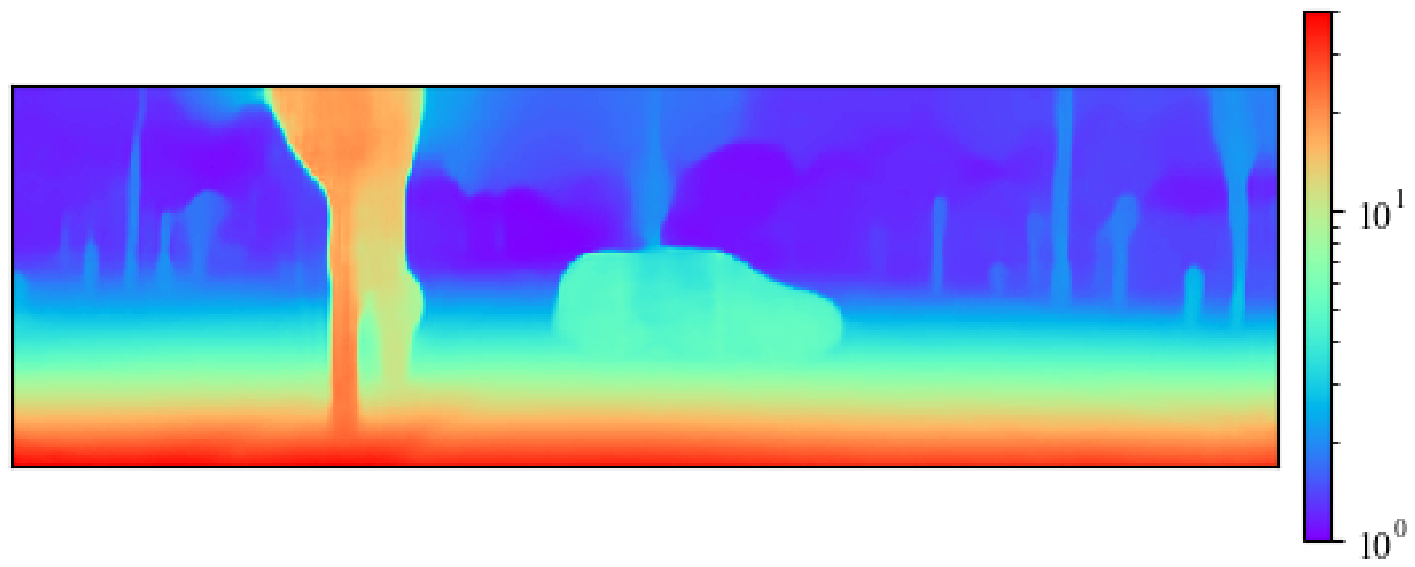}}
        & {\includegraphics[width=28mm]{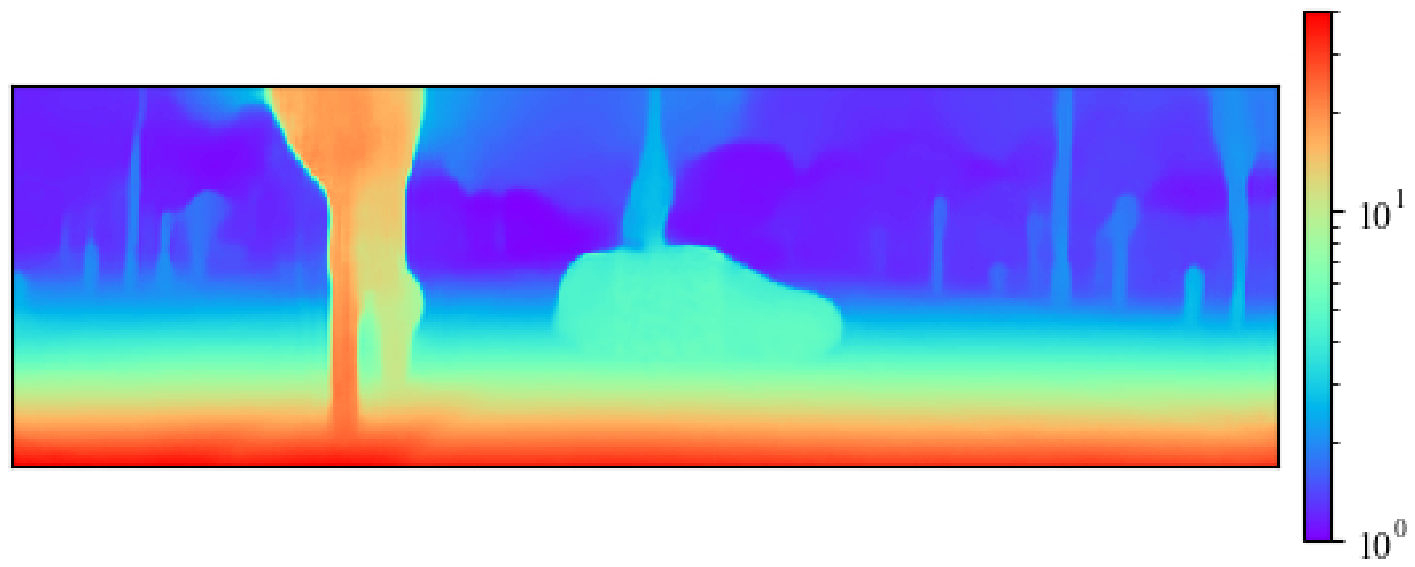}}
        \\
        ~&&&
        \\ \hline 
        ~\\
      \end{tabular}}
      \caption{Results on Godard's model~\cite{godard2019digging} in outdoor scenes (Experiment 2)}
      \label{tab:table_Godard}
    \end{figure*}

  \subsection{Experiment 2: Analysis of a model trained with outdoor scenes
  }

The second target DNN in our experiment is the model proposed by
Godard et al.~\cite{godard2019digging} trained on the outdoor scene dataset KITTI.
Similar to Experiment 1, we employed the trained model~\footnote{%
\url{https://github.com/nianticlabs/monodepth2} 
}
and did not trained the model by ourselves.
The tested scenes $S^{(out)}_1$, $S^{(out)}_2$, and $S^{(out)}_3$ in
this experiment were selected from the training data of the dataset.

Fig.~\ref{tab:table_Godard}
shows the results.
Scenes $S^{(out)}_1$ and $S^{(out)}_2$ were misestimated as scenes
where the target objects did not exist by adding perturbations, while 
the depth of the target object of $S^{(out)}_3$ was not changed.
In detail, in $S^{(out)}_1$, the depth was estimated such that the top
and middle of the target car disappeared, leaving only the bottom of
the vehicle at about the same height as the sidewalk.
Similarly, in $S^{(out)}_2$, the depth was misestimated such that 
the hood and cabin disappeared, leaving only the front bumper.
On the other hand, in scene $S^{(out)}_3$, 
there was little change in the depth of the target vehicle.

\subsection{Experiment 3: Comparison with previous work}

Finally, we compared our method with previous
method~\cite{mathew2020monocular}, which produces universal
adversarial patchs.
This comparison was not fair in the following two points;
first, the previous method designed the universal patch whereas the
proposed method designed perturbations for each scene, and second, the
previous method was white-box and untargeted attack whereas the
proposed method was black-box and targeted attack.
However, the two methods have in common that they apply perturbations
only to target objects.
In addition, most of other previous methods added perturbations to the
entire scene images, making it difficult to develop them into physical
attacks.
For these reasons, this paper compared our method with Mathew's method.

The right two columns of Fig.~\ref{tab:table_Godard} shows the results
using two universal patches~\cite{mathew2020monocular} designed for
Godard's model~\cite{godard2019digging}.
The estimated disparity maps were not changed by the patches designed
by the previous method in all the tested three scenes, whereas the
perturbations by the proopsed method succeeded in misleading Godard's
model in scene $S^{(out)}_1$ and $S^{(out)}_2$.

\section{Conclusion
}
This paper proposed a method to generate adversarial examples that
cause errors of monocular depth estimation under black-box conditions.
The proposed method employs black-box optimization-based approach that
does not require to have a substitute DNN and train it nor any
information about training dataset.
The proposed method successfully suppressed the number of design
variables by introducing a block-wise perturbation scheme and by
adding perturbation only on target object textures.

As demonstrated in the experiment, the proposed black-box attack
method successfully changed the estimated depth of the target objects
by adding perturbations only to the target, whereas previous studies
added perturbations to the entire
image~\cite{hu2019analysis,zhang2020adversarial,wong2020targeted} or
patches the compeletely overwrite the certain area of the
scenes~\cite{weko_200840_1,mathew2020monocular}.
On the other hand, the experiments revealed the future task of this
study where getting DNNs to misestimate the depth of rod-shaped
objects such as dining chairs is more challenging.

At this stage, ths study has not considered reducing computational
costs because we assumed the situation where DNN models are externally
inspected with the permission of the developers to find
vulnerabilities that cannot be found white-box attacks and transfer-based
attacks.
In future, we plan to speed up the proposed method by hybridizing it with 
boundary-based attack methods.

\clearpage

\bibliographystyle{elsarticle-num} 
\bibliography{BBatacks_MDE_arxiv}

\end{document}